\documentclass{article}

\usepackage{microtype}
\usepackage{graphicx}
\usepackage{subfigure}
\usepackage{booktabs} 

\usepackage{amsmath,amsfonts,bm}

\def\eqref#1{equation~\ref{#1}}

\def\1{\bm{1}}

\def\vx{{\bm{x}}}
\def\vy{{\bm{y}}}

\DeclareMathAlphabet{\mathsfit}{\encodingdefault}{\sfdefault}{m}{sl}
\SetMathAlphabet{\mathsfit}{bold}{\encodingdefault}{\sfdefault}{bx}{n}

\usepackage{wrapfig}
\usepackage{lipsum}
\usepackage{enumitem}
\usepackage{multirow}
\usepackage{epsfig}
\usepackage{url}
\usepackage{amsmath}
\usepackage{comment}
\usepackage[group-separator={,},group-minimum-digits=4]{siunitx}
\usepackage{threeparttable}
\usepackage{tabularx}
\usepackage{colortbl}
\usepackage{float}
\usepackage{tikz}
\usepackage{sidecap}
\sidecaptionvpos{figure}{t}

\definecolor{natural}{rgb}{0.7137,0.3333,0.3333}
\definecolor{specialized}{rgb}{0.4118,0.6431,0.4314}
\definecolor{structured}{rgb}{0.3254,0.4431,0.6666}
\definecolor{all}{rgb}{0.7529,0.4902,0.6471}

\usepackage{hyperref}

\usepackage[capitalise]{cleveref}

\newcommand{\myparagraph}[1]{\noindent\textbf{#1}\quad}

\def\t{T}

\DeclareRobustCommand{\taskNatural}{\raisebox{0.5pt}{\tikz{\fill[natural] (0cm,0cm) circle (.5ex);}}\,\textsc{natural}}
\DeclareRobustCommand{\taskSpecialized}{\raisebox{0.5pt}{\tikz{\fill[specialized] (0,0) circle (.5ex);}}\,\textsc{specialized}}
\DeclareRobustCommand{\taskStructured}{\raisebox{0.5pt}{\tikz{\fill[structured] (0,0) circle (.5ex);}}\,\textsc{structured}}

\newcommand{\imagenet}{ImageNet}

\newcommand{\dataspace}{X}
\newcommand{\labelspace}{Y}

\renewcommand{\paragraph}[1]{{\bf #1}\,\,}

\usepackage[accepted]{icml2020}

\icmltitlerunning{A Large-scale Study of Representation Learning with the Visual Task Adaptation Benchmark}

\begin{document}

\twocolumn[
\icmltitle{A Large-scale Study of Representation Learning with the\\Visual Task Adaptation Benchmark}



\icmlsetsymbol{equal}{*}

\begin{icmlauthorlist}
  \icmlauthor{Xiaohua Zhai}{equal}
  \icmlauthor{Joan Puigcerver}{equal}
  \icmlauthor{Alexander Kolesnikov}{equal}
  \icmlauthor{Pierre Ruyssen}{}
  \icmlauthor{Carlos Riquelme}{}
  \icmlauthor{Mario Lucic}{}
  \icmlauthor{Josip Djolonga}{}
  \icmlauthor{Andr\'{e} Susano Pinto}{}
  \icmlauthor{Maxim Neumann}{}
  \icmlauthor{Alexey Dosovitskiy}{}
  \icmlauthor{Lucas Beyer}{}
  \icmlauthor{Olivier Bachem}{}
  \icmlauthor{Michael Tschannen}{}
  \icmlauthor{Marcin Michalski}{}
  \icmlauthor{Olivier Bousquet}{}
  \icmlauthor{Sylvain Gelly}{}
  \icmlauthor{Neil Houlsby}{}\\
  \vspace{2mm}
  Google Research, Brain Team, Z\"urich
\end{icmlauthorlist}

\icmlcorrespondingauthor{Neil Houlsby}{neilhoulsby@google.com}

\icmlkeywords{Machine Learning, ICML}

\vskip 0.3in
]



\printAffiliationsAndNotice{\icmlEqualContribution} 

\begin{abstract}
Representation learning promises to unlock deep learning for the long tail of vision tasks without expensive labelled datasets. Yet, the absence of a unified evaluation for general visual representations hinders progress. Popular protocols are often too constrained (linear classification), limited in diversity (ImageNet, CIFAR, Pascal-VOC), or only weakly related to representation quality (ELBO, reconstruction error). We present the Visual Task Adaptation Benchmark (VTAB), which defines good representations as those that adapt to \emph{diverse}, \emph{unseen} tasks with \emph{few examples}. With VTAB, we conduct a large-scale study of many popular publicly-available representation learning algorithms. We carefully control confounders such as architecture and tuning budget. We address questions like: How effective are ImageNet representations beyond standard natural datasets? How do representations trained via generative and discriminative models compare? To what extent can self-supervision replace labels? And, how close are we to general visual representations?
\end{abstract}

\section{Introduction}

Distributed representations learned from raw pixels have enabled unprecedented performance on many visual understanding tasks.
Hand-crafted features have been replaced with hand-annotated datasets, with thousands to millions of examples~\citep{cifar10,imagenet}.
By contrast, humans learn a wide range vision tasks using just a few examples per task.
A key research challenge is to close this gap in sample efficiency, and unlock deep learning for the long tail of problems without many labels.

Improving sample efficiency has been approached from many angles:
few-shot learning~\citep{li2006},
transfer learning~\citep{pan2009survey},
domain adaptation~\citep{wang2018deep},
and representation learning~\citep{bengio2013representation}.
Representation learning is studied in many contexts:
supervised pre-training~\citep{razavian2014},
self-supervised learning~\citep{doersch2015unsupervised},
semi-supervised learning~\citep{chapelle2009semi},
generative modeling~\citep{donahue2016adversarial},
and disentanglement learning~\citep{higgins2017beta}.
Each sub-domain has its own evaluation protocol, and the lack of a common benchmark impedes progress.
Benchmarks have been critical in other sub-fields, such as RL~\citep{atari}, image classification~\citep{imagenet}, and NLP~\citep{glue}.
Inspired by these successes, we propose a benchmark with similar principles:
(i) minimal constraints to encourage creativity,
(ii) a focus on practical considerations, and
(iii) make it challenging.

We present the Visual Task Adaptation Benchmark (VTAB).
Using VTAB, we perform the first extensive cross sub-field study of representation learning.
VTAB defines a good representation as one that can be used to solve many diverse \emph{previously unseen} tasks with the \emph{fewest possible labels}.
We consider low sample complexity to be the key objective of representation learning.
Task diversity is also crucial to assess generality.
Therefore, VTAB goes beyond standard natural tasks, and includes those related to sensorimotor control, medical imaging, and scene understanding.

In our study, we investigate many representation learning algorithms, pre-trained on \imagenet{}.
We carefully control confounding factors such as tuning budget, architecture, and pre-training data.
This study quantifies existing intuitions and reveals new insights:
(i)~Supervised \imagenet{} pre-training yields excellent representations for natural image classification tasks.
Interestingly, it also yields a smaller, but consistent improvement on specialized tasks (e.g. medical imaging).
However, these representations are extremely limited for tasks that require structured understanding.
(ii)~Self-supervised is less effective than supervised learning overall, but surprisingly, can improve structured understanding.
(iii)~Combining supervision and self-supervision is effective, and to a large extent, self-supervision can replace, or compliment labels.
(iv)~Discriminative representations appear more effective than those trained as part of a generative model, with the exception of adversarially trained encoders.
(v)~GANs perform relatively better on data similar to their pre-training source (here, \imagenet{}), but worse on other tasks.
(vi)~Evaluation using a linear classifier leads to poorer transfer and different conclusions.

\section{The Visual Task Adaptation Benchmark}

\begin{figure}[h]
\centering
\includegraphics[width=\linewidth]{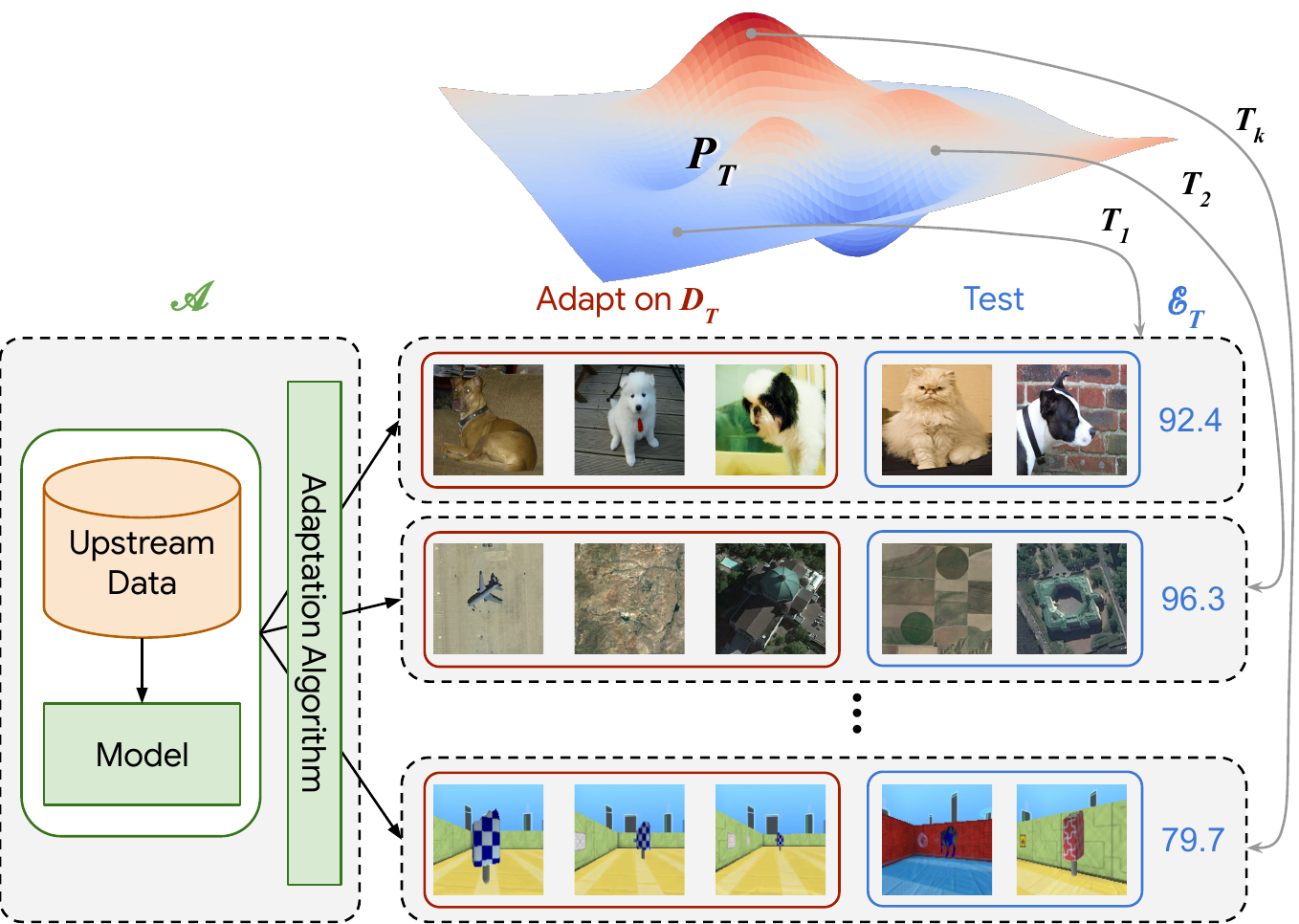}
\caption{Overview of the VTAB evaluation protocol.}
\label{fig:vtab_protocol}
\end{figure}

We seek algorithms that perform well on a wide variety of unseen visual understanding tasks with few labels per task.\footnote{
In some practical settings additional \emph{unlabelled} data may be available for unseen tasks in addition to the labelled data.
We omit this setting, leaving it to future work.}
We first formalize this objective and then specify a practical benchmarking procedure to measure progress.

A \emph{dataset} $D^n$ is a set of $n$ instances $\{(\vx_i,\, \vy_i)\}_{i=1}^n$ with observations  $\vx_i \in \dataspace$ and labels $\vy_i \in \labelspace$.
A \emph{prediction function} is any mapping $F: X \to Y$ (e.g.\ a classifier).
A (learning) \emph{algorithm}, $\mathcal{A}$, takes as input a dataset and outputs a prediction function.
For example, $\mathcal{A}$ may be a pre-trained network coupled with a training mechanism.
An \emph{evaluation procedure}, $\mathcal{E}_T$, takes $F$ and outputs a scalar measuring $F$'s performance (e.g.\ test-set accuracy). 
We seek the algorithm that maximizes the expected performance over a distribution of tasks $P_\t$, where a task $\t$ is a tuple containing a task-specific dataset distribution $D_\t$ and evaluation procedure.
Given only $n$ samples per task we want to maximize:
\begin{align}
\textsc{score}_n(\mathcal{A}) = \mathbb{E}_{\t \sim P_\t}\; \mathcal{E}_\t\left[ \mathcal{A}(D_\t^n) \right],\label{eq:ta}
\end{align}
This general formulation requires a few clarifications.
First, the task distribution needs to be appropriate; we desire a spectrum that covers tasks solvable by a vision algorithm with human-like capabilities.
Second, $n$ may be varied to measure an algorithm's sample complexity.
In practice we choose $n$ to be similar to a modest labelling budget (\cref{sec:setup}).
Third, we assume that $P_\t$ is known, and we aim to build an algorithm with the best inductive biases for solving samples from it.
We desire an ``open world'', where evaluation data is extremely diverse and can always be freshly sampled.
Unfortunately, in practice the benchmark must contain a finite number of test tasks. Therefore, we must ensure that the algorithms are not pre-exposed to specific evaluation samples, as described below.

\subsection{A Practical Benchmark}
We now describe the Visual Task Adaptation Benchmark (VTAB), which is designed to be the best possible proxy for \cref{eq:ta}.
\cref{fig:vtab_protocol} present an overview.

\myparagraph{Task Distribution}
We aim for universal visual understanding, so we informally define $P_\t$ as ``Tasks that a human can solve, from visual input alone.''. 
We validate this empirically, see \cref{app:human}.
Intuitively, such tasks should benefit from visual representations learned by observing and interacting with the natural world. 
The second clause eliminates tasks that require external knowledge that is (currently) unreasonable for a vision algorithm to acquire -- an extreme example would be to classify objects grouped by their spelling in a natural language.
Section~\ref{sec:tasks} details the samples.

\myparagraph{Expectation Over Tasks}
We approximate the expectation over tasks by an empirical average over a number of hand-picked samples.
Ideally, we would sample a new task for each evaluation, as is possible in procedural environments, e.g.~\citep{finn2017}.
However, real-world vision tasks are expensive to collect, so we define a fixed, representative set of samples from $P_{\t}$.
While fixing the set reduces variance, it introduces a risk of meta-overfitting.

\myparagraph{Mitigating Meta-Overfitting}
To reduce meta-overfitting, we treat the evaluation tasks like a test set, and consider them unseen.
Therefore, algorithms that use pre-training must not pre-train on any of the evaluation tasks (even their unlabelled images).
Upstream training should provide useful inductive biases for any draw from $P_{\t}$, so should not use test samples. 
Fortunately, despite numerous test re-evaluations on popular ML benchmarks, such as \imagenet{}, progress seems to transfer to new data~\citep{recht18cifar,recht19imagenet,kornblith2018better}.

\myparagraph{Unified Implementation}
With many diverse tasks, usage could become impractical.
As discussed, the algorithms must have no prior knowledge of the downstream tasks;
while it is permitted to run a hyperparameter search on each task, the search space cannot be task-dependent.
To get meaningful results, we need to define hyperparameter searches that work well across the benchmark.

For this, we convert all tasks into classification problems.
For example, a detection task requiring localization of an object can be mapped to classification of the $(x,y,z)$ coordinates.
With a homogeneous task interface, we may control for possible confounding factors.
For example, we may use the same architecture and hyperparameter sweep everywhere.
Not all tasks can be efficiently modeled as image-level classification, such as those requiring per-pixel predictions.
Nonetheless, we design the tasks such that success on VTAB requires learning of diverse set of visual features:
object identification, scene classification, pathology detection, counting, localization, and 3D geometry.

\subsection{Tasks\label{sec:tasks}}
VTAB contains 19  tasks which cover a broad spectrum of domains and semantics. \cref{app:tasks} contains details. These are grouped into three sets: \taskNatural{}, \taskSpecialized{}, and \taskStructured{}.

The \taskNatural{} group represents classical vision problems.
These tasks contain natural images captured using standard cameras.
The classes may represent generic, fine-grained, or abstract objects.
The group includes: Caltech101, CIFAR-100, DTD, Flowers102, Pets, Sun397, and SVHN.

The \taskSpecialized{} group also contains images of the world, but captured through specialist equipment.
These images have different invariances to those in the \taskNatural{} tasks.
Nonetheless, humans recognize the structures therein, thus generic visual representations should also capture the visual concepts.
We have two sub-groups: remote sensing, and medical.
Remote sensing includes Resisc45 and EuroSAT: aerial images of the earth captured using satellites or aerial photography.
Medical includes Patch Camelyon, metastases detection from microscopy images, and Diabetic Retinopathy, retinopathy classification from fundus images.

The \taskStructured{} group assesses comprehension of the structure of a scene, for example, object counting, or 3D depth prediction.
Most of these tasks are generated from simulated environments, whose structure is easy for a human to determine, but whose domain differs greatly to datasets like \imagenet{}.
These tasks are intended as a step towards useful representations for perceptual control.
We include:
\emph{Clevr}: Simple shapes rendered in a 3D scene, with two tasks: counting and depth prediction.
\emph{dSprites}: Simple black/white shapes rendered in 2D, with two tasks: location and orientation prediction.
\emph{SmallNORB}: Artificial objects viewed under varying conditions, with two tasks: object-azimuth and camera-elevation prediction.
\emph{DMLab}: Frames from a rendered 3D maze. The task involves predicting the time for a pre-trained RL agent to navigate to an object.
\emph{KITTI}: frames captured from a car driver's perspective.
We predict the depth of the nearest vehicle.

\subsection{Representation and Transfer Learning}

Success on VTAB, and ultimately optimizing~\cref{eq:ta}, requires some knowledge of $P_{\t}$.
For example, CNN architectures have a useful inductive bias for vision.
While these biases are usually manually designed (e.g. through architecture choice), they can also be learned.
For example by learning: architectures~\citet{zoph2017}, optimization algorithms~\citet{bello2017neural}, initialization distributions~\citet{raghu2019},
or pre-trained data representations.

Whilst many strategies merit investigation, we focus on representation learning.
Human visual perception is refined through years of observation and interaction with the world, resulting in a system that solves new tasks with few in-domain labels.
Likewise, we aim to pre-train a network that extracts useful features, or representations, from raw data.

\myparagraph{Transfer Strategy}
The representations are not pre-trained on the evaluation tasks themselves (which VTAB forbids), so they must be adapted to solve the new tasks (using limited in-domain data).
The simplest adaptation strategy in deep representation learning is to first pre-train the network, then \emph{freeze} the network's weights and train another -- usually smaller -- model on top.
When the upstream and downstream datasets differ significantly,
\emph{fine-tuning} the original weights is more effective~\citep{yosinski2014,kornblith2018better}.
VTAB does not constrain the transfer strategy; here we use fine-tuning as it tends to perform best.

\myparagraph{Upstream Training}
Representation learning literature often focuses on unsupervised learning which may be applied to any dataset. However, supervised data, where available, can yield good representations. Indeed, the most popular models used in practice are pre-trained on \imagenet{} labels~\citep[and refs therein]{huh2016makes}.
VTAB \emph{does not constrain the type of data} used for pre-training.

\section{Large-Scale Study\label{sec:experiments}}

With VTAB, we evaluate many popular, publicly-available representation learning algorithms across the different types of tasks.
Here, we study the pre-training losses, and therefore, control for other factors such as architecture, preprocessing, transfer hyperparameters, and pre-training data.
Of course, VTAB may be used subsequently to study and improve these aspects as well.
Finally, we provide additional analysis of various of VTAB's design choices and assess other evaluation protocols.

\subsection{Setup\label{sec:setup}}

\myparagraph{Downstream Data Size}
Section~\ref{sec:tasks} describes the tasks.
VTAB aims to assess adaptation with limited data, so we evaluate primarily using \num{1000} labelled examples per task (called VTAB-1k).
We define train (\num{800} samples) and validation (\num{200} samples) sets for users' convenience.
We emphasize that to avoid meta-overfitting, one should not use extra images for hyperparameter selection, but may use the \num{1000} examples in any manner.
We also use the full datasets (VTAB-full).
This allows us to assess the value of representation learning as in-domain data increases, and to check performance against prior art.

\myparagraph{Representation Learning Algorithms}
We evaluate supervised, semi-supervised, self-supervised, and generative models, selecting popular, public methods from each class.
We also compare to ``from-scratch'' models, which use no pre-training, yielding a total of 18 algorithms.
All models (except ``from-scratch'') are pre-trained on \imagenet{} (1.28M labelled images).
Some use the \imagenet{} labels, others use some or none of the labels.

We train two supervised models: one using $100\%$ of the \imagenet{} labels (\textsc{Sup-100\%}), and one using $10\%$ (\textsc{Sup-10\%}).
For self-supervised learning, we include both image-based and patch-based models.
The image-based models include \textsc{rotation}~\citep{gidaris2018unsupervised} and \textsc{Exemplar}~\citep{dosovitskiy2014exemplar}.
Patch-based include
\textsc{Relative Patch Location}~\citep{doersch2015unsupervised} and \textsc{Jigsaw}~\citep{noroozi2016unsupervised}.
We use the public implementations of~\citet{kolesnikov2019revisiting}.
The patch-based models are converted into image-level classifiers by averaging the representations from a $3\times3$ grid of patches (see \cref{app:architectures}).
The semi-supervised models are trained using 10\% of the \imagenet{} labels with an auxiliary loss on all of the data, see~\citep{zhai2019s4l}.
We use either rotation (\textsc{Sup-Rotation-10\%}) or Exemplar (\textsc{Sup-Exemplar-10\%}) auxiliary losses.
These models can also use all of the labels, denoted \textsc{Sup-Rotation-100\%} and \textsc{Sup-Exemplar-100\%}.
For the generative models, we evaluate GANs and VAEs.
As is common, we take the GAN's image representations from the discriminator, replacing the final linear layer with a classification layer.
We use both the label-conditional and unconditional BigGAN discriminators~\citep{brock2018large} (\textsc{Cond-BigGAN} and \textsc{Uncond-BigGAN}) using the implementations of~\citet{chen2019self,lucic2019high}.
We also evaluate the encoder from \textsc{BigBiGAN}~\citep{donahue2019large}.
For the autoencoders, we use the encoder as the representation.
We evaluate VAEs~\citep{kingma2013auto}, and WAEs with three distribution matching losses: GAN, MMD~\citep{tolstikhin2017wasserstein}, and UKL~\citep{rubenstein2019practical}.

\myparagraph{Transfer Algorithm}
VTAB permits any method of transfer.
Here, we fine-tuning the entire network on the task data, since that performs best~\citep{kornblith2018better}.
It is also popular to add a linear model to a frozen network, so we compare to this approach in \cref{sec:linear}.

\myparagraph{Hyperparameters}
Upstream, we contol the data (\imagenet{}) and architecture.
We find that bigger architectures perform better on VTAB (\cref{app:scale-architecture} and~\citet{kolesnikov2019revisiting}).
For comparability, we use ResNet50-v2, or similar, in this study.
For the supervised and semi-supervised methods we use the architecture in~\citet{he2016identity}.
For GANs and auto-encoders we use very similar deep ResNets, but with appropriate modifications to pre-train them successfully as a generative model.
\cref{app:architectures} contains details.

Downstream, the transfer hyperparameters influence performance, and different datasets require different settings.
We run VTAB in two modes: \emph{lightweight} and \emph{heavyweight}.
Lightweight mode performs a restricted per-task hyperparameter search.
Many hyperparameters are fixed, including optimizer, batch size, pre-processing, and weight decay.
For each task, the lightweight mode sweeps: 2 initial learning rates, and 2 learning rate schedules. See \cref{app:hyperparameter_sweep} for details.
We choose short schedules to limit cost, but show in \cref{sec:analysis} that these yield near-optimal performance.

In heavyweight mode we perform a large random search over learning rate, schedule, optimizers, batch size, train pre-processing functions, evaluation pre-processing, and weight decay.
We include longer training schedules and higher resolution images.
This mode is used to understand the impact of a larger computational budget on the performance, and establish performance upper bounds on the methods used.
\cref{app:hyperparameter_sweep} contains details.

We perform our main study in lightweight mode.
In \cref{sec:heavy} we show that although extensive tuning improves all methods, the relative performances are mostly unaffected.
For future study using VTAB, we recommend defining a similar lightweight setup (or using the one here) that facilitates fair comparison without high cost.
If ample computational resources are available, a heavyweight mode may be used to improve the state-of-the-art.

\myparagraph{Tuning and Evaluation Protocol}
We perform adaptation on the training set and perform model selection for each task on the pre-defined validation sets.
We re-train the best model on the union of training and validation sets and evaluate on the test set.
Note, for VTAB-1k we define custom train set with 800 examples and validation set with 200 examples.
We run the final evaluation on the test set with three random seeds and report the median score.

\myparagraph{Metrics}
We evaluate with top-1 accuracy.
We consider other metrics in \cref{app:metrics}, and find the conclusions are the same.
To aggregate scores across tasks, we take the mean accuracy.
We investigate more complex aggregation strategies in Section~\ref{sec:analysis}, but the relative performances are unaffected, so we use mean accuracy for simplicity.

\subsection{VTAB Results}

\begin{SCfigure*}[][t]
  \includegraphics[width=1.5\linewidth]{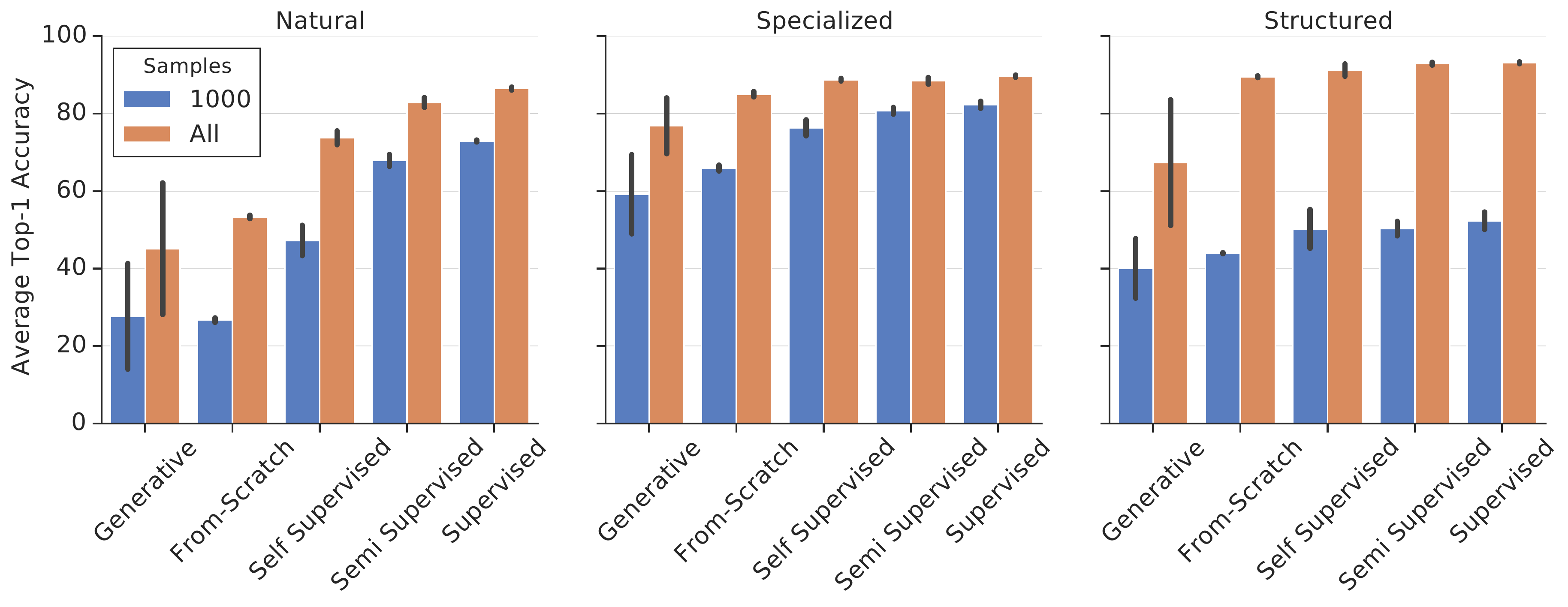}\\
  \vspace{-15pt}
  \caption{
  The methods are divided into five groups:
  Generative,
  training from-scratch,
  all methods using $10\%$ labels (Semi Supervised),
  and all methods using $100\%$ labels (Supervised),
  For each task group, bars show the mean accuracy across all methods with 3 repeats.
  Error bars indicate standard deviation across methods.
  The BigBiGAN outlier causes the high variance in generative models.
  \vspace{-15pt}
  }
  \label{fig:all-methods-grouped}
\end{SCfigure*}

\begin{SCfigure*}[][t]
  \centering
  \begin{tabular}{cc}
  \includegraphics[width=0.8\linewidth]{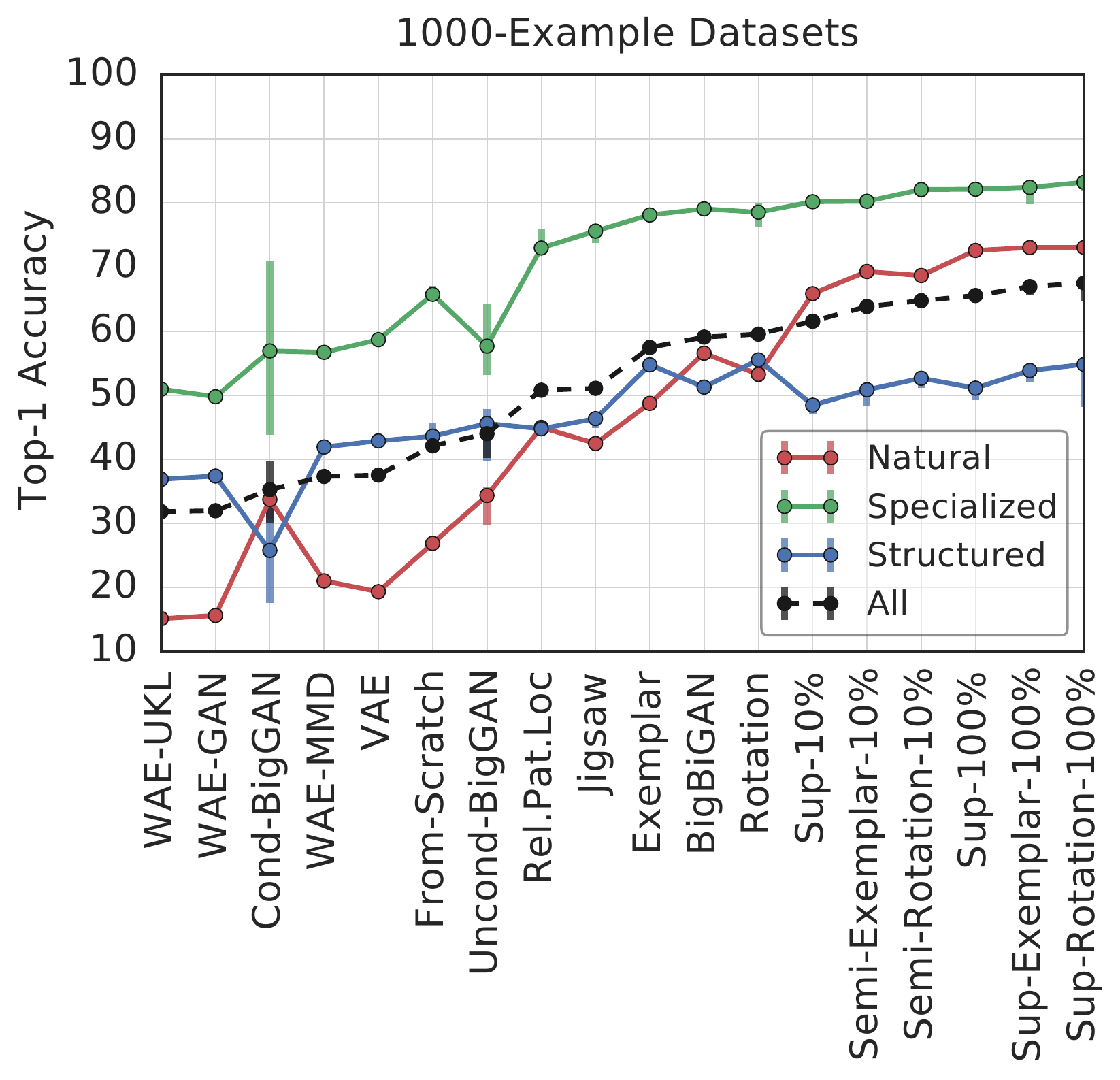}
  \includegraphics[width=0.8\linewidth]{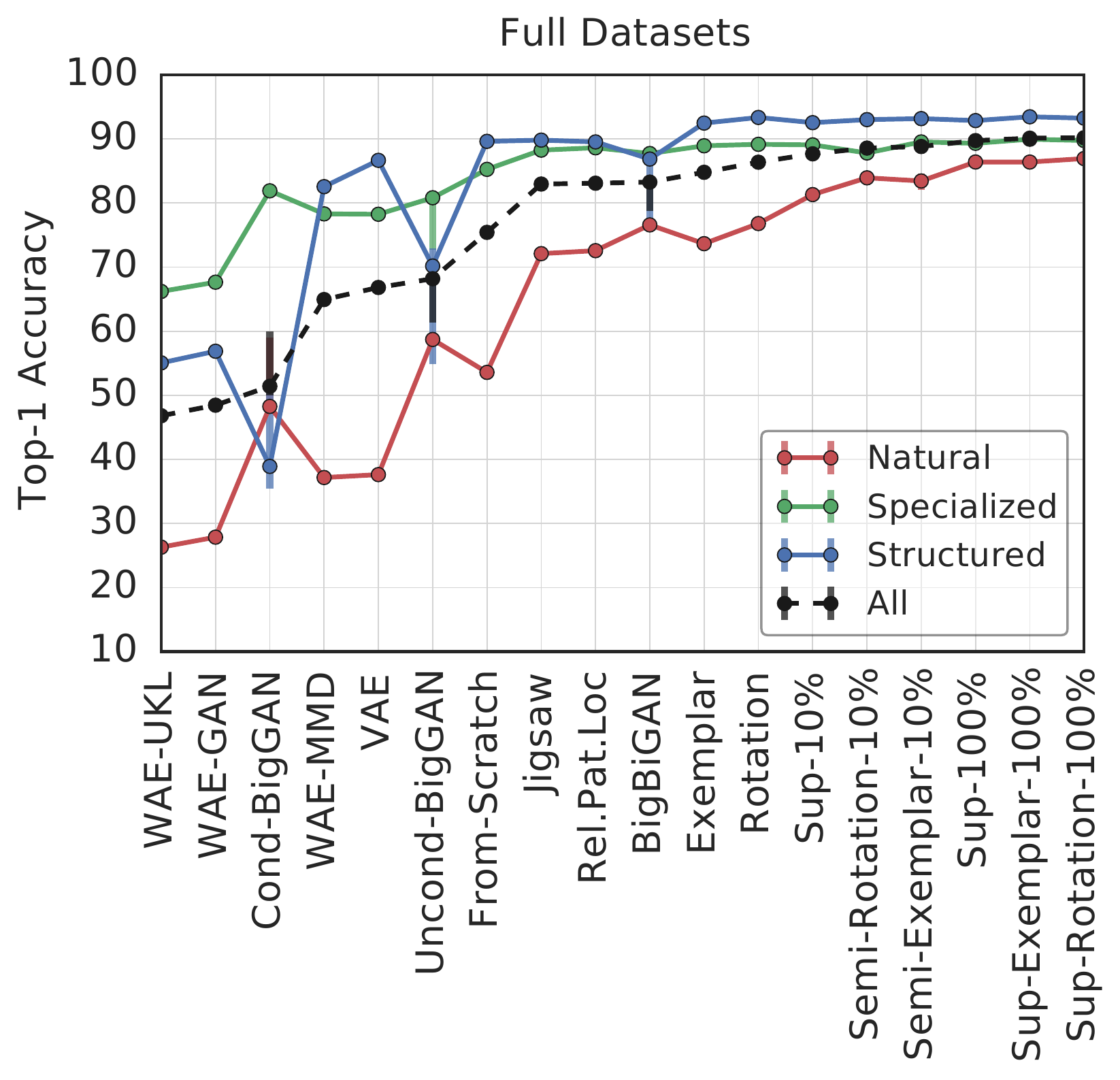}
  \end{tabular}
  \caption{
  Average top-1 accuracy across the tasks in each group for VTAB-1k.
  The x-axis indexes the methods, ordered ordered according to their average accuracy across all tasks (dashed curve).
  Error bars indicate 95\% confidence interval using bootstrap resampling from the 3 experiment repeats.
  }
  \label{fig:all-methods}
\end{SCfigure*}

\begin{figure}[t]
\centering
\begin{tabular}{c}
    \includegraphics[width=0.9\linewidth]{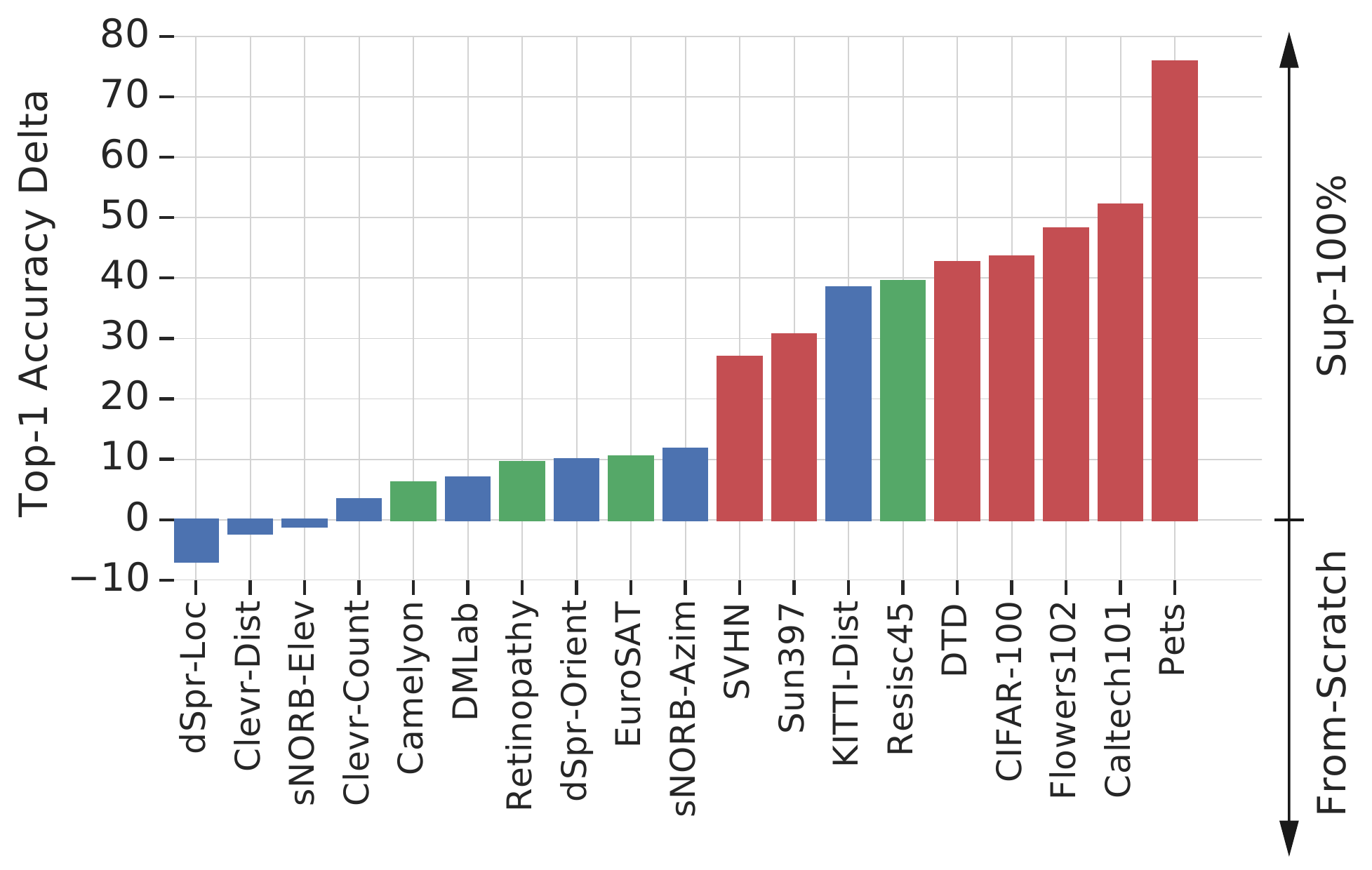}\\
    \includegraphics[width=0.9\linewidth]{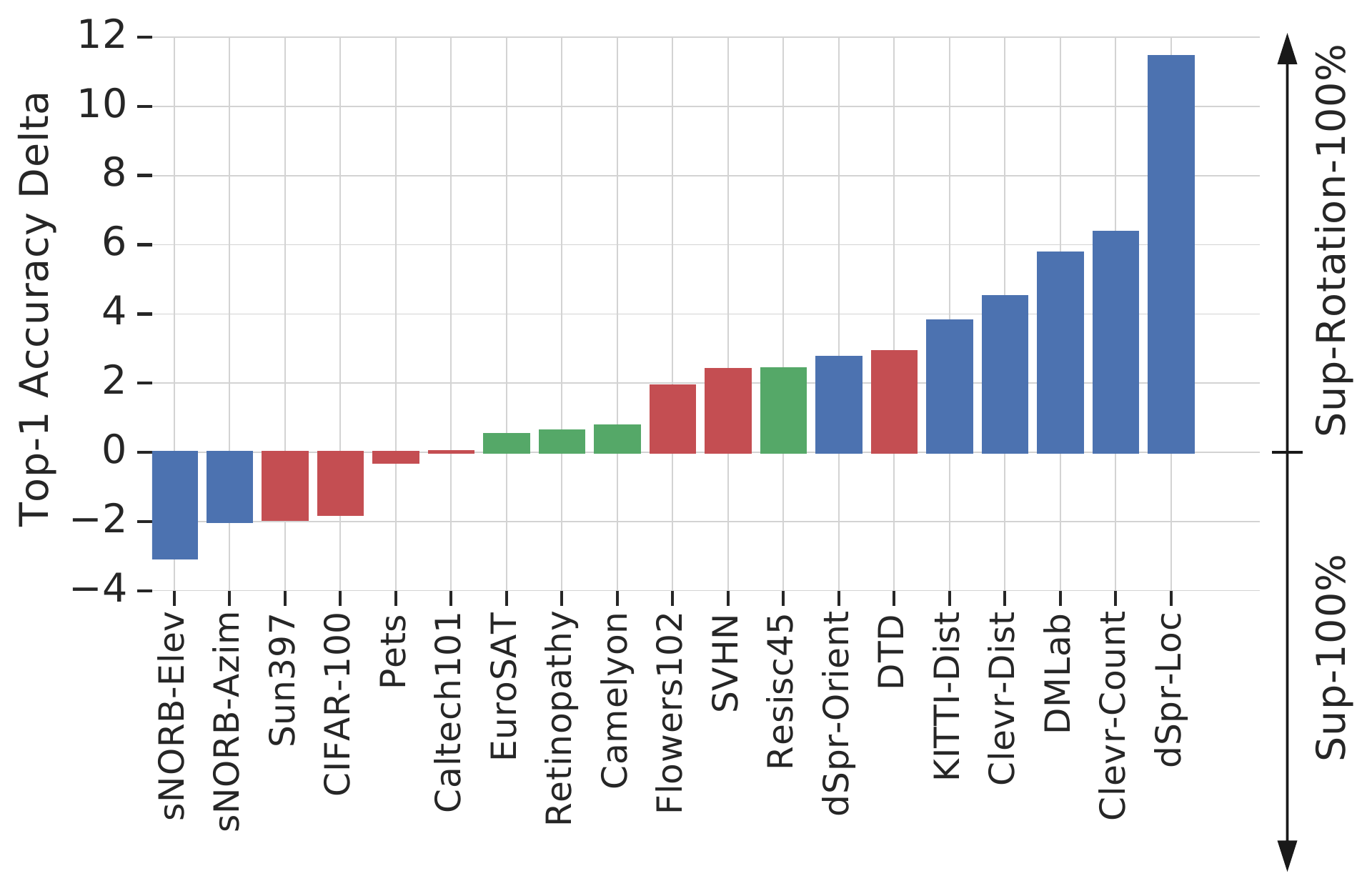}
\end{tabular}
\caption{
    Absolute difference in top-1 accuracy between method pairs for each dataset, using 1k examples.
    The bar colour indicates the task group: \taskNatural{}, \taskSpecialized{}, and \taskStructured{}.
    Left: \textsc{Sup-100\%} versus \textsc{From-Scratch} --- supervised pre-training yields a large improvement on the \taskNatural{} datasets and some others.
    Right: \textsc{Sup-Rotation-100\%} versus \textsc{Sup-100\%} --- the additional self-supervised loss on top of supervised loss yields improvements, especially on the \taskStructured{} tasks.
   Note that the y-scales differ.
}
\label{fig:atari}
\vspace{-15pt}
\end{figure}

We run all 18 methods on 19 tasks for both VTAB-1k and VTAB-full using the lightweight sweep.
\cref{fig:all-methods-grouped} shows aggregate performance of each method group (generative, self-sup, etc.) on each VTAB task group (\taskNatural{}, etc.).
\cref{fig:all-methods} presents a breakdown of the performance of each algorithm on each taks group.
\cref{fig:atari} shows two detailed per-task comparisons:
\textsc{Sup-100\%} (\imagenet{}) versus \textsc{From-Scratch}, and
\textsc{Sup-Rotation-100\%} (best overall on VTAB-1k) versus \textsc{Sup-100\%}.
\cref{app:lightweight},~\cref{tab:lightweight} contains the complete results table.

\myparagraph{Generative Models}
Overall, these perform worst, \cref{fig:all-methods-grouped}.
The autoencoders perform more poorly than training from scratch.
Despite SOTA generative quality on \imagenet{}, the GANs' discriminators do not provide useful representations.
The GAN discriminators appear unstable, with large error bars in \cref{fig:all-methods}.
The \textsc{BigBiGAN} encoder stands out, performing much better -- similar to the best self-supervised models, \cref{fig:all-methods}.
Interestingly, \textsc{BigBiGAN} performs relatively well on natural tasks, but less so on structured tasks.
\textsc{Uncond-GAN} shows a similar pattern.
This indicates that GANs fit more strongly to \imagenet's domain (natural images), than self-supervised alternatives.
Despite huge advances in generative quality, these models do not yield great representation quality.
Similarly,~\citet{ravuri2019classification}~show that models with high sample quality fail to generate data from which an accurate classifier can be trained.
However, the BigBiGAN result holds promise for adversarially-trained encoders, and consideration of downstream representation quality may lead to further progress.

\myparagraph{Self-supervised}
All self-supervised representations outperform from-scratch training.
The best, \textsc{Rotation}, attains $59.6\%$ on VTAB-1k, while \textsc{From-Scratch} attains $42.1\%$ (\cref{app:lightweight}, \cref{tab:lightweight}).
Methods applied to the entire image (\textsc{Rotation}, \textsc{Exemplar}) outperform patch-based methods (\textsc{Jigsaw}, \textsc{Rel.Pat.Loc.}).
However, the patch based methods perform slightly better on DTD (Describable Textures) and Retinopathy (Fundus images), see \cref{app:lightweight}, \cref{tab:lightweight}.
Intuitively, these tasks require sensitivity to local textures, which are indeed captured by patch-based methods.
On \taskNatural{} tasks, self-supervision is far behind supervised methods (\cref{fig:all-methods-grouped}).
On \taskSpecialized{}, the performances are similar, and most interestingly, on \taskStructured{} self-supervised methods performs slightly \emph{better} than supervised.
This indicates supervised \imagenet{} models are invariant to useful features required for structured understanding, but self-supervised methods can capture these to some degree.

\myparagraph{(Semi-)Supervised}
Overall, supervised models perform best.
The benefits are most pronounced on the \taskNatural{} tasks, whose domain and semantics are arguably most similar \imagenet{} classification.
However, with self-supervision, the benefit can be attained with fewer labels.
\textsc{Sup-10\%} (120k labels) attains $61.6\%$ on VTAB-1k , while \textsc{Sup-100\%} (1.2M labels) attains $65.6\%$.
With self-supervision, \textsc{Sup-Rotation-10\%} closes $80\%$ of the gap, attaining $64.8\%$.
Recent work has shown in-domain benefits of semi-supervised learning~\citep{henaff2019data,berthelot2019remixmatch}; our study reveals that semi-supervised methods also learn good representations for transfer to new tasks.
Finally, additional self-supervision even improves on top of $100\%$ labelled \imagenet{}, particularly on \taskStructured{} tasks (\cref{fig:atari}.
\textsc{Sup-100\%} attains $65.6\%$, whereas \textsc{Sup-Rotation-100\%} attains $67.5\%$.

\subsection{Heavyweight Hyperparameter Sweeps \label{sec:heavy}}

\begin{table*}[ht]
\vspace{-20pt}
\caption{Top-1 accuracy of the best models and ResNet50 from-scratch using the heavyweight hyperparameter sweep.}
\fontsize{7pt}{7pt}\selectfont
\newcolumntype{C}{>{\centering\arraybackslash}X}
\setlength{\tabcolsep}{0pt}
\setlength{\extrarowheight}{5pt}
\renewcommand{\arraystretch}{0.75}
\begin{tabularx}{\linewidth}{p{10pt}p{1.6cm}!{\color{lightgray}\vline} CCCCCCC!{\color{lightgray}\vline}CCCC!{\color{lightgray}\vline}CCCCCCCC!{\color{lightgray}\vline}C}
\toprule
 &
 & \rotatebox{90}{\raisebox{0.5pt}{\tikz\fill[natural] (0,0) circle (.5ex);} Caltech101}
 & \rotatebox{90}{\raisebox{0.5pt}{\tikz\fill[natural] (0,0) circle (.5ex);} CIFAR-100}
 & \rotatebox{90}{\raisebox{0.5pt}{\tikz\fill[natural] (0,0) circle (.5ex);} DTD}
 & \rotatebox{90}{\raisebox{0.5pt}{\tikz\fill[natural] (0,0) circle (.5ex);} Flowers102}
 & \rotatebox{90}{\raisebox{0.5pt}{\tikz\fill[natural] (0,0) circle (.5ex);} Pets}
 & \rotatebox{90}{\raisebox{0.5pt}{\tikz\fill[natural] (0,0) circle (.5ex);} Sun397}
 & \rotatebox{90}{\raisebox{0.5pt}{\tikz\fill[natural] (0,0) circle (.5ex);} SVHN}
 & \rotatebox{90}{\raisebox{0.5pt}{\tikz\fill[specialized] (0,0) circle (.5ex);} Camelyon}
 & \rotatebox{90}{\raisebox{0.5pt}{\tikz\fill[specialized] (0,0) circle (.5ex);} EuroSAT}
 & \rotatebox{90}{\raisebox{0.5pt}{\tikz\fill[specialized] (0,0) circle (.5ex);} Resisc45}
 & \rotatebox{90}{\raisebox{0.5pt}{\tikz\fill[specialized] (0,0) circle (.5ex);} Retinopathy}
 & \rotatebox{90}{\raisebox{0.5pt}{\tikz\fill[structured] (0,0) circle (.5ex);} Clevr-Count}
 & \rotatebox{90}{\raisebox{0.5pt}{\tikz\fill[structured] (0,0) circle (.5ex);} Clevr-Dist}
 & \rotatebox{90}{\raisebox{0.5pt}{\tikz\fill[structured] (0,0) circle (.5ex);} DMLab}
 & \rotatebox{90}{\raisebox{0.5pt}{\tikz\fill[structured] (0,0) circle (.5ex);} dSpr-Loc}
 & \rotatebox{90}{\raisebox{0.5pt}{\tikz\fill[structured] (0,0) circle (.5ex);} dSpr-Ori}
 & \rotatebox{90}{\raisebox{0.5pt}{\tikz\fill[structured] (0,0) circle (.5ex);} KITTI-Dist}
 & \rotatebox{90}{\raisebox{0.5pt}{\tikz\fill[structured] (0,0) circle (.5ex);} sNORB-Azim}
 & \rotatebox{90}{\raisebox{0.5pt}{\tikz\fill[structured] (0,0) circle (.5ex);} sNORB-Elev}
 & \rotatebox{90}{\raisebox{0.5pt}{\tikz\fill[all] (0,0) circle (.5ex);} Mean} \\
\midrule

\multirow{4}{*}{\rotatebox{90}{\hspace*{-2pt}1000}}
& From-Scratch &       61.2 &      20.4 &     46.4 &       73.4 &     48.0 &     13.1 &     82.5 &     77.0 &     91.5 &     59.5 &        73.4 &        66.0 &   \bf 59.2 &     36.9 &     88.2 &        62.6 &       63.8 &       22.6 &   \bf 79.1 &     59.2 \\
      & Sup-100\% &   \bf 83.6 &      48.9 &     65.9 &       93.0 & \bf 90.4 &     31.1 &     87.0 &     80.2 &     95.8 &     82.0 &    \bf 79.0 &        78.8 &       58.3 &     53.8 &     89.0 &    \bf 71.2 &   \bf 73.9 &       34.2 &       57.8 &     71.2 \\
      & Sup-Rot-100\% &       80.0 &      42.6 & \bf 66.5 &       91.5 &     88.6 &     32.7 & \bf 90.0 &     78.5 & \bf 96.4 & \bf 82.1 &        78.4 &        96.2 &       58.1 &     55.3 &     89.3 &        71.1 &       69.1 &   \bf 39.3 &       53.6 &     71.5 \\
      & Sup-Ex-100\% &       83.3 &  \bf 49.3 &     61.8 &   \bf 94.2 &     90.2 & \bf 35.2 &     87.6 & \bf 84.2 &     95.7 &     81.0 &        78.7 &    \bf 96.4 &       56.3 & \bf 56.1 & \bf 90.0 &        69.8 &       73.3 &       38.6 &       59.5 & \bf 72.7 \\

\arrayrulecolor{lightgray}\specialrule{.5pt}{0.6pt}{-0.5pt}\arrayrulecolor{black}

\multirow{4}{*}{\rotatebox{90}{\hspace*{-2pt}Full}}
& From-Scratch &       74.5 &      77.8 &     67.1 &       85.8 &     70.9 &     70.1 &     97.0 & \bf 91.2 &     98.8 &     94.3 &        82.8 &        99.8 &       96.7 &     76.6 & \bf 100.0 &        96.7 &       68.4 &       99.9 &       94.0 &     86.4 \\
      & Sup-100\% &       93.0 &      83.4 &     73.7 &       97.3 &     92.7 &     75.6 &     97.5 &     87.3 &     98.8 &     96.1 &        83.4 &       100.0 &   \bf 97.0 &     78.8 & \bf 100.0 &    \bf 96.8 &       81.0 &  \bf 100.0 &       98.5 &     91.1 \\
      & Sup-Rot-100\% &       93.5 &  \bf 84.0 & \bf 76.8 &       97.4 &     92.6 & \bf 75.6 &     97.4 &     86.5 & \bf 99.1 & \bf 96.3 &    \bf 83.7 &   \bf 100.0 &       96.8 & \bf 79.6 & \bf 100.0 &    \bf 96.8 &       82.4 &  \bf 100.0 &       96.7 &     91.3 \\
      & Sup-Ex-100\% &   \bf 93.8 &      83.1 &     76.5 &   \bf 97.8 & \bf 92.9 &     75.3 & \bf 97.5 &     86.5 &     99.0 &     96.3 &        83.7 &        99.9 &       96.8 &     79.3 & \bf 100.0 &        96.7 &   \bf 82.8 &  \bf 100.0 &   \bf 99.1 & \bf 91.4 \\

\bottomrule
\end{tabularx}

\label{tab:heavyweight_full_test_selection}
\vspace{-10pt}
\end{table*}

We evaluate \textsc{From-Scratch} and the best models: \textsc{Sup-100\%}, \textsc{Sup-Rotation-100\%}, and \textsc{Sup-Exemplar-100\%} using the heavyweight hyperparameter search:~\cref{tab:heavyweight_full_test_selection}.
\cref{app:hyperparameter_sweep} shows the selected hyperparameters.
As expected, all methods improve significantly.
Prior work~\citep{he2018rethinking} shows that with sufficient data and training time, from-scratch training is competitive for detection.
However, we observe that across all task groups, pre-trained representations are better than a tuned from-scratch model.
On a couple of tasks \textsc{From-Scratch} performs competitively --- Clevr-Dist and sNORB-Elev, which require localization and camera elevation respectively --- indicating that even the best \imagenet{} representations fail to capture these aspects.
The ranking of the best methods is similar, but not identical, with a combination supervision and self-supervision (\textsc{Sup-Exemplar-100\%}) getting the best performance.

We check our results against those recently reported in the literature (\cref{app:heavyweight}, \cref{tab:literature}).
Our results are comparable; behind on highly popular tasks on which complex architectures have been optimized (e.g. CIFAR), but ahead in others.
In \cref{app:scale-architecture} we show that simply increasing the architecture size improves VTAB performance.

\subsection{Frozen Features Extractors\label{sec:linear}}

\begin{figure}
\centering
\includegraphics[width=0.9\linewidth]{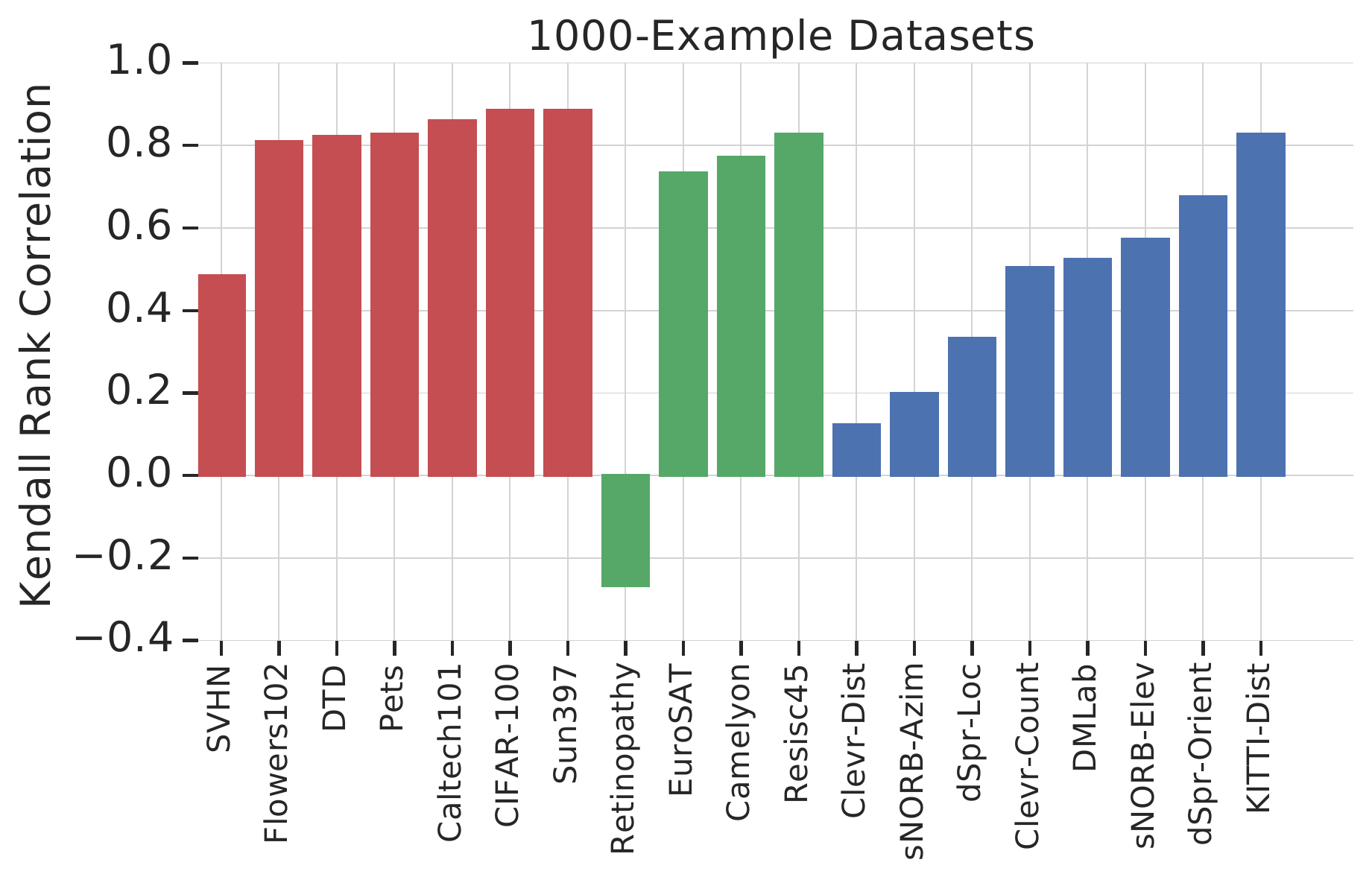}
\vspace{-5pt}
\caption{Kendall's correlation between fine-tuning and linear evaluation on each dataset.}
\label{fig:finetune_vs_linear}
\end{figure}

Representations are often evaluated by training a linear layer on a frozen model~\citep{kolesnikov2019revisiting}.
\imagenet{} is often used for linear evaluation, but this is meaningless for the models we consider because many use \imagenet{} labels for pre-training.
However, linear evaluation is also used in a transfer setting~\citep{goyal2019scaling}, so we apply this protocol to the VTAB tasks, and contrast it to fine-tuning.
The full protocol resembles the lightweight fine-tuning sweep, see \cref{app:linear}.

\cref{fig:finetune_vs_linear} shows the per-task correlation between linear and fine-tuning.
\cref{app:linear}, \cref{tab:lightweight_linear_test} contains the full table of results.
We first note that linear evaluation significantly lowers performance, even when downstream data is limited to $1000$ examples.
\textsc{Sup-100\%} attains $65.6\%$ with fine-tuning (lightweight sweep), but $57.3\%$ with linear.
Linear transfer would not by used in practice unless infrastructural constraints required it.
Second, \cref{fig:finetune_vs_linear} shows that on many datasets, particularly \taskSpecialized{} and \taskStructured{}, the correlation is low.
Linear evaluation may lead to different conclusions, for example, \textsc{Cond-BigGAN} attains $43.3\%$ (1000-examples) on linear, outperforming both \textsc{Rel.Pat.Loc} and \textsc{Jigsaw}.
Yet when fine-tuned these self-supervised methods significantly outperform the GAN discriminator.
Another discrepancy is between semi-supervised and supervised.
\textsc{Semi-Rotation-10\%} and \textsc{Semi-Exemplar-10\%} are $1-2\%$ behind \textsc{Sup-100\%} with fine-tuning, but $4-5\%$ behind with linear evaluation.
These self-supervised methods extract useful representations, just without linear separability.
Previous works~\citep{kornblith2018better,kolesnikov2019revisiting} claim that linear evaluation results are sensitive to additional factors that we do not vary, such as ResNet version or pre-training regularization parameters.
Overall linear evaluation is a poor proxy for overall reduced sample complexity.

\subsection{Analysis\label{sec:analysis}}

\myparagraph{Metrics}
We explore alternatives to using top-1 accuracy, and aggregation of scores accross tasks using the mean.
Instead of top-1 accuracy, we used mean-per-class accuracy and Cohen's quadratic kappa.
These metrics create only minor ranking differences, leaving the overall picture unchanged.
Kendall's ranking correlation with respect to top-1 accuracy is always $>0.97$.
Details are in~\ref{app:metrics}.
For aggregation across tasks, we consider seven alternative strategies:
(i)~Macro-average across datasets (grouping tasks with the same input data).
(ii)~Macro-average across groups (merging tasks in the same group).
(iii)~Geometric mean.
(iv)~Average rank.
(v)~Average rank after small perturbation of the scores with noise.
(vi)~Robust (binned) average rank.
(vii)~Elimination rank -- equivalent to an ``Exhaustive Ballot''.
All strategies have a high Kendall's correlation with the vanilla mean across tasks ($\tau>0.87$).
Appendix~\ref{app:alt-model-ranking} contains details.
Based on these results we choose mean top-1 for VTAB since it is simple, interpretable, can be computed for each task independently, and is highly correlated with more complex appraoches.

\myparagraph{Representative Subset}
For prototyping, a representative subset of VTAB may be useful.
We compute the rank correlation between the mean scores produced by each $\binom{20}{5}$  subsets of five tasks, and the full suite.
The top-5 subsets tend to span different domains, but differ to each other.
Appendix~\ref{app:subset-model-ranking} contains the results.
Although a subset might be useful for screening models, these were computed on our set of models, and may correlate less well in other experiments.
Using subsets also increases the risk of meta-overfitting, which VTAB aims to avoid by having many tasks.

\begin{figure}
  \centering
  \includegraphics[width=0.7\linewidth]{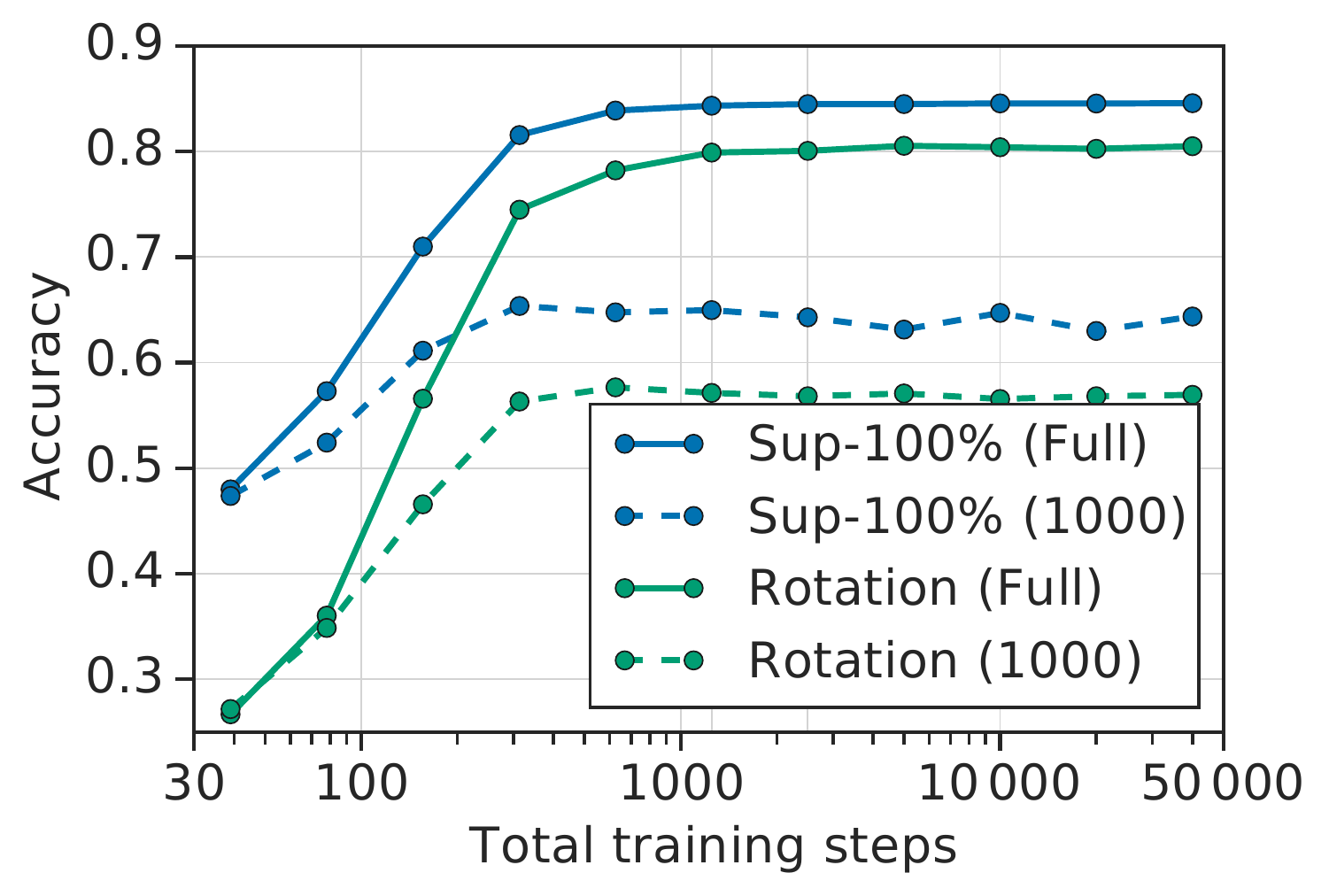}
  \vspace{-10pt}
  \caption{Performance under various time constraints.}
  \label{fig:budget-plot-mean}
  \vspace{-15pt}
\end{figure}

\myparagraph{Limiting Evaluation Cost}
The cost is near linear in the schedule length.
To determine a minimal, yet meaningful, schedule, we sweep over schedules ranging from $40$ to \num{40000} steps, using batch size $512$.
\cref{fig:budget-plot-mean} summarizes the results, details Figs.~\ref{fig:budget-plot-1k} and \ref{fig:budget-plot-full},~\cref{app:budget}.
Most runs reach their optimal performance within \num{1000} steps and do not improve significantly when trained for longer.

Finetuning and evaluating a given ResNet-50 model on a single Nvidia \texttt{P100} GPU (batch size of 64 images, 0.01 initial learning rate) takes 3 hours for VTAB-1k.
To reduce the time and cost we use Google Cloud TPU-v3-16 accelerators and verify that we obtain similar resuts in this setting.

\section{Related Work \label{sec:related_work}}

\myparagraph{Vision Benchmarks}
The Visual Decathlon~\citep{rebuffi2017} contains ten classification tasks: 
Omniglot, and nine using natural images. 
Four overlap with VTAB, but VTAB includes several other domains and tasks.
Importantly, these benchmarks have opposite evaluation protocols:
The Visual Decathlon allows direct joint optimization on the tasks, but forbids external data. 
By contrast, VTAB forbids multi-task learning on the downstream tasks, but permits transfer from any arbitrary dataset --- this simulates one-off training of representations which are then applied to novel tasks.
Further, VTAB considers a low-sample regime (1000-examples), which reflects performance under a reasonable labelling budget.
We conduct experiments showing that the methods ranked according to  VTAB are more likely to transfer to new tasks, than those ranked according to the Visual Decathlon (\cref{sec:visual_decathlon}).

The Facebook AI SSL challenge~\citep{fai2019} was proposed to evaluate self-supervision.
It contains four tasks on two datasets of natural images, including classification, detection, and low-label classification.
The original paper~\citep{goyal2019scaling} also includes navigation and surface normal estimation.
VTAB uses classification tasks only to admit task-independent implementations,
and attains diversity with many alternative domains and semantics (localization, counting, etc.).
Our study includes self-supervision, but are not restricted to it.
Indeed, one challenge is to outperform supervised \imagenet{} on VTAB.
Most importantly, Facebook SSL requires transfer via a shallow network stacked on a 
fixed CNN, whereas VTAB permits any transfer, and focuses on performance with few samples.

Meta-Dataset~\citep{triantafillou2019meta} contains natural image
classification tasks. 
The protocol differs to VTAB: most of Meta-Dataset's evaluation datasets are also in the training set, and the train/test split is made across classes.
Meta-Dataset is designed for few-shot learning, rather than 1000 examples, which may entail different solutions.

\myparagraph{Representation Learning Evaluation}
Popular evaluations for representation learning are linear/MLP and semi-supervised. 
In linear/MLP evaluation, widely used in for self-supervised representations~\citep{doersch2015unsupervised,zhang2016colorful,noroozi2016unsupervised,doersch2017multi}, the weights of a pre-trained network are frozen, and a linear layer/MLP is trained on top to predict the labels. 
Evaluation is often performed on the same dataset that was used to train the representations, but using all labels, defeating the goal of sample efficiency.
A semi-supervised protocol is more realistic and performs better~\citep{zhai2019s4l}. 
This evaluation is performed on a single dataset, by contrast VTAB concerns transfer to \emph{unseen} tasks.
Linear evaluation is sensitive to small upstream training details, such as the ResNet version or regularization, that make little difference when training end-to-end~\citep{kornblith2018better,kolesnikov2019revisiting}.
Other intrinsic scores have been proposed to measure the disentanglement of representations~\citep{locatello2018challenging}, or the mutual information between the input and representations~\citep{hjelm2018learning}.
Both were shown to be weakly correlated with representation utility~\citep{locatello2018challenging,tschannen2019mutual}.

\myparagraph{Generative Evaluation}
Evaluation of generative quality is a difficult goal with numerous proxy metrics, such as reconstruction error, FID~\citep{heusel2017gans}, IS~\citep{salimans2016improved}, precision-recall~\citep{sajjadi2018assessing}, log likelihood (for auto-regressive models), or the ELBO (for VAEs).
Sometimes generative models are also evaluated using linear classification~\citep{radford2016,donahue2016adversarial,dumoulin2016adversarially}, or semi-supervised learning~\citep{kingma2014semi, narayanaswamy2017learning,tschannen2018recent}.

\section{Discussion}
Our study answers important and timely questions on representation learning.
First, how effective are supervised \imagenet{} representations?
\imagenet{} labels are indeed effective for natural tasks.
Further, although the literature reports mixed results~\citep{liu2017detecting,raghu2019,neumann2019domain}, we find that they also work for specialized tasks despite the domain-shift (medical or remote sensing).
However, for structured understanding, supervised representations can be poorer than unsupervised ones.
As future work, training on diverse data sources, such as video, may enable moving beyond the ``\imagenet{}-like'' tasks.

Second, how do representations trained via generative and discriminative models compare?
Generation is a useful task in of itself, but generative losses, in their current form, seem less promising as means towards learning how to represent data.
BigBiGAN is a notable recent exception, that is on par with self-supervision and warrants more exploration.

Third, to what extent can self-supervision replace labels?
Self-supervised learning appears promising, even outperforming supervision on some tasks that require structured understanding.
Interestingly, self-supervision can almost (but not quite) replace $90\%$ of \imagenet{} labels; the gap between pre-training on $10\%$ labels with self-supervison, and $100\%$ labels, is small.
Further, self-supervision adds value on top of \imagenet{} labels \emph{on the same data}.
Overall, it seems we are quite far from general visual representations;
simply adding more data on the \taskSpecialized{} and \taskStructured{} tasks is better than the pre-training strategies we evaluated.
However, self-supervison can be applied to any dataset of images or videos, so combining large scale open-domain self-supervision with \imagenet{} or other label sources may be promising future work.

VTAB measures representation quality as the ability to adapt with few examples to diverse, unseen tasks.
Here, we provide a large study of upstream training losses and control other factors such as
hyperparameter sweep, architecture, transfer algorithm,
preprocessing, and pre-training data.
However, VTAB may be used to analyze and optimze many factors involved in learning generalizable representations.
Varying other factors to improve VTAB is valuable future research.
To isolate the effect of each factor, confounders such as hyperparameter sweep size should be controlled.
The only approach that is out-of-bounds is to condition the algorithm explicitly on the VTAB tasks, since this would compromise representation generalization.

Code and data to run VTAB are made available, and progress is monitored at~\url{github.com/google-research/task_adaptation}.
The models presented here are released.
We hope that VTAB drives representation learning towards making deep learning accessible to the many problems without vast labelling budgets.

\subsubsection*{Acknowledgments}
We thank Raphael Marinier and Anton Raichuk for help building the DMLab classification task,
and Samy Bengio and Kevin Murphy for useful discussions.

\bibliography{task_adapt}
\bibliographystyle{icml2020}

\appendix
\clearpage
\onecolumn

\begin{center}
{\Large\bf Supplementary Material: A Large-scale Study of Representation Learning with the Visual Task Adaptation Benchmark}
\end{center}

\section{Tasks \label{app:tasks}}

\begin{table}[h]
\centering
\begin{tabular}{llrrl}
\toprule
\bf Category & \bf Dataset & \bf Train size & \bf Classes & \bf Reference \\
\toprule
\raisebox{0.5pt}{\tikz\fill[natural] (0,0) circle (.5ex);} Natural & Caltech101 & \num{3060} & 102 & \cite{li2006one} \\
\raisebox{0.5pt}{\tikz\fill[natural] (0,0) circle (.5ex);} Natural & CIFAR-100 & \num{50000} & 100 & \cite{cifar10} \\
\raisebox{0.5pt}{\tikz\fill[natural] (0,0) circle (.5ex);} Natural & DTD & \num{3760} & \num{47} & \cite{cimpoi14describing} \\
\raisebox{0.5pt}{\tikz\fill[natural] (0,0) circle (.5ex);} Natural & Flowers102 & \num{2040} & 102 & \cite{Nilsback08} \\
\raisebox{0.5pt}{\tikz\fill[natural] (0,0) circle (.5ex);} Natural & Pets & \num{3680} & 37 & \cite{parkhi12a} \\
\raisebox{0.5pt}{\tikz\fill[natural] (0,0) circle (.5ex);} Natural & Sun397 & \num{87003} & 397 & \cite{xiao2010sun} \\
\raisebox{0.5pt}{\tikz\fill[natural] (0,0) circle (.5ex);} Natural & SVHN & \num{73257} & 10 & \cite{netzer2011reading} \\
\arrayrulecolor{lightgray}\midrule[0.5pt]\arrayrulecolor{black}
\raisebox{0.5pt}{\tikz\fill[specialized] (0,0) circle (.5ex);} Specialized & EuroSAT & \num{21600} & 10 & \cite{helber2017eurosat} \\
\raisebox{0.5pt}{\tikz\fill[specialized] (0,0) circle (.5ex);} Specialized & Resisc45 & \num{25200} & 45 & \cite{cheng2017remote} \\
\raisebox{0.5pt}{\tikz\fill[specialized] (0,0) circle (.5ex);} Specialized & Patch Camelyon & \num{294912} & 2 & \cite{veeling2018rotation} \\
\raisebox{0.5pt}{\tikz\fill[specialized] (0,0) circle (.5ex);} Specialized & Retinopathy & \num{46032} & 5 & \cite{kaggle-diabetic-retinopathy} \\
\arrayrulecolor{lightgray}\midrule[0.5pt]\arrayrulecolor{black}
\raisebox{0.5pt}{\tikz\fill[structured] (0,0) circle (.5ex);} Structured & Clevr/count & \num{70000} & 8 & \cite{johnson2017clevr} \\
\raisebox{0.5pt}{\tikz\fill[structured] (0,0) circle (.5ex);} Structured & Clevr/distance & \num{70000} & 6 & \cite{johnson2017clevr} \\
\raisebox{0.5pt}{\tikz\fill[structured] (0,0) circle (.5ex);} Structured & dSprites/location & \num{663552} & 16 & \cite{dsprites17} \\
\raisebox{0.5pt}{\tikz\fill[structured] (0,0) circle (.5ex);} Structured & dSprites/orientation & \num{663552} & 16 & \cite{dsprites17} \\
\raisebox{0.5pt}{\tikz\fill[structured] (0,0) circle (.5ex);} Structured & SmallNORB/azimuth & \num{36450} & 18 & \cite{lecun2004learning} \\
\raisebox{0.5pt}{\tikz\fill[structured] (0,0) circle (.5ex);} Structured & SmallNORB/elevation & \num{36450} & 9 & \cite{lecun2004learning} \\
\raisebox{0.5pt}{\tikz\fill[structured] (0,0) circle (.5ex);} Structured & DMLab & \num{88178} & 6 & \cite{beattie2016deepmind} \\
\raisebox{0.5pt}{\tikz\fill[structured] (0,0) circle (.5ex);} Structured & KITTI/distance & \num{5711} & 4 & \cite{Geiger2013IJRR} \\
\bottomrule
\end{tabular}
\caption{Description of the datasets used for the tasks in VTAB.}
\label{tab:tasks}
\end{table}

Table~\ref{tab:tasks} provides statistics and references for all of the tasks in VTAB.

We now provide a brief description of each task.

\begin{description}
    \item[Caltech101] \citep{li2006one} The task consists in classifying pictures of objects (101 classes plus a background clutter class), including animals, airplanes, chairs, or scissors. The image size varies, but it typically ranges from 200-300 pixels per edge.
    \item[CIFAR-100] \citep{cifar10}  The task consists in classifying natural images (100 classes, with 500 training images each). Some examples include apples, bottles, dinosaurs, and bicycles.
    The image size is 32x32.
    \item[DTD] \citep{cimpoi14describing} The task consists in classifying images of textural patterns (47 classes, with 120 training images each). Some of the textures are banded, bubbly, meshed, lined, or porous.
    The image size ranges between 300x300 and 640x640 pixels.
    \item[Flowers102] \citep{Nilsback08} The task consists in classifying images of flowers present in the UK (102 classes, with between 40 and 248 training images per class). Azalea, Californian Poppy, Sunflower, or Petunia are some examples.
    Each image dimension has at least 500 pixels.
    \item[Pets] \citep{parkhi12a} The task consists in classifying pictures of cat and dog breeds (37 classes with around 200 images each), including Persian cat, Chihuahua dog, English Setter dog, or Bengal cat. Images dimensions are typically 200 pixels or larger.
    \item[Sun397] \citep{xiao2010sun} The Sun397 task is a scenery benchmark with 397 classes and, at least, 100 images per class. Classes have a hierarchy structure, and include cathedral, staircase, shelter, river, or archipelago. The images are (colour) 200x200 pixels or larger.
    \item[SVHN] \citep{netzer2011reading} This task consists in classifying images of Google's street-view house numbers (10 classes, with more than 1000 training images each).
    The image size is 32x32 pixels.
    \item[EuroSAT] \citep{helber2017eurosat} The task consists in classifying Sentinel-2 satellite images into 10 different types of land use (Residential, Industrial, River, Highway, etc). The spatial resolution corresponds to 10 meters per pixel, and the image size is 64x64 pixels.
    \item[Resisc45] \citep{cheng2017remote} The Remote Sensing Image Scene Classification (RESISC) dataset is a scene classification task from remote sensing images. There are 45 classes, containing 700 images each, including tennis court, ship, island, lake, parking lot, sparse residential, or stadium. The image size is RGB 256x256 pixels.
    \item[Patch Camelyon] \citep{veeling2018rotation} The Patch Camelyon dataset contains \num{327680} images of histopathologic scans of lymph node sections. The classification task consists in predicting the  presence of metastatic tissue in given image (i.e., two classes). All images are 96x96 pixels. 
    \item[Retinopathy] \citep{kaggle-diabetic-retinopathy} The Diabetic Retinopathy dataset consists of image-label pairs with high-resolution retina images, and labels that indicate the presence of Diabetic Retinopahy (DR) in a 0-4 scale (No DR, Mild, Moderate, Severe, or Proliferative DR).
    \item[Clevr/count] \citep{johnson2017clevr} CLEVR is a visual question and answer dataset designed to evaluate algorithmic visual reasoning. We use just the images from this dataset, and create a synthetic task by setting the label equal to the number of objects in the images.
    \item[Clevr/distance] \citep{johnson2017clevr} Another synthetic task we create from CLEVR consists of predicting the depth of the closest object in the image from the camera. The depths are bucketed into size bins.
    \item[dSprites/location] \citep{dsprites17} The dSprites dataset was originally designed to asses disentanglement properties of unsupervised learning algorithms. In particular, each image is a 2D shape where six factors are controlled: color, shape, scale, rotation, and (x,y) center coordinates. Images have 64x64 black-and-white pixels.
    This task consists in predicting the x (horizontal) coordinate of the object.
    The locations are bucketed into 16 bins.
    \item[dSprites/orientation] \citep{dsprites17} We create another task from dSprites consists in predicting the orientation of each object, bucketed into 16 bins.
    \item[SmallNORB/azimuth] \citep{lecun2004learning} The Small NORB dataset contains images of 3D-toys from 50 classes, including animals, human figures, airplanes, trucks, and cars. The image size is 640x480 pixels. In this case, we define labels depending on the azimuth 	(angle of horizontal deviation), in intervals of 20 degrees (18 classes).
    \item[SmallNORB/elevation] \citep{lecun2004learning} Another synthetic task we create from Small NORB consists in predicting the elevation in the image. There are 9 classes, corresponding to 9 different elevations ranging from 30 to 70 degrees, in intervals of 5 degrees.
    \item[DMLab] \citep{beattie2016deepmind} The DMLab (DeepMind Lab) is a set of control environments focused on 3D navigation and puzzle-solving tasks.
    The Dmlab dataset contains frames observed by the agent acting in the DeepMind Lab environment,
    which are annotated by the distance between the agent and various objects present in the environment.
    The goal is to evaluate the ability of a visual model to reason about distances from the visual input in 3D environments.
    The Dmlab dataset consists of 360x480 color images in 6 classes.
    The classes are \{close, far, very far\} $\times$ \{positive reward, negative reward\} respectively.
    \item[KITTI-Dist] \citep{Geiger2013IJRR} The KITTI task consists in predicting the (binned) depth to the vehicle (car, van, or truck)  in the image. There are 4 bins / classes.
\end{description}

\clearpage
\section{Human Evaluation \label{app:human}}

We define $P(\t)$ informally as ``All tasks that a human can solve using visual input alone''. Intuitively, the tasks in \cref{app:tasks} satisfy this property because the task semantics involve simple visual concepts (for a human), and the images contain recognisable objects -- either in a natural or artificial environment. However, we also check empirically that humans can solve the types of tasks using in VTAB \emph{from examples alone}. We therefore evaluate human raters on a representative subset of the tasks used in VTAB: Pets (natural images, fine-grained object labels), DTD (natural, textures), Camelyon (specialized, medical images), EuroSAT (specialized, aerial images), DMLab (structured, distance prediction), and Clevr-count (structured, object counting). 

For each human-evaluated dataset, raters are given 20 random examples for each class, taken from the training split. Raters are asked to classify between 50 and 100 images each, for a total of 1K images (except for DMLab: 534 images), randomly taken from the test split, based on the provided training examples. Beside the examples, no hints nor explanations are given on the nature of the tasks. The raters have to deduce which properties of each image should be used to classify those, for example: the breed of the animal, the distance between the camera and the objects, or the number of objects. The raters are asked not to exchange on the tasks, so each rater produces independent work. 

All datasets rated by humans are the same as the one rated by the models, except for DMLab, where we only assess the distance prediction aspect and not object type. This is because there are too many object types for the raters to learn the two groups of objects from 20 examples per class. Therefore, we asses distance-prediction alone. The human DMLab task contains only 3 classes, each of which contains many object types, and differ only in object distance.

Note that this evaluation is not meant to quantify the relative performances of humans and machines, since protocol differences e.g. number of training examples, render the performance incomparable. Instead, it is meant as a verification that the kinds of domains (natural images, aerial imagery, etc.) and semantics (object type classification, localization, counting, etc.) are possible to learn from examples alone using human-level visual representations.

\begin{table}[h]
\centering
\begin{tabular}{lrr}
\toprule
& Random Guess & Human \\
\midrule
Pets & $2.7\%$ & $63.1\%$ \\
DTD  & $2.1\%$ & $64.0\%$  \\
Camelyon & $50\%$ & $68.0\%$  \\
EuroSAT & $10\%$ & $86.5\%$  \\
DMLab & $33.3\%$ & $49.0\%$  \\
Clevr-count & $12.5\%$  & $99.0\%$  \\
\bottomrule
\end{tabular}
\caption{
Human evaluation scores, measured using mean-per-class accuracy.
\label{tab:app-human}}
\end{table}

We measure human performance using mean-per-class accuracy. This is because the human training sets are class-balanced, so the raters cannot learn the class priors which algorithms that see an i.i.d sample of the training set could. \cref{tab:app-human} shows the results. The results indicate that some tasks are harder due to more subtle distinctions (Camelyon) or noise (DMLab), and some very easy (Clevr-count). However, in all cases the raters perform significantly better than random guessing. This demonstrates that the types of tasks used in VTAB are ones for which human-like visual representations are useful to solve them, using few labels and visual-input alone.

\clearpage
\section{Network Architectures\label{app:architectures}}

All of the methods (except for \emph{BigGAN} models) we evaluate in the paper use the standard ResNet50-v2~\citep{he2016identity} architecture.

This architecture consists of 50 convolutional layers.
The representation produced by this model has 2048 dimensions.
Exact details of this architecture can be found in~\citep{he2016identity}.

In the \emph{Cond-BigGAN} and \emph{Uncond-BigGAN} models we use publicly available\footnote{\url{https://github.com/google/compare_gan/blob/master/compare_gan/architectures/resnet_biggan.py}} implementation by \cite{lucic2019high} and \citep{chen2019self} of the custom ResNet-like architecture proposed and described in~\citep{brock2018large}. It has 1536 dimensions in the final representation layer.

We use a specialized procedure for evaluating patch-based models (\emph{Relative Patch Location} \citep{doersch2015unsupervised} and \emph{Jigsaw} \citep{noroozi2016unsupervised}). These models use ResNet50 model with the overall stride reduced from 64 to 16 (by substituting the first and the last strided convolution of the standard ResNet50 model by a convolution with stride one). During the adaptation phase, we apply ResNet50 with reduced stride in the following way (assuming that input image size is $224 \times 224$):
\begin{itemize}
\item Perform central crop of size $192 \times 192$.
\item Cut image into $3 \times 3$ patches of size $64 \times 64$.
\item Apply the ResNet50 model independently to every patch.
\item Output the final representation and element-wise average of 9 individual patch representations.
\end{itemize}

\clearpage
\section{Alternative Metrics\label{app:metrics}}

\begin{figure}[h]
    \centering
    \includegraphics[width=1.0\textwidth]{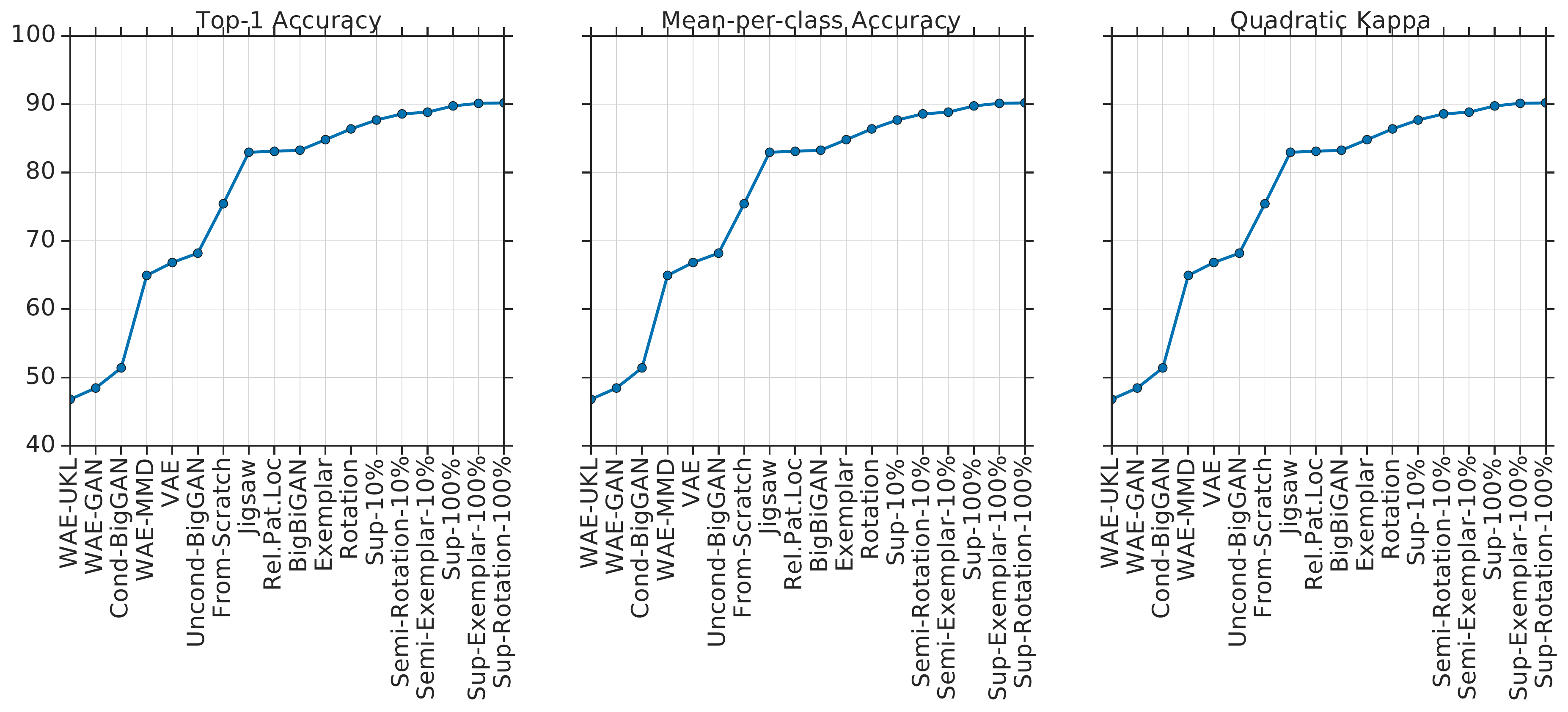}
    \caption{Ranking of the methods using the average scores across datasets (validation split) using three different metrics:
    top-1 accuracy (left),
    mean-per-class accuracy (center),
    Cohen's quadratic kappa (right).
    The methods on the x-axis are sorted according to the highest scores according to each metric.
    Although there are some minor changes in the ranking between top-1 and Cohen's quadratic kappa, 
    the overall performance of groups of methods remains unchanged.}
    \label{fig:metrics}
\end{figure}

In previous works, some datasets are
reported using alternative metrics to top-1 accuracy:
mean-per-class accuracy (Caltech101)
or Cohen's quadratic kappa (Retinopathy).
We study whether these metrics reveal different results to top-1 accuracy.
We find that, although there are minor ranking differences, the overall picture remains unchanged.
Kendall's ranking correlation scores, with respect to top-1 accuracy are 1.0 (mean per-class accuracy) and 0.97 (quadratic kappa).

Figure~\ref{fig:metrics} shows different test metric scores
(top-1 accuracy, mean-per-class accuracy, and Cohen's quadratic kappa),
averaged across all datasets in the benchmark, for the different methods
studied. The methods were ranked in the x-axis according to their score 
in each metric.
Observe that top-1 and mean-per-class accuracy give exactly the same
ranking. There are subtle differences when using
Cohen's quadratic kappa, but the overall picture
remains unchanged: supervised methods outperform semi-supervised methods
by a small margin, followed by self-supervised methods, etc.
Kendall's ranking correlation scores, with respect to top-1 accuracy, are:
1.0 for mean per-class accuracy, and 0.97 for Cohen's quadratic kappa.
This confirms that our conclusions hold even if we are not using the standard
metrics for a few datasets.

\clearpage
\section{Alternative weighting and ranking schemes for models\label{app:alt-model-ranking}}

Mean top-1 accuracy across tasks creates an implicit weighting.
First, some tasks use the same input data (e.g Clevr-Count and Clevr-Dist), thus upweighting those domains.
Second, the task groups differ in size.
Third, the tasks exhibit different performance ranges across methods.
Therefore, we compare seven different ranking aggregation strategies:
(i)~Macro-average across datasets (grouping tasks with the same input data).
(ii)~Macro-average across groups (merging tasks in the same group).
(iii)~Geometric mean.
(iv)~Average rank.
(v)~Average rank after small perturbation of the scores with noise.
(vi)~Robust (binned) average rank.
(vii)~Elimination rank -- equivalent to an ``Exhaustive Ballot''.

All strategies have a high Kendall's correlation with the vanilla mean across tasks ($\tau>0.87$).
The most dissimilar strategy is the average rank, with $\tau=0.873$.
Therefore, we use mean top-1 accuracy because it is interpretable and can be computed independently for each method, unlike rank-based aggregation.

Throughout, we use (unweighted) mean top-1 accuracy across all tasks to rank the  different models. This assumes that the samples represent our desired task distribution $P_T$ in an unbiased manner. However, there may be other sensible weighting schemes that are not uniform. Here we explore three alternative  weighting schemes (i) assigning equal weight to every dataset -- there are three data sets that are used in two tasks, and the others are used in one task, (ii) assigning equal weight to each task group: \taskNatural{}, \taskSpecialized{}, and \taskStructured{}, and (iii) the weighting scheme introduced in \citet{balduzzi2018}.

Another issue inherent in (weighted or unweighted) mean accuracy is that the mean accuracies of individual tasks vary significantly depending on the task's difficulty. Since maximum accuracy is bounded, this may limit the range of performances, implicitly upweighting tasks with more ``headroom''. Therefore, we explore a few simple alternative ranking strategies (see \citet{dwork2001rank} for an introduction raking aggregation methods): 
\begin{enumerate}[label=(\roman*)]
\item Ranking according to the geometric mean.
\item The average rank obtained by ranking the models for each task according to accuracy and then computing the average rank across tasks.
\item A noise-perturbed version of (i) in which the accuracies are perturbed by Gaussian noise with variance $1.0$ prior to ranking.
\item A robust variant of the average rank, where, prior to averaging ranks across tasks, the accuracy is binned into buckets of size $1\%$ and all models in the same bucket obtain the same rank.
\item An elimination ranking scheme equivalent to the ``Exhaustive Ballot'' voting system, see e.g. \citep{shahandashti2016electoral}.
\end{enumerate}
We measure the agreement between these ranking strategies using the Kendall rank correlation coefficient~\citep{kendall1945treatment}. To account for the training/evaluation stochasticity in VTAB, we sample (independently for every pair of model and task) one out of the three test-set repetitions, and compute the rank correlation between ranking according to the mean accuracy and each alternative ranking. We average over 100 such samples. The results are in Table~\ref{tab:altrank}. The mean rank correlation between the ranking according to the mean across tasks correlates very well with alternative ranking schemes. In particular, weighted means for 1000 samples and the full dataset as well as the geometric mean have a rank correlation with the ranking according to the mean that exceeds $0.95$. The agreement with different types of average rank somewhat lower for 1000 samples, but still considerable. We hence conclude that the mean accuracy across tasks is a fairly robust metric to rank models in VTAB.

\begin{table}[h]
\begin{center}
\begin{tabular}{lrr}
\toprule
\bf Ranking Method & \bf 1000 Samples & \bf Full Dataset \\
\toprule
Reweighted data sets & $0.959\pm0.017$ & $0.978\pm0.015$ \\
Reweighted groups & $0.969\pm0.017$ & $0.991\pm0.010$ \\
\citet{balduzzi2018} weighting & $0.891\pm0.027$ & $0.950\pm0.018$ \\
\midrule
Geometric mean & $0.951\pm0.020$ & $0.973\pm0.016$ \\
Average rank & $0.873\pm0.023$ & $0.925\pm0.019$ \\
Average rank (perturbed) & $0.874\pm0.024$ & $0.926\pm0.028$ \\
Average rank (robust) & $0.877\pm0.024$ & $0.957\pm0.016$ \\
Elimination rank & $0.930\pm0.031$ & $0.929\pm0.018$ \\
\bottomrule
\end{tabular}

\caption{\label{tab:altrank} Kendall rank correlation coefficient \citep{kendall1945treatment} (with standard deviation) measuring the agreement of the ranking of models according to the mean accuracy across tasks with different types of weighted means, the geometric mean, and different types of average ranks. Agreement of ranking according to mean accuracy with alternative ranking schemes is high.}
\end{center}
\end{table}

\clearpage
\section{Representative Subset of Tasks \label{app:subset-model-ranking}}

We explore another direction related to ranking of models: Which subset of tasks is most representative of the performance of the entire benchmark? Such a subset allows for cheap iteration  which is beneficial during early development of new models. 

We search the most representative $5$ tasks from the full set of $20$ tasks by performing exhaustive search over all $20 \choose 5$ subsets. For each subset we compute the mean accuracy and compute the Kendall rank correlation coefficient between the resulting ranking of models and the ranking according to the mean over all tasks (averaged over 10 trials sampled as described in \cref{app:alt-model-ranking}). \cref{tab:subsetrank} shows the subsets which produce the highest rank correlation. There are many subsets which lead to an excellent agreement with the ranking according to the full mean. These subsets can be quite different provided that they are sufficiently diverse.

To assess how well we can expect this approach generalize to unseen models, we perform the following experiment. For each pair of models, we perform subset selection based on the accuracies of the remaining models, and check whether the ranking of the left out pair based on the mean across all data sets agrees with the ranking of the pair according to the mean over the subset. The mean outcome of this binary test across all pairs is an estimate of the probability for correctly ranking an unseen pair of models: $0.952$ for the full data sets and $0.956$ for 1000 examples. The subset selection hence generalizes to an unseen pair of models with high probability.

The representative subsets in \cref{tab:subsetrank} may be used for rapid prototyping of new methods before running the full VTAB.
However, a caveat to the above analyses is that we evaluate the ranking of a very diverse set of models, from those whose performance is worse than from-scratch training to performance combinations of self-supervision and supervision.
Therefore, the ranking is reasonably robust.
To make fine-grained distinctions between models of a similar class, the representative subsets above may be less reliable.
Repeated iteration on just a few tasks exposes one more to the risk of meta-overfitting.

\begin{table}[h]
\begin{center}
\begin{tabular}{lcl}
\toprule
\bf Dataset & \bf Rank Correlation & \bf Subset \\
\toprule
\multirow{5}{*}{1000}
 & $0.965\pm0.020$ & Clevr-Count, DMLab, Pets, Camelyon, Sun397 \\
 & $0.963\pm0.016$ & Clevr-Dist, Clevr-Count, DTD, Flowers102, Pets \\
 & $0.963\pm0.013$ & Clevr-Count, DTD, Pets, Camelyon, sNORB-Azim \\
 & $0.962\pm0.014$ & Clevr-Count, DTD, Pets, sNORB-Azim, SVHN \\
 & $0.961\pm0.019$ & Clevr-Count, DTD, Pets, Camelyon, SVHN \\
\midrule
\multirow{5}{*}{full}
 & $0.976\pm0.011$ & CIFAR-100, Clevr-Count, Flowers102, Pets, sNORB-Elev \\
 & $0.976\pm0.011$ & Clevr-Count, Flowers102, Camelyon, sNORB-Elev, Sun397 \\
 & $0.976\pm0.010$ & CIFAR-100, Clevr-Count, Retinopathy, DMLab, Flowers102 \\
 & $0.975\pm0.021$ & CIFAR-100, Clevr-Count, DMLab, Flowers102, SVHN \\
 & $0.975\pm0.012$ & CIFAR-100, Clevr-Count, Pets, Camelyon, Sun397 \\
\bottomrule
\end{tabular}

\caption{\label{tab:subsetrank} Task subsets of size 5 that lead to the largest rank correlation with the mean accuracy across all tasks. There are many different subsets that lead to an high rank correlation with the full mean.}
\end{center}
\end{table}

\clearpage
\section{Lightweight experiments\label{app:lightweight}}

\begin{table}[h]
\fontsize{7pt}{7pt}\selectfont
\newcolumntype{C}{>{\centering\arraybackslash}X}
\setlength{\tabcolsep}{0pt}
\setlength{\extrarowheight}{5pt}
\renewcommand{\arraystretch}{0.75}
\begin{tabularx}{\linewidth}{p{10pt}p{1.6cm}!{\color{lightgray}\vline} CCCCCCC!{\color{lightgray}\vline}CCCC!{\color{lightgray}\vline}CCCCCCCC!{\color{lightgray}\vline}C}
\toprule
 &
 & \rotatebox{90}{\tikz\fill[natural] (0,0) circle (.5ex); Caltech101}
 & \rotatebox{90}{\tikz\fill[natural] (0,0) circle (.5ex); CIFAR-100}
 & \rotatebox{90}{\tikz\fill[natural] (0,0) circle (.5ex); DTD}
 & \rotatebox{90}{\tikz\fill[natural] (0,0) circle (.5ex); Flowers102}
 & \rotatebox{90}{\tikz\fill[natural] (0,0) circle (.5ex); Pets}
 & \rotatebox{90}{\tikz\fill[natural] (0,0) circle (.5ex); Sun397}
 & \rotatebox{90}{\tikz\fill[natural] (0,0) circle (.5ex); SVHN}
 & \rotatebox{90}{\tikz\fill[specialized] (0,0) circle (.5ex); Camelyon}
 & \rotatebox{90}{\tikz\fill[specialized] (0,0) circle (.5ex); EuroSAT}
 & \rotatebox{90}{\tikz\fill[specialized] (0,0) circle (.5ex); Resisc45}
 & \rotatebox{90}{\tikz\fill[specialized] (0,0) circle (.5ex); Retinopathy}
 & \rotatebox{90}{\tikz\fill[structured] (0,0) circle (.5ex); Clevr-Count}
 & \rotatebox{90}{\tikz\fill[structured] (0,0) circle (.5ex); Clevr-Dist}
 & \rotatebox{90}{\tikz\fill[structured] (0,0) circle (.5ex); DM-Lab}
 & \rotatebox{90}{\tikz\fill[structured] (0,0) circle (.5ex); dSpr-Loc}
 & \rotatebox{90}{\tikz\fill[structured] (0,0) circle (.5ex); dSpr-Ori}
 & \rotatebox{90}{\tikz\fill[structured] (0,0) circle (.5ex); KITTI-Dist}
 & \rotatebox{90}{\tikz\fill[structured] (0,0) circle (.5ex); sNORB-Azim}
 & \rotatebox{90}{\tikz\fill[structured] (0,0) circle (.5ex); sNORB-Elev}
 & \rotatebox{90}{\tikz\fill[all] (0,0) circle (.5ex); Mean} \\
 
\midrule

\multirow{16}{*}{\rotatebox{90}{1000}}
& WAE-UKL &       31.0 &       8.4 &      6.6 &       11.9 &      7.7 &      2.7 &     37.8 &     65.9 &     60.7 &     16.2 &        61.1 &        30.6 &       54.2 &     27.4 &     70.5 &        20.0 &       49.7 &       15.3 &       27.6 &     31.9 \\
      & WAE-GAN &       30.3 &       7.1 &      6.0 &       11.7 &      7.4 &      3.2 &     43.9 &     67.4 &     56.9 &     15.1 &        59.8 &        33.4 &       52.6 &     27.6 &     68.8 &        19.6 &       51.3 &       20.6 &       25.3 &     32.0 \\
      & Cond-BigGAN &       61.8 &      20.9 &      2.1 &       60.2 &     23.6 &     10.9 &     56.8 &     74.7 &     29.6 &     49.8 &    \bf 73.6 &        12.7 &       24.5 &     22.2 &      6.2 &        51.3 &       34.5 &       22.2 &       32.8 &     35.3 \\
      & WAE-MMD &       37.7 &      10.3 &      8.8 &       18.8 &     10.4 &      4.8 &     56.5 &     68.8 &     72.5 &     22.6 &        63.0 &        35.5 &       56.4 &     29.0 &     89.7 &        26.9 &       50.6 &       19.2 &       27.9 &     37.3 \\
      & VAE &       35.7 &       8.8 &      9.8 &       17.6 &      8.9 &      4.4 &     50.2 &     72.4 &     73.4 &     23.7 &        65.3 &        36.0 &       60.7 &     30.5 &     91.4 &        20.5 &       40.7 &       26.1 &       37.1 &     37.5 \\
      & From-Scratch &       37.7 &      11.0 &     23.0 &       40.2 &     13.3 &      3.9 &     59.3 &     73.5 &     84.8 &     41.6 &        63.1 &        38.5 &       54.8 &     35.8 &     87.9 &        37.3 &       36.9 &       20.9 &       36.9 &     42.1 \\
      & Un.C.-BigGAN &       58.0 &      16.4 &     24.8 &       48.8 &     17.3 &      8.8 &     66.5 &     76.7 &     73.2 &      7.2 &    \bf 73.6 &        43.5 &       57.7 &     32.4 &     77.7 &        47.7 &       38.1 &       25.4 &       42.2 &     44.0 \\
      & Rel.Pat.Loc. &       67.8 &      17.9 &     51.0 &       67.2 &     38.8 &     10.5 &     61.7 &     73.4 &     92.6 &     66.2 &        59.7 &        44.3 &       55.7 &     39.4 &     63.6 &        32.7 &       57.8 &       29.9 &       34.9 &     50.8 \\
      & Jigsaw &       66.2 &      14.8 &     50.7 &       65.3 &     34.0 &     11.4 &     54.9 &     73.0 &     91.5 &     66.7 &        71.3 &        44.1 &       56.2 &     42.2 &     66.0 &        34.2 &       63.8 &       32.9 &       31.7 &     51.1 \\
      & Exemplar &       69.1 &      12.5 &     48.4 &       66.7 &     42.0 &     14.5 &     88.0 &     76.8 &     94.7 &     68.1 &        73.1 &        48.3 &   \bf 62.0 &     45.3 &     92.0 &        45.5 &       73.4 &       30.9 &       40.8 &     57.5 \\
      & BigBiGAN &       80.8 &      39.2 &     56.6 &       77.9 &     44.4 &     20.3 &     76.8 &     77.4 &     95.6 &     74.0 &        69.3 &    \bf 53.9 &       55.6 &     38.7 &     70.6 &        46.7 &       71.4 &       27.2 &   \bf 46.3 &     59.1 \\
      & Rotation &       77.1 &      27.4 &     52.6 &       66.1 &     49.0 &     11.0 & \bf 89.6 &     77.8 &     93.9 &     70.3 &        72.3 &        44.4 &       58.6 &     46.1 &     87.0 &        49.2 &       77.2 &   \bf 39.3 &       42.5 &     59.5 \\
      & Sup-10\% &       84.9 &      46.1 &     55.4 &       82.8 &     80.2 &     24.9 &     86.9 &     80.6 &     95.2 &     73.1 &        71.9 &        39.4 &       55.7 &     43.5 &     76.0 &        45.9 &       63.3 &       30.9 &       32.9 &     61.6 \\
      & Semi-Ex-10\% &       87.7 &      52.8 &     60.4 &       84.0 &     84.0 &     29.2 &     87.1 &     79.1 &     94.9 &     77.1 &        70.1 &        39.4 &       56.0 &     42.3 &     73.7 &    \bf 52.3 &       72.0 &       37.6 &       33.5 &     63.9 \\
      & Semi-Rot-10\% &       87.4 &      45.6 &     61.9 &       86.8 &     84.1 &     27.0 &     88.0 &     81.2 & \bf 95.9 &     79.6 &        71.6 &        44.2 &       55.4 &     43.9 &     86.8 &        51.5 &       68.7 &       34.5 &       36.5 &     64.8 \\
      & Sup-100\% &       89.8 &      54.6 &     65.6 &       88.4 & \bf 89.1 & \bf 34.5 &     86.3 &     79.7 &     95.3 &     81.0 &        72.6 &        41.8 &       52.5 &     42.7 &     81.0 &        47.3 &       75.3 &       32.6 &       35.8 &     65.6 \\
      & Sup-Ex-100\% &       89.8 &  \bf 55.3 &     66.9 &       87.8 &     88.8 &     34.4 &     88.3 & \bf 82.1 &     95.5 &     80.6 &        71.6 &        47.1 &       56.2 &     48.3 &     83.0 &        45.6 &       75.6 &       35.9 &       39.2 &     67.0 \\
      & Sup-Rot-100\% &   \bf 89.9 &      52.8 & \bf 68.6 &   \bf 90.3 &     88.8 &     32.5 &     88.7 &     80.5 &     95.9 & \bf 83.4 &        73.2 &        48.2 &       57.0 & \bf 48.5 & \bf 92.5 &        50.0 &   \bf 79.1 &       30.6 &       32.8 & \bf 67.5 \\

\arrayrulecolor{lightgray}\specialrule{.5pt}{0.6pt}{-0.5pt}\arrayrulecolor{black}

\multirow{16}{*}{\rotatebox{90}{Full}}
& WAE-UKL &       41.7 &      23.2 &     12.3 &       17.2 &     12.3 &     12.0 &     65.5 &     76.4 &     78.1 &     36.8 &        73.6 &        44.5 &       67.8 &     36.7 &      98.1 &        51.4 &       55.1 &       35.9 &       51.0 &     46.8 \\
      & WAE-GAN &       42.0 &      24.8 &      8.7 &       15.5 &     13.1 &     12.8 &     78.2 &     77.1 &     81.5 &     38.4 &        73.6 &        52.2 &       70.2 &     37.3 &      97.7 &        49.9 &       62.3 &       33.4 &       52.2 &     48.5 \\
      & Cond-BigGAN &        0.1 &      56.3 &     44.8 &       68.8 &     31.6 &     44.9 &     91.4 &     81.3 &     94.5 &     76.5 &        75.3 &        12.4 &       24.5 &     51.4 &       6.2 &         7.4 &       49.7 &       80.6 &       79.2 &     51.4 \\
      & WAE-MMD &       50.8 &      38.8 &     11.0 &       20.8 &     16.2 &     31.6 &     90.9 &     80.6 &     94.1 &     64.8 &        73.8 &        98.1 &       89.3 &     52.6 & \bf 100.0 &        90.2 &       61.6 &       96.3 &       72.4 &     64.9 \\
      & VAE &       48.4 &      44.2 &     16.0 &       18.4 &     14.0 &     29.3 &     93.1 &     81.3 &     92.5 &     65.0 &        74.2 &        98.4 &       90.1 &     59.7 &     100.0 &        94.7 &       57.0 &       97.9 &       95.6 &     66.8 \\
      & Un.C.-BigGAN &       73.6 &      58.1 &     44.9 &       63.5 &     30.9 &     46.9 &     93.0 &     82.2 &     89.8 &     75.4 &        75.9 &        47.6 &       54.9 &     54.8 &      86.1 &        95.9 &       57.4 &       88.1 &       76.6 &     68.2 \\
      & From-Scratch &       55.9 &      64.4 &     31.3 &       50.6 &     23.8 &     52.7 &     96.3 &     81.2 &     96.2 &     86.8 &        76.8 &        99.7 &       89.4 &     71.5 & \bf 100.0 &        96.3 &       68.4 &       99.9 &       91.7 &     75.4 \\
      & Jigsaw &       79.1 &      65.3 &     63.9 &       77.9 &     65.4 &     59.2 &     93.9 &     83.0 &     97.9 &     92.0 &    \bf 80.1 &        99.6 &       88.6 &     72.0 &     100.0 &        90.3 &       74.7 &       99.9 &       93.6 &     83.0 \\
      & Rel.Pat.Loc. &       79.9 &      65.7 &     65.2 &       78.8 &     66.8 &     58.0 &     93.7 &     85.3 &     97.8 &     91.5 &        79.8 &        99.5 &       87.7 &     71.5 &     100.0 &        90.4 &       75.0 &       99.7 &       92.6 &     83.1 \\
      & BigBiGAN &       89.2 &      75.1 &     67.2 &       85.6 &     62.7 &     62.1 &     94.3 &     83.0 &     98.4 &     91.4 &        78.2 &        95.2 &       89.8 &     62.3 &     100.0 &        84.7 &       75.8 &       95.9 &       91.3 &     83.3 \\
      & Exemplar &       81.9 &      70.7 &     61.1 &       79.3 &     67.8 &     58.2 &     96.7 &     84.7 &     98.5 &     93.5 &        79.0 &        99.8 &   \bf 93.3 &     74.7 & \bf 100.0 &        96.5 &       78.2 &       99.9 &       97.4 &     84.8 \\
      & Rotation &       88.3 &      73.6 &     63.3 &       83.4 &     71.8 &     60.5 &     96.9 &     86.4 &     98.3 &     93.4 &        78.6 &        99.8 &       93.3 &     76.8 & \bf 100.0 &        96.5 &       82.6 &       99.9 &       98.0 &     86.4 \\
      & Sup-10\% &       90.0 &      78.5 &     67.6 &       88.5 &     85.0 &     63.1 &     96.5 &     85.0 &     98.4 &     94.2 &        78.7 &        99.7 &       92.1 &     74.6 & \bf 100.0 &        96.4 &       81.3 &       99.8 &       96.2 &     87.7 \\
      & Semi-Rot-10\% &       88.1 &      82.4 &     72.4 &       93.2 &     87.9 &     66.7 &     96.9 &     78.6 &     98.7 &     94.9 &        79.0 &    \bf 99.8 &       93.2 &     76.1 & \bf 100.0 &        96.5 &       81.0 &       99.9 &       97.5 &     88.6 \\
      & Semi-Ex-10\% &       85.3 &      82.7 &     70.5 &       92.2 &     89.0 &     67.4 &     97.0 &     86.0 &     98.6 &     94.7 &        78.8 &        99.8 &       93.1 &     76.8 & \bf 100.0 &    \bf 96.5 &       81.5 &  \bf 100.0 &       97.8 &     88.8 \\
      & Sup-100\% &       94.1 &      83.8 &     74.0 &       93.2 & \bf 91.9 & \bf 70.7 &     97.0 &     83.9 & \bf 98.8 & \bf 95.3 &        79.3 &        99.8 &       92.1 &     76.4 & \bf 100.0 &        96.4 &       80.7 &       99.8 &       97.7 &     89.7 \\
      & Sup-Ex-100\% &       94.4 &      84.1 &     74.5 &       93.4 &     91.8 &     69.4 & \bf 97.1 & \bf 86.7 &     98.6 &     95.1 &        79.5 &        99.8 &       92.7 & \bf 76.8 & \bf 100.0 &        96.4 &   \bf 84.0 &       99.8 &       98.0 &     90.1 \\
      & Sup-Rot-100\% &   \bf 94.6 &  \bf 84.8 & \bf 75.9 &   \bf 94.7 &     91.5 &     70.2 &     97.0 &     85.9 & \bf 98.8 &     94.9 &        79.5 &        99.8 &       92.5 &     76.5 & \bf 100.0 &        96.5 &       82.3 &      100.0 &   \bf 98.4 & \bf 90.2 \\

\bottomrule
\end{tabularx}

\caption{Top-1 accuracy of all the models evaluated on VTAB in lightweight mode.
Each entry represents the median score of three runs with different random seeds evaluated on the test set. 
Within each dataset-size group (1000-example and full), the methods are sorted from best to worst according to their mean accuracy.}
\label{tab:lightweight}
\end{table}

\clearpage
\section{Heavyweight experiments\label{app:heavyweight}}

The literature results may use substantially more complex architectures, or additional task-specific logic.
For example, the Retinopathy result uses three neural networks combined with decision trees, and uses additional information, such as combining the images from two eyes of the same patient.

\begin{table}[h]
\centering

\caption{
Top-1 accuracy of all the models evaluated on VTAB in heavyweight mode.
}
\label{tab:heavyweight_appendix}
\end{table}

\begin{table}[h]
\centering
\begin{threeparttable}[b]
\begin{tabular}{lccl}
\toprule
 & \textsc{Sup-Ex.-100\%} &  Result & Reference \\
\midrule
\raisebox{1pt}{\tikz\fill[natural] (0,0) circle (.5ex);} Caltech101   & 90.4 ~/~ 95.1\tnote{*} & 86.9 ~/~ 95.1\tnote{*} & \citet{cubuk2019} ~/~ \citet{kornblith2018better} \\
\raisebox{1pt}{\tikz\fill[natural] (0,0) circle (.5ex);} CIFAR-100    & 83.1 & 91.7 & \citet{tan2019efficientnet} \\
\raisebox{1pt}{\tikz\fill[natural] (0,0) circle (.5ex);} DTD          & 76.5 & 78.1 & \citet{kornblith2018better} \\
\raisebox{1pt}{\tikz\fill[natural] (0,0) circle (.5ex);} Flowers102   & 97.8 & 98.8 & \citet{tan2019efficientnet} \\
\raisebox{1pt}{\tikz\fill[natural] (0,0) circle (.5ex);} Pets         & 92.9 & 95.9 & \citet{huang2018gpipe} \\
\raisebox{1pt}{\tikz\fill[natural] (0,0) circle (.5ex);} SVHN         & 97.5 & 99.0 & \citet{cubuk2019} \\
\raisebox{1pt}{\tikz\fill[natural] (0,0) circle (.5ex);} Sun397       & 75.3 & 72.0 & \citet{Wang:2017} \\
\raisebox{1pt}{\tikz\fill[specialized] (0,0) circle (.5ex);} Camelyon     & 86.5 & 90.6 & \citet{teh2019metric} \\
\raisebox{1pt}{\tikz\fill[specialized] (0,0) circle (.5ex);} EuroSAT      & 99.0 & 96.4 & \citet{helber2017eurosat} \\
\raisebox{1pt}{\tikz\fill[specialized] (0,0) circle (.5ex);} Resisc45     & 96.3 & 90.4 & \citet{cheng2017remote} \\
\raisebox{1pt}{\tikz\fill[specialized] (0,0) circle (.5ex);} Retinopathy  & 74.7\tnote{$\dagger$} & 85.4\tnote{$\dagger$} & \citet{Wang:2017} \\
\bottomrule
\end{tabular}
\begin{tablenotes}[para]
\item[*] Mean per-class accuracy
\item[$\dagger$] Quadratic Kappa
\end{tablenotes}
\end{threeparttable}

\caption{
Comparison of the best method on the full datasets using heavyweight sweep (\textsc{Sup-Exemplar-100\%}) to results published in the literature (where available).
For some datasets prior work does not use top-1 accuracy, therefore, we present our the performance of \textsc{Sup-Exemplar-100\%} using the same metric.
}
\label{tab:literature}
\end{table}

\clearpage
\section{Hyperparameter Sweeps\label{app:hyperparameter_sweep}}

\paragraph{Lightweight sweep}
The lightweight mode performs a restricted hyperparameter search for all tasks, permitting fair comparison with few resources.
It uses fixed values for most hyperparameters.
We set the batch size to 512 and use SGD with momentum of 0.9.
When fine-tuning, we do not use weight decay, and when training from scratch we set it to 0.01 times the learning rate~\citep{loshchilov2019}.
We resize all images to $224\times224$, except for generative models, where we resize to $128\times128$ since training at a higher resolution is currently very challenging.

Lightweight mode sweeps four hyperparameters:
\begin{itemize}
  \item Learning rate: $\{0.1, 0.01\}$
  \item Traning schedule: In all cases we decay the learning rate by a factor of 10 after $\frac{1}{3}$ and $\frac{2}{3}$ of the training time, and one more time shortly before the end. We train for $\{\allowbreak2\,500, \allowbreak10\,000\}$ training steps (\emph{i.e.} model updates).
\end{itemize}

Note that when training on 1000 examples, we perform model selection using the regular validation set for each task.
This setting is somewhat unrealistic, since in practice one would train on the larger validation set.
However, we use this setting, which is common in prior art~\citep{olivier2018}, because it significantly reduces evaluation noise.
Further,~\citet{zhai2019s4l} show that using a small validation yields the same conclusions.

\paragraph{Heavyweight sweep}
We define a relatively large search-space over relevant hyperparameters for the fine-tuning adaptation phase and perform an individual random search for each downstream task's validation set in that space.
The random search consists of 100 independent trials.
Specifically, we search:
\begin{itemize}
    \item Batch-size: $\{128,\allowbreak 256,\allowbreak 512\}$
    \item Training schedule: In all cases we decay the learning rate by a factor of 10 after $\frac{1}{3}$ and $\frac{2}{3}$ of the training time, and one more time shortly before the end. We train for either
    \begin{itemize}
        \item any of $\{100,\allowbreak 200,\allowbreak 300,\allowbreak 400,\allowbreak 500\}$ epochs over the dataset, or
        \item any of $\{2\,500,\allowbreak 5\,000,\allowbreak 10\,000,\allowbreak 20\,000,\allowbreak 40\,000\}$ training steps (\emph{i.e.} model updates).
    \end{itemize}
    \item Preprocessing: during training, optionally include random horizontal flipping.
    \item Preprocessing: during training, optionally include random color distortions.
    \item Preprocessing: during training, use any of the following techniques for resizing:
    \begin{itemize}
        \item \texttt{res.}: resize the image to 224;
        \item \texttt{res.|crop}: resize the image to 256 and take a random crop of size 224;
        \item \texttt{res.sma|crop}: resize the image keeping its aspect ratio such that the smaller side is 256, then take a random crop of size 224;
        \item \texttt{inc.crop}: ``inception crop'' from \cite{szegedy2015going};
        \item \texttt{cif.crop}: resize the image to 224, zero-pad it by 28 on each side, then take a random crop of size 224.
    \end{itemize}
    \item Preprocessing: for evaluation, either
    \begin{itemize}
        \item resize the image to one of $\{224,\allowbreak 320,\allowbreak 384\}$, or
        \item resize it such that the smaller side is of size $\{256,\allowbreak 352,\allowbreak 448\}$ and then take a central crop of size $\{224,\allowbreak 320,\allowbreak 384\}$.
    \end{itemize}
    \item Weight decay: we sample log-uniform randomly in $[10^{-5}, 10^{-1}]$.
    \item Optimizer: we randomly choose between
    \begin{itemize}
        \item SGD with learning-rate chosen log-uniform randomly in $[0.001, 0.0]$, or
        \item Adam with learning-rate chosen log-uniform randomly in $[10^{-5}, 10^{-2}]$.
    \end{itemize}
\end{itemize}
We then choose the set of hyperparameters with the best performance at the end of training for each task individually.

\begin{figure}[!h]
\centering
\includegraphics[width=0.9\textwidth]{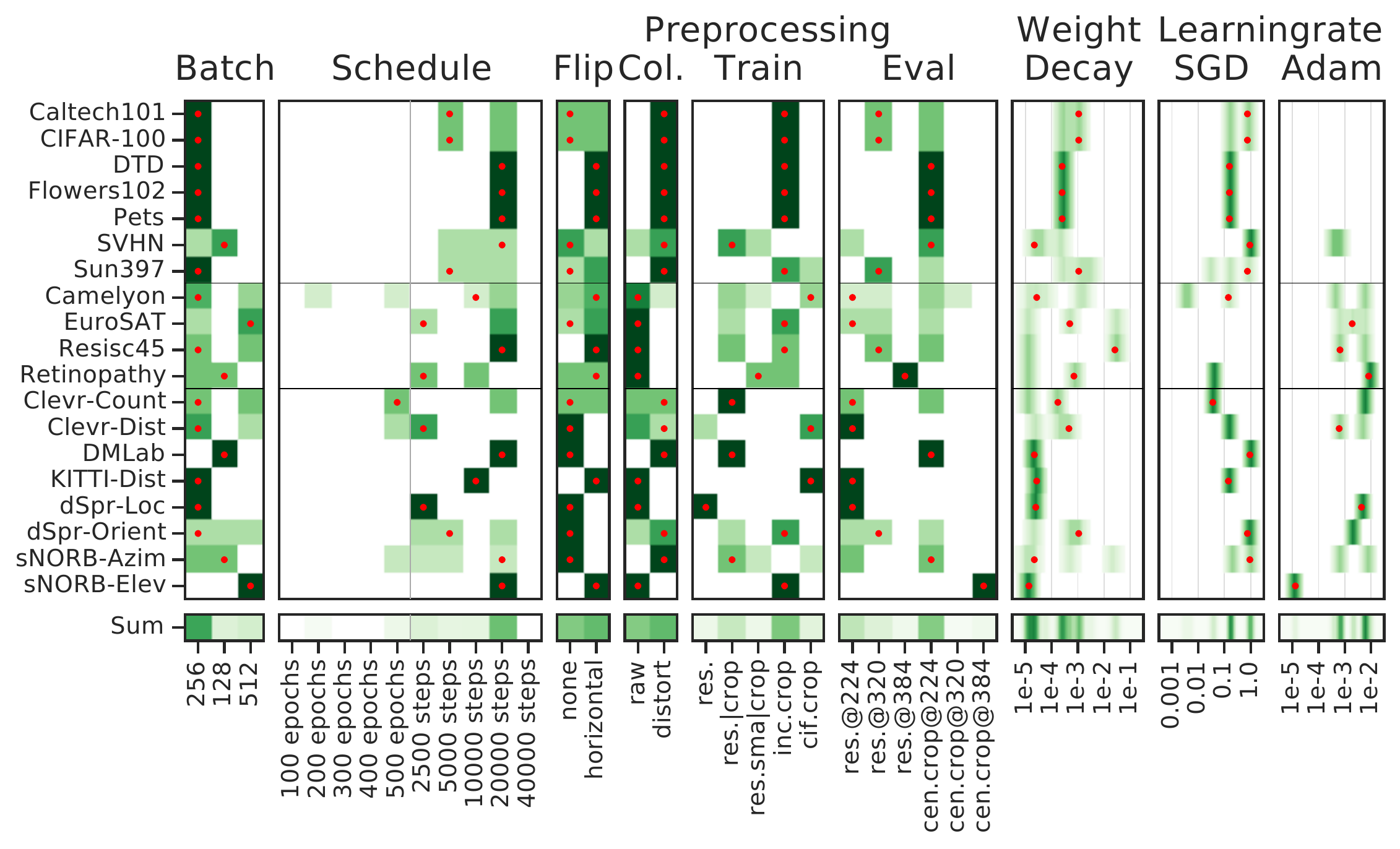}
\caption{Training from scratch with only 1000 examples. Best hyperparameter values are marked in red and those within $2\%$ in green to show sensitivity.}\label{fig:heavy:scratch:1000}
\end{figure}

\begin{figure}[!h]
\centering
\includegraphics[width=0.9\textwidth]{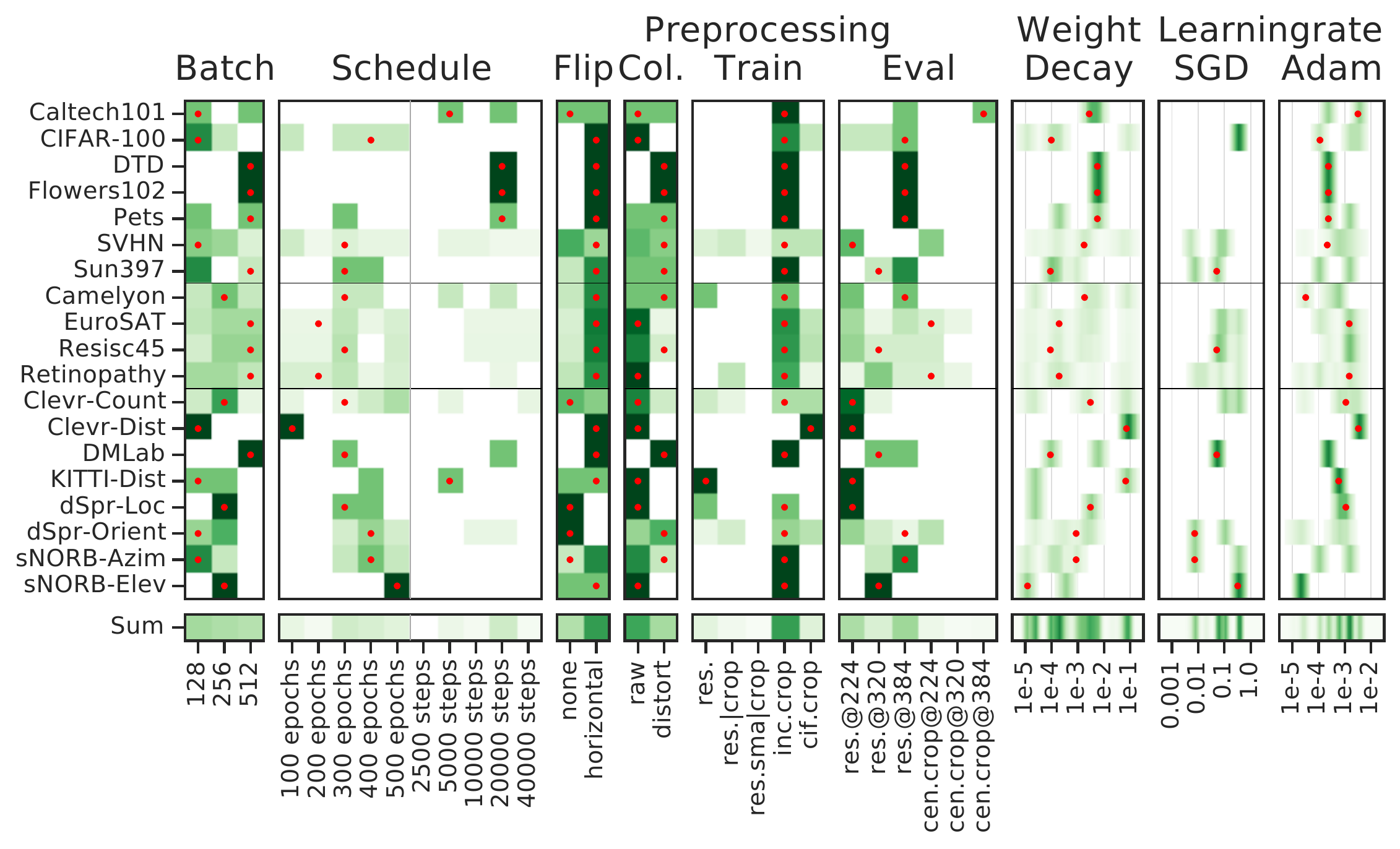}
\caption{Training from scratch on the full datasets. Best hyperparameter values are marked in red and those within $2\%$ in green to show sensitivity.}\label{fig:heavy:scratch:full}
\end{figure}

\begin{figure}[!h]
\centering
\includegraphics[width=0.9\textwidth]{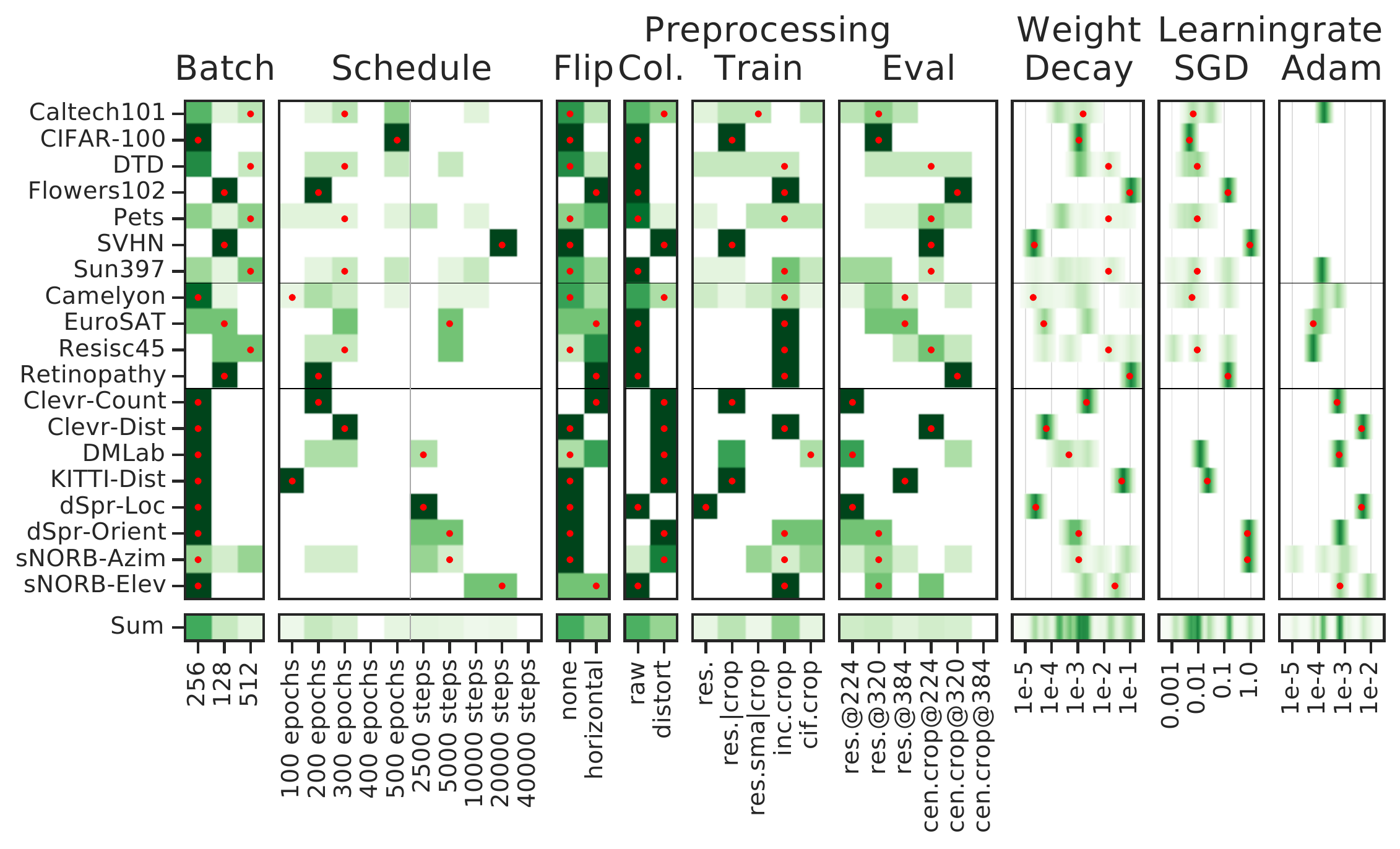}
\caption{\imagenet{} supervised representation fine-tuned on only 1000 examples. Best hyperparameter values are marked in red and those within $2\%$ in green to show sensitivity.}\label{fig:heavy:sup:1000}
\end{figure}

For both training from scratch, and from a pre-trained supervised \imagenet{} training, we present the best set of hyperparameters for each task in \cref{fig:heavy:scratch:1000,fig:heavy:scratch:full,fig:heavy:sup:1000,fig:heavy:scratch:full} as a red dot.
The green background shows the distribution of hyperparameters which performed within 2\% of the best one for each task, providing a sense for the importance of tuning that hyperparameter.

\begin{figure}[H]
\centering
\begin{tabular}{c}
\includegraphics[width=1.0\textwidth]{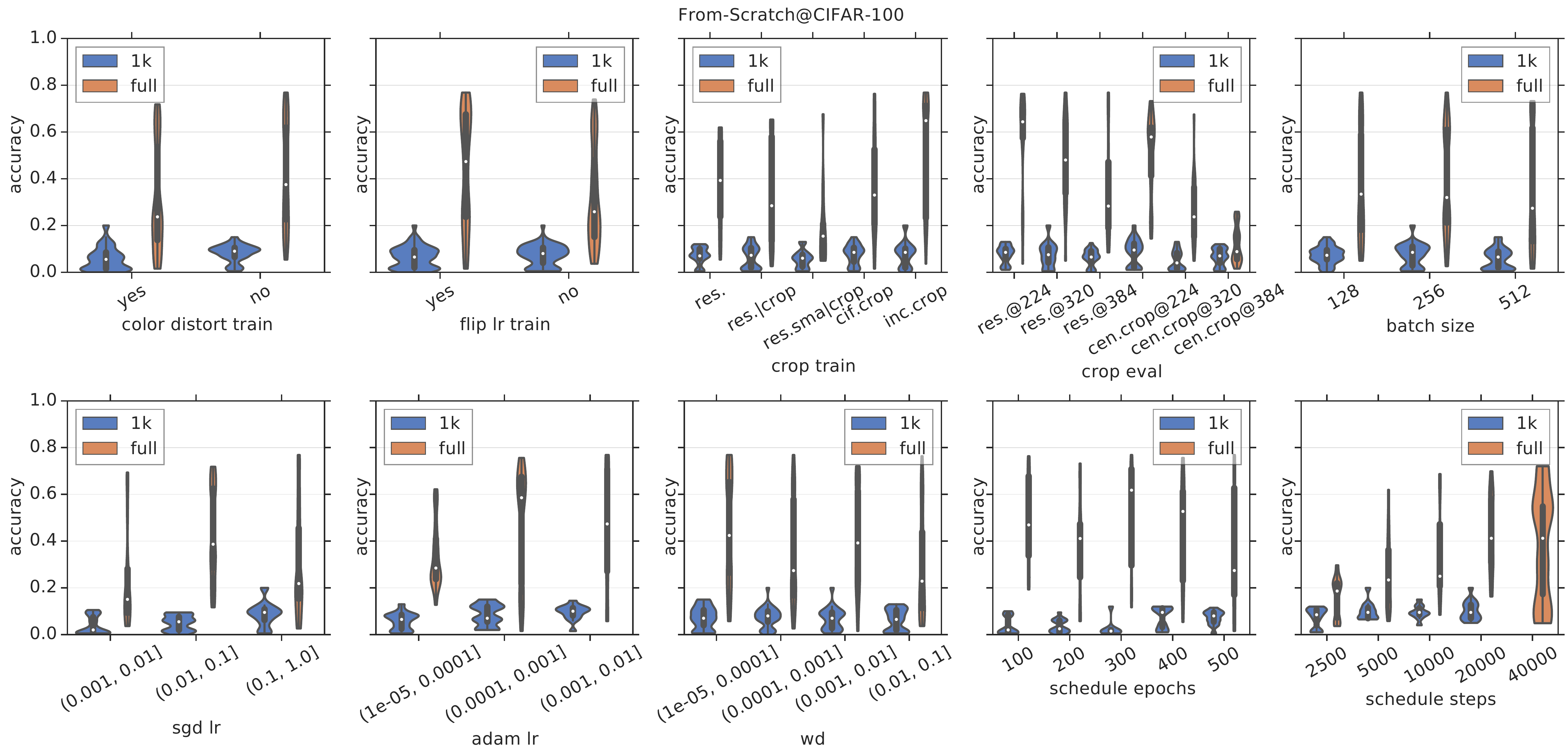}\\
\includegraphics[width=1.0\textwidth]{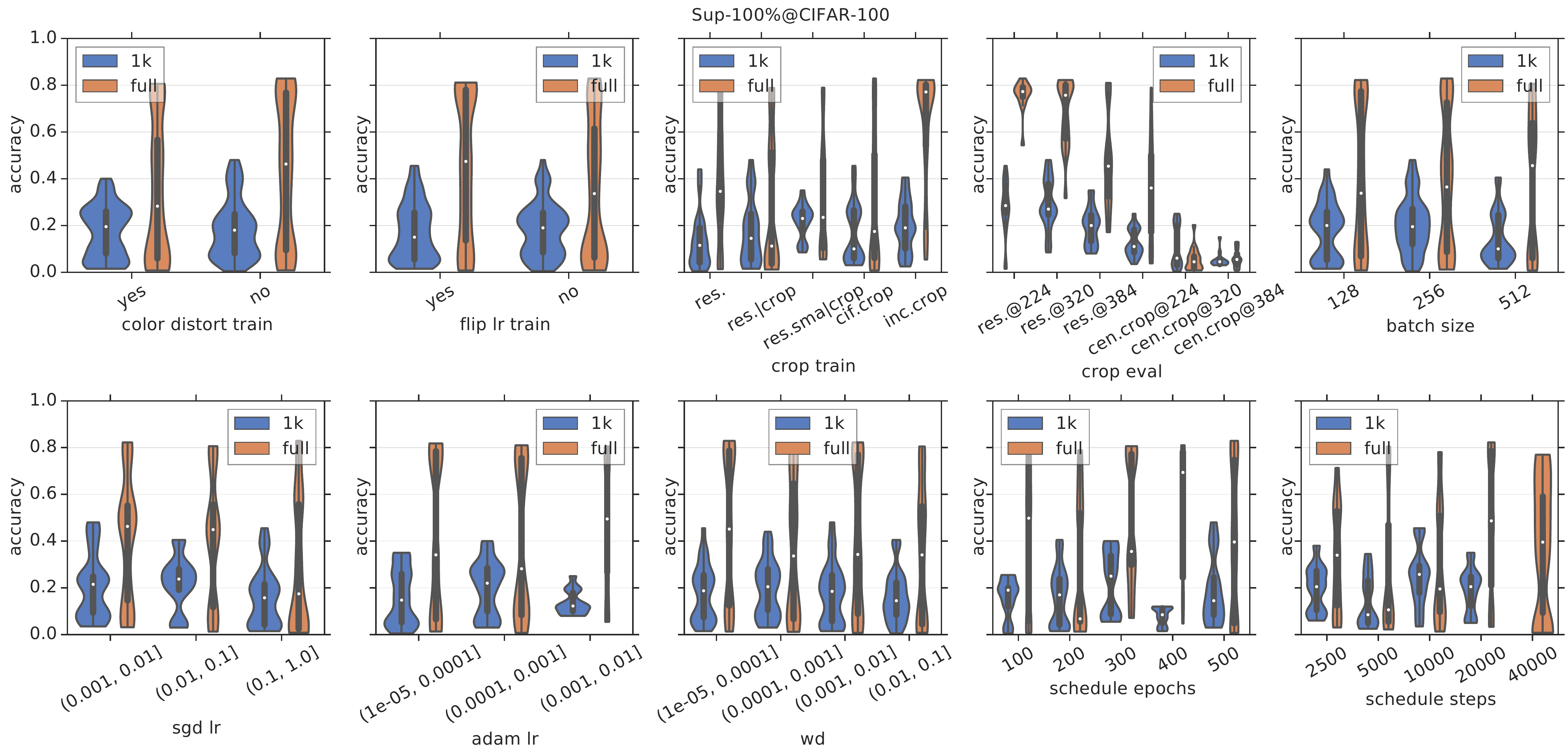}\\
\end{tabular}
\caption{
Heavy parameter sweep on CIFAR-100 dataset.
Top 2 rows show the results of from scratch algorithm and bottom 2 rows show the results of supervised 100\% algorithm.
Each plot shows the violin plot of the results with respect to a given parameter.
Here we show the results evaluated on full datasets and 1k datasets.
}\label{fig:heavy:cifar100}
\end{figure}

\begin{figure}[H]
\centering
\begin{tabular}{c}
\includegraphics[width=1.0\textwidth]{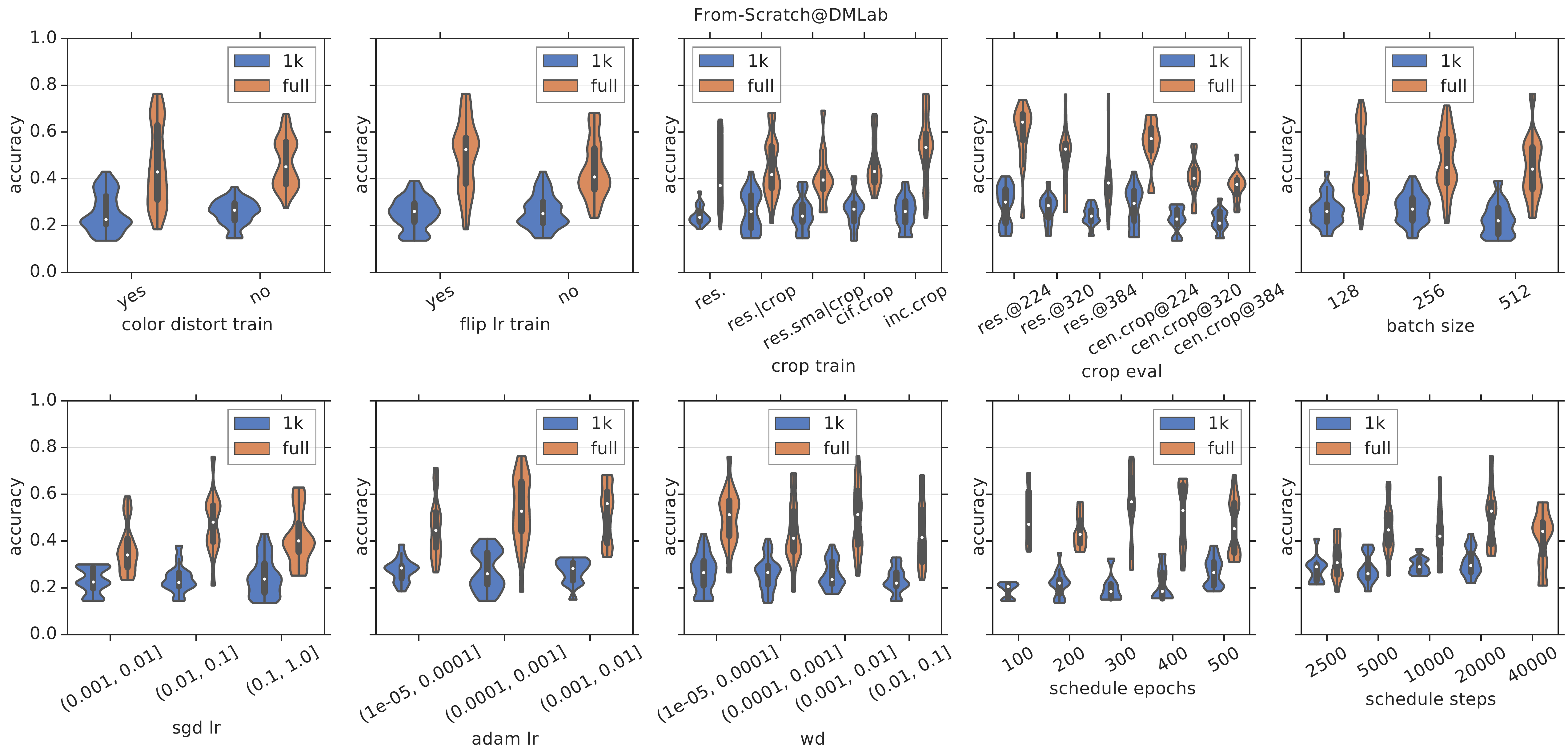}\\
\includegraphics[width=1.0\textwidth]{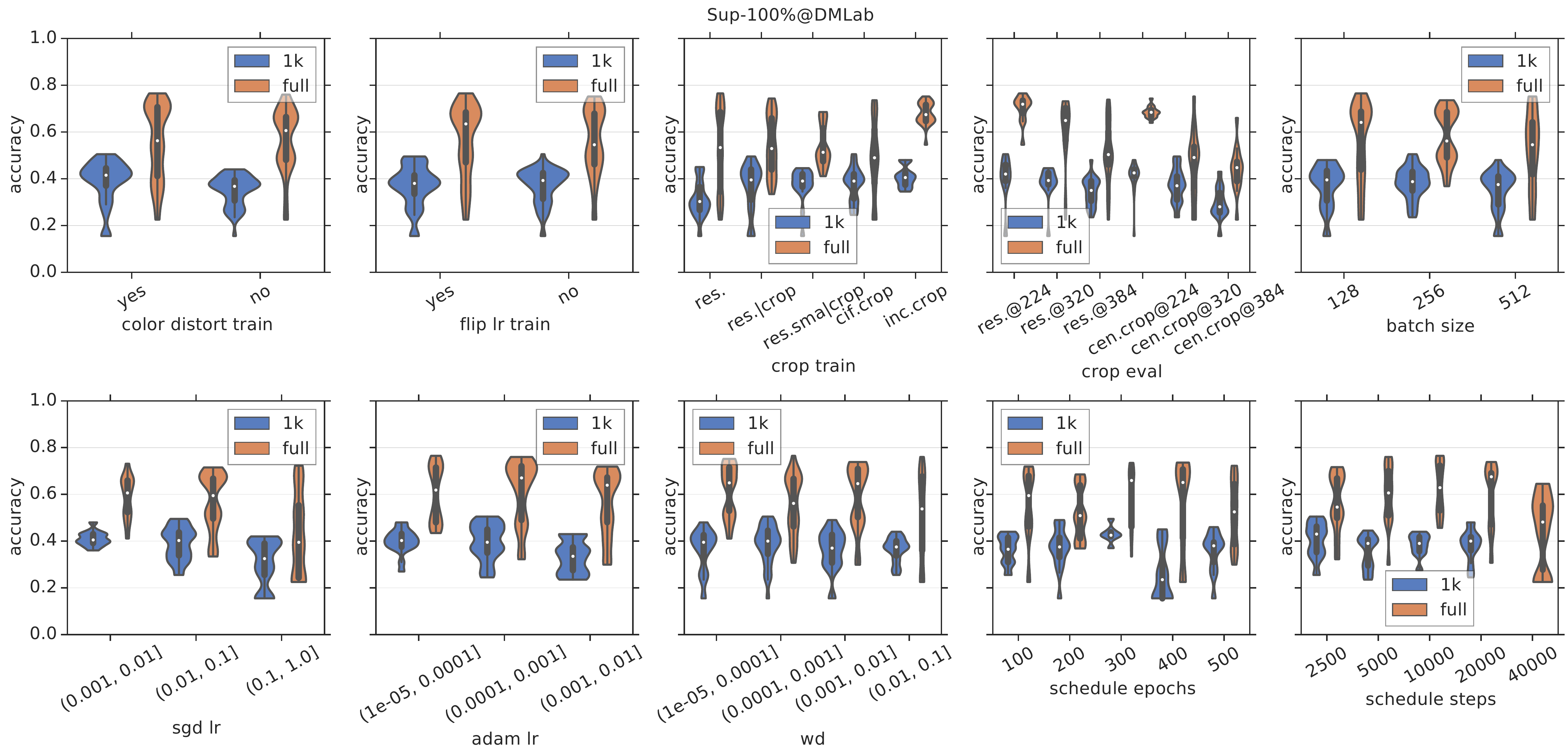}\\
\end{tabular}
\caption{
Heavy parameter sweep on DMLab dataset.
Top 2 rows show the results of from scratch algorithm and bottom 2 rows show the results of supervised 100\% algorithm.
Each box shows the violin plot of the results with respect to a given parameter.
We show the results evaluated on full datasets and 1k datasets.
}\label{fig:heavy:dmlab}
\end{figure}

In \cref{fig:heavy:cifar100,fig:heavy:dmlab}, we show the achieved performance for the full distribution of hyperparameters as violin plots for the CIFAR-100 and DMLab downstream tasks, both for the small and full variants.
The supervised pre-training not only achieves better accuracy than training from scratch, but it also displays much smaller variance across hyperparameter values.
\clearpage

\begin{figure}[!h]
\centering
\includegraphics[width=0.9\textwidth]{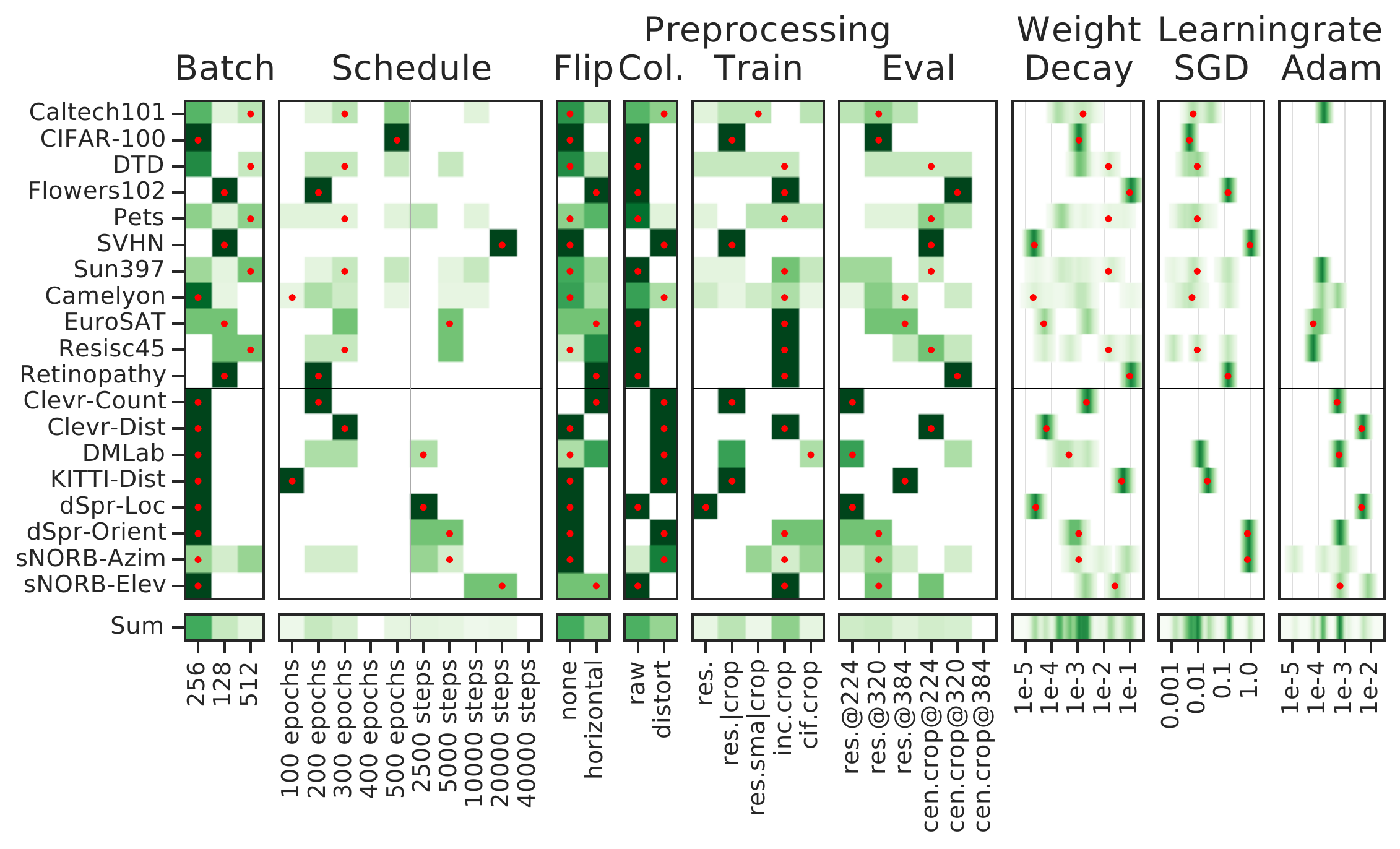}
\caption{\imagenet{} supervised representation fine-tuned on the full datasets. Best hyperparameter values are marked in red and those within $2\%$ in green to show sensitivity.}\label{fig:heavy:sup:full}
\end{figure}

\clearpage
\section{Lightweight versus Heavyweight Searches\label{app:sweeps}}

\begin{figure}[H]
\centering
\begin{tabular}{cc}
\includegraphics[width=0.35\textwidth]{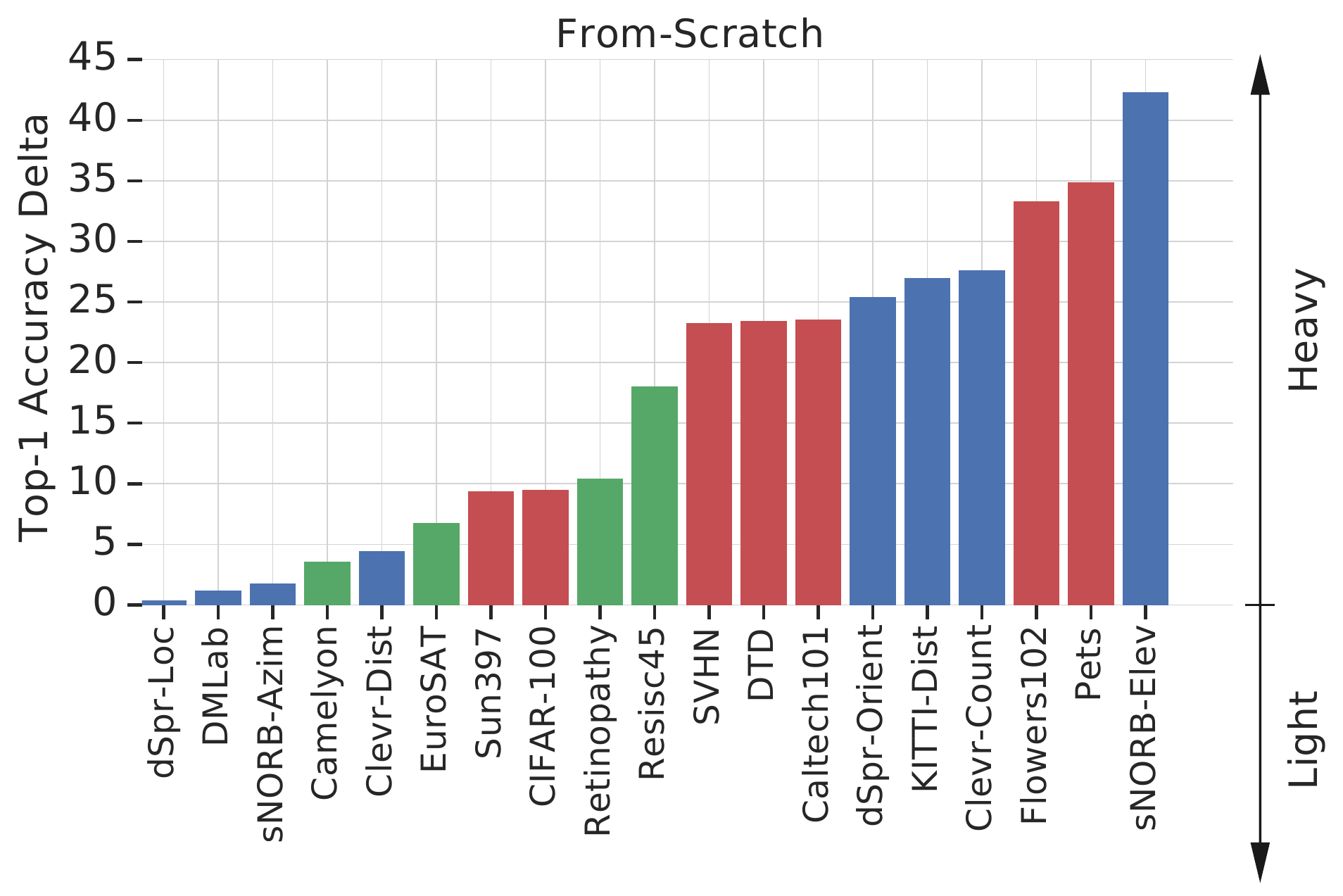}&
\includegraphics[width=0.35\textwidth]{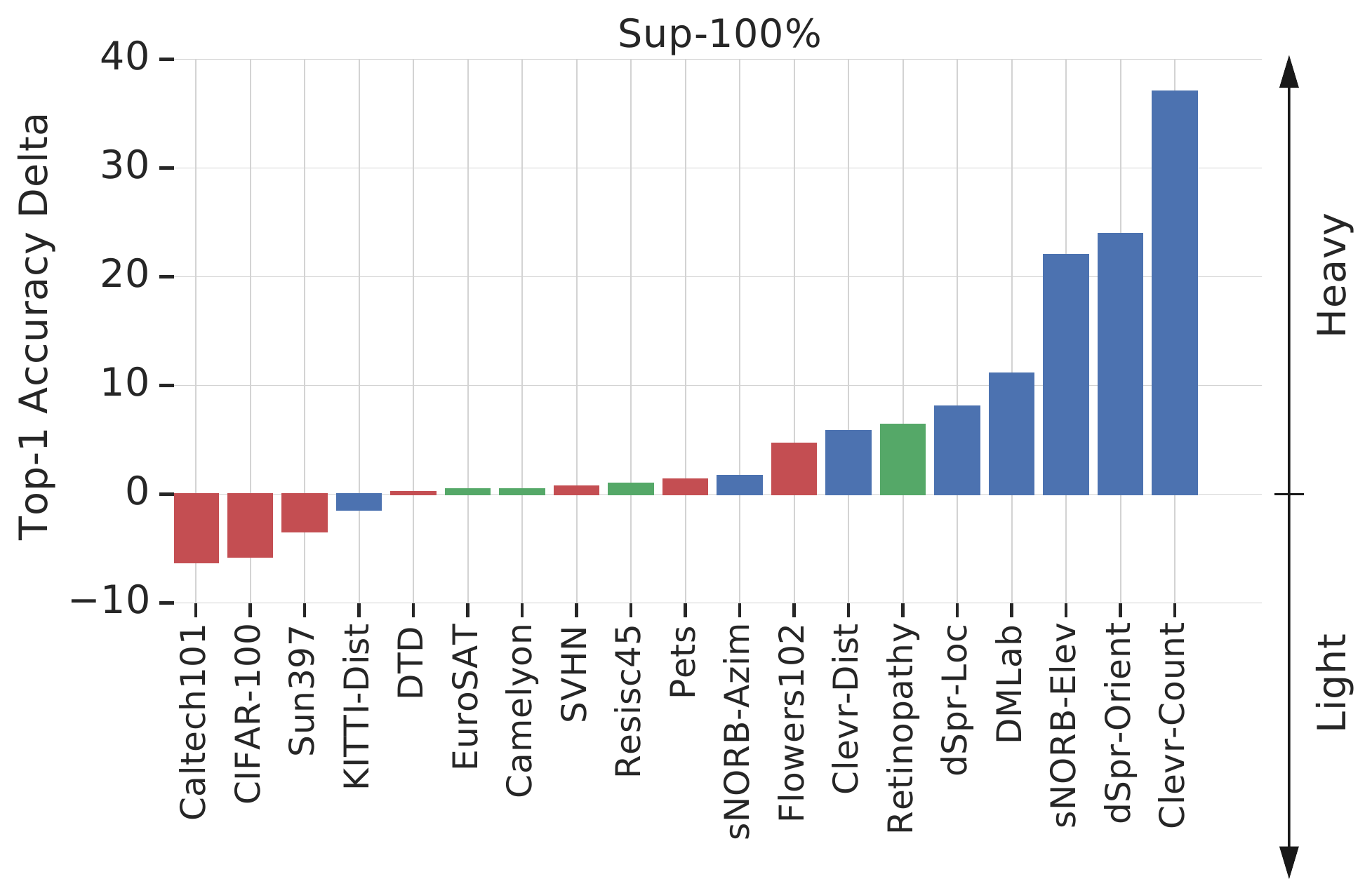}\\
\includegraphics[width=0.35\textwidth]{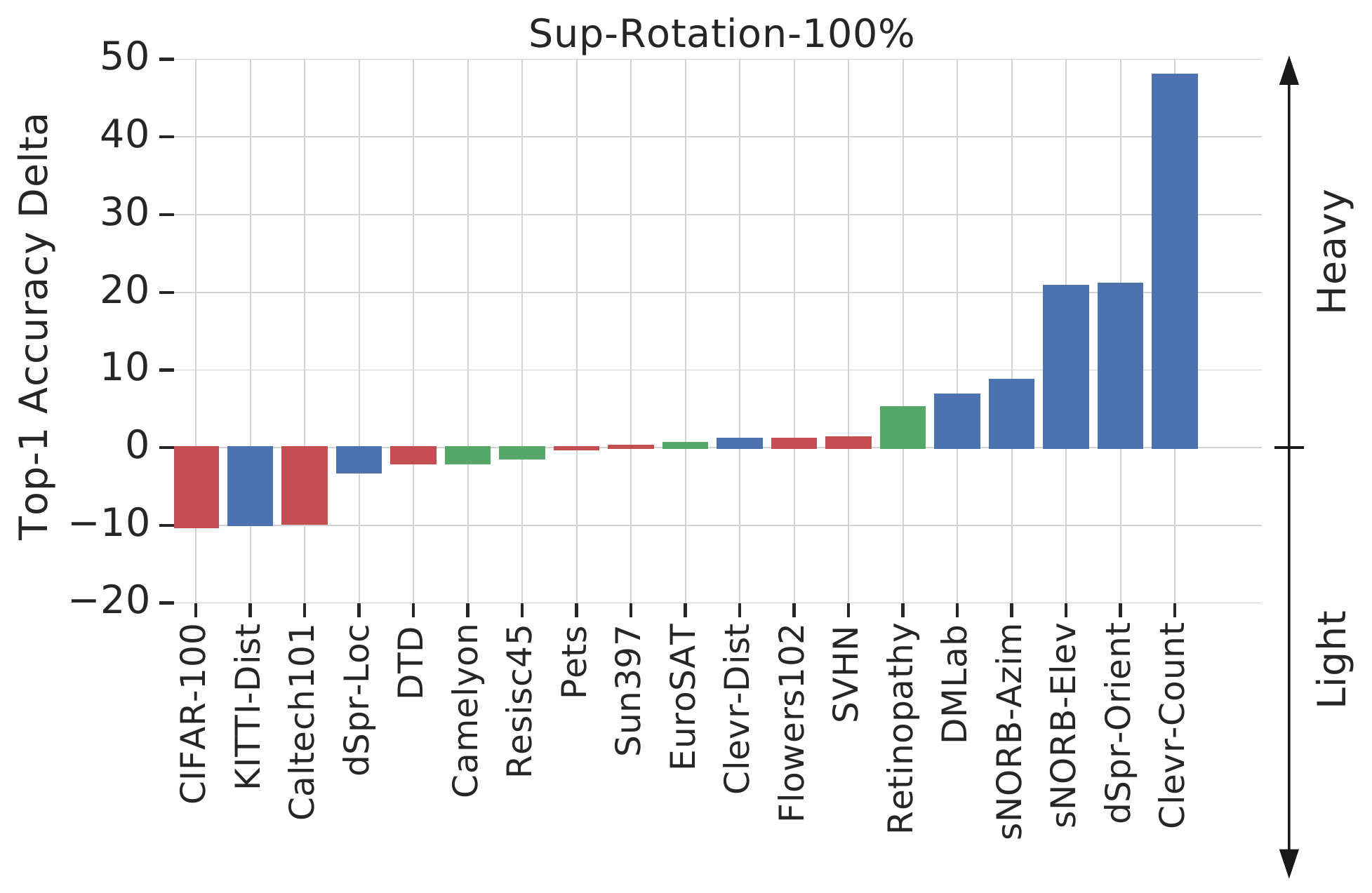}&
\includegraphics[width=0.35\textwidth]{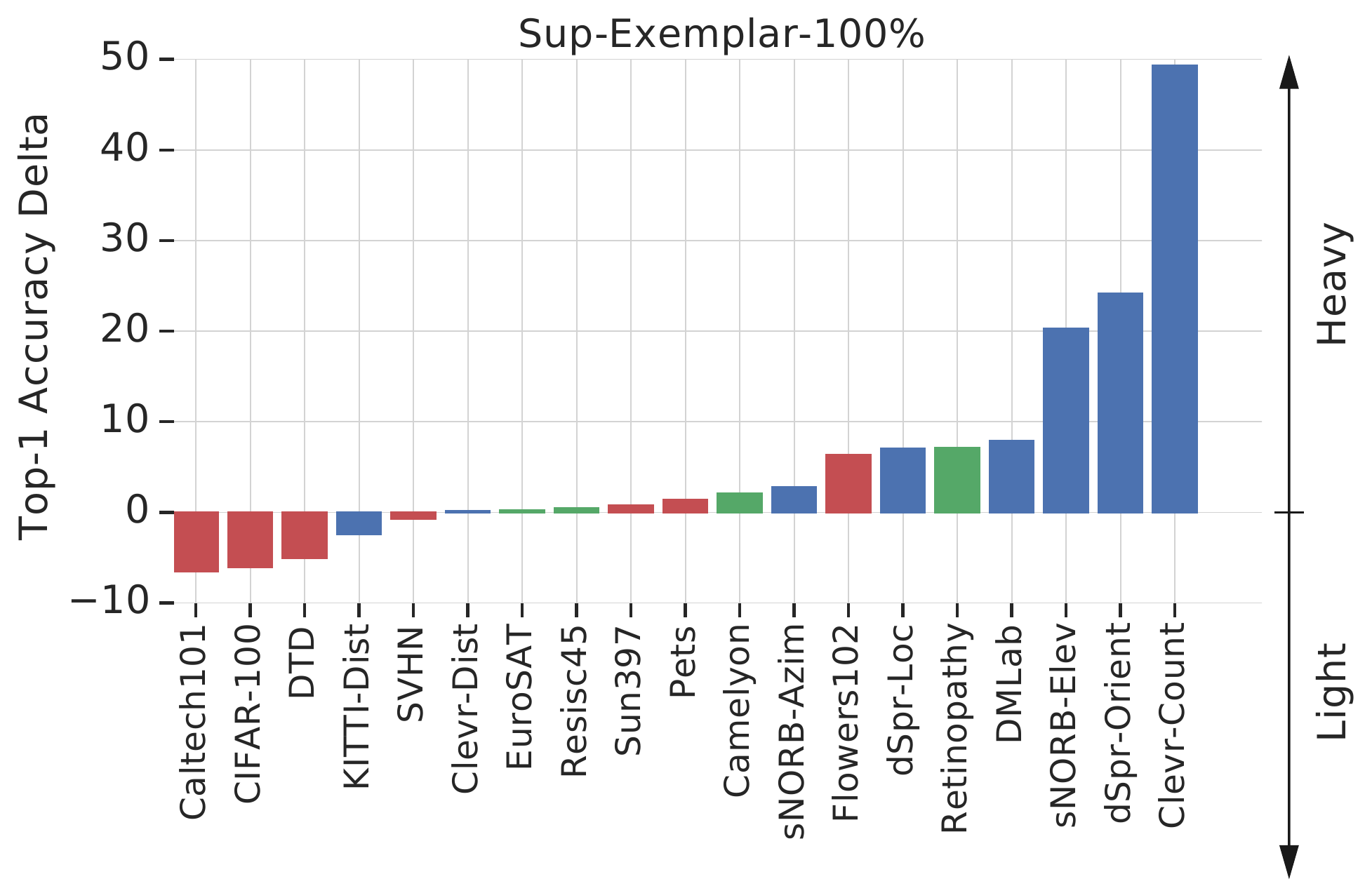}
\end{tabular}
\caption{
Per-dataset, absolute difference in top-1 accuracy of two hyper parameter sweep strategies.
Top left: \textsc{from-scratch}.
Top right: \textsc{sup-100\%}.
Bottom left: \textsc{sup-rotation-100\%}.
Bottom right: \textsc{sup-exemplar-100\%}
The bar colour denotes the task group.
}
\label{fig:atari-heavy-light}
\end{figure}

\begin{figure}[H]
\centering
\begin{tabular}{cc}
    \includegraphics[width=0.35\textwidth]{plots/atari/atari_Sup-100_v_From-Scratch_1000_light_test.pdf}&
    \includegraphics[width=0.35\textwidth]{plots/atari/atari_Sup-Rotation-100_v_Sup-100_1000_light_test.pdf}\\
    \includegraphics[width=0.35\textwidth]{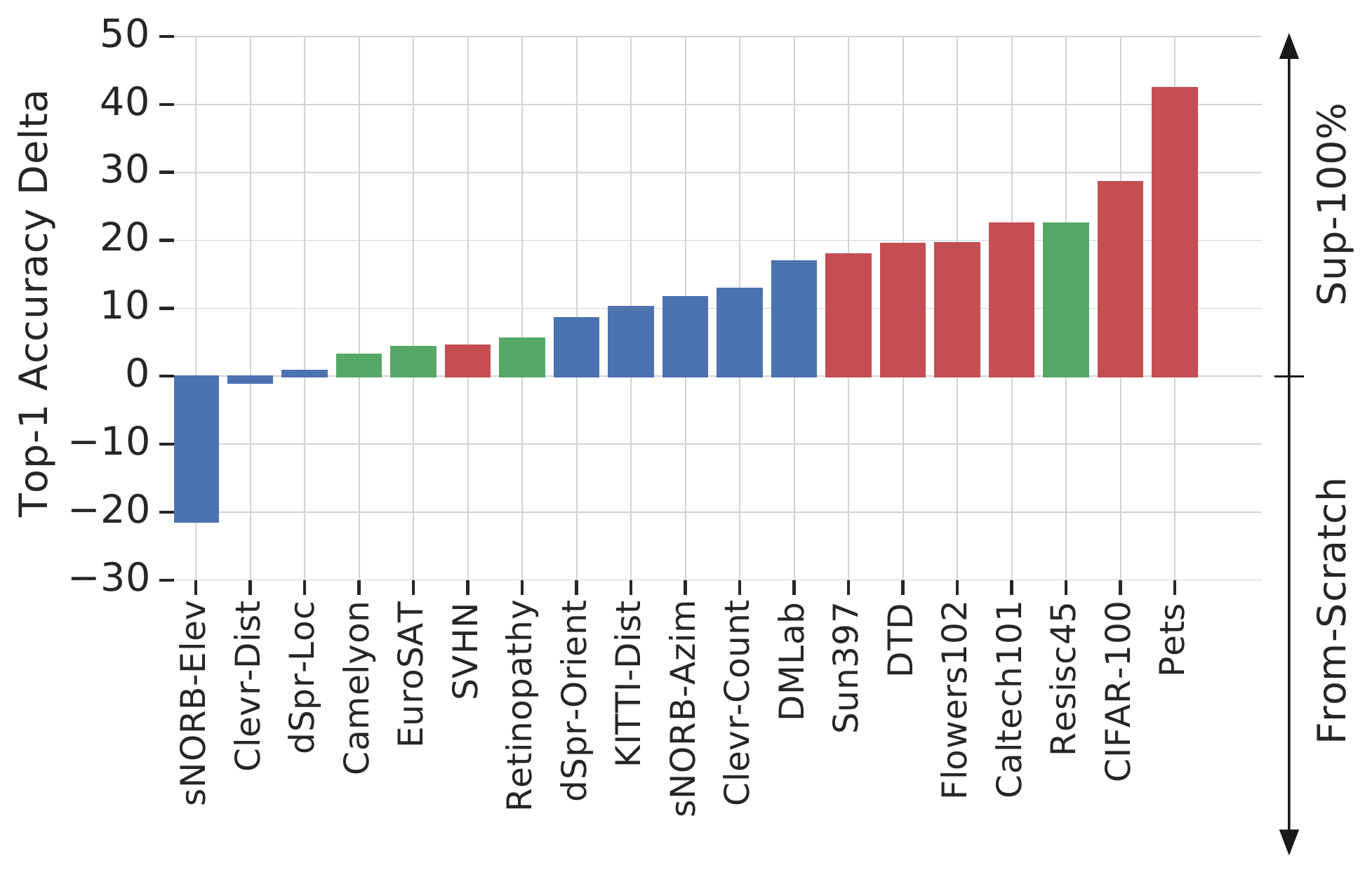}&
    \includegraphics[width=0.35\textwidth]{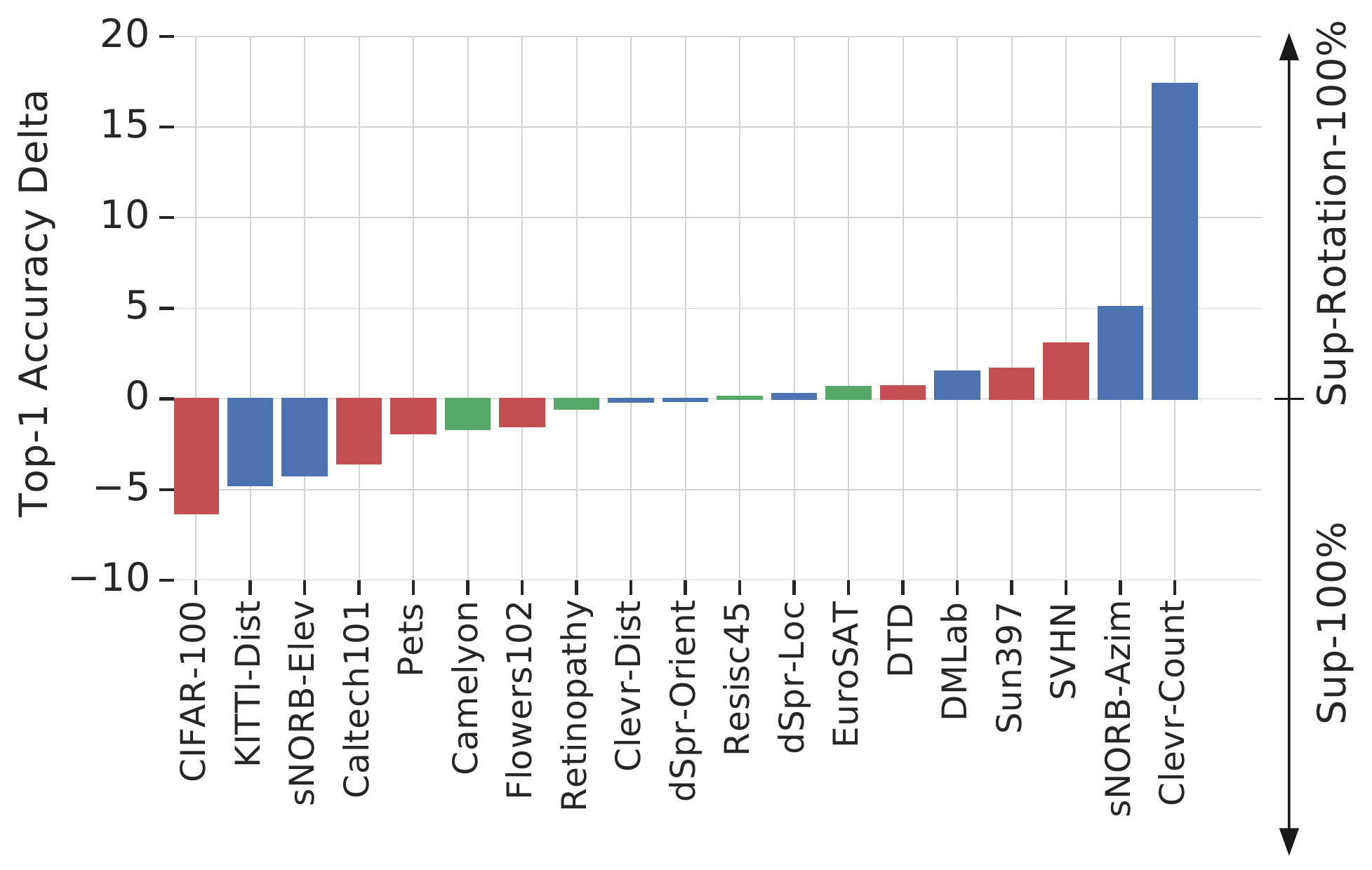}
\end{tabular}
\caption{
Absolute difference in top-1 accuracy between pairs of methods for each dataset.
The bar colour denotes the task group as usual.
Top: Lightweight hyperparameter sweep.
Bottom: Heavyweight hyperparameter sweep.
Left: \textsc{Sup-100\%} versus \textsc{From-Scratch} -- supervised pre-training yields a substantial improvement on the \taskNatural{} datasets and some others.
Right: \textsc{Sup-Rotation-100\%} versus \textsc{Sup-100\%} -- the additional self-supervised loss yields better representations for the \taskStructured{} tasks.
}
\label{fig:atari-all}
\end{figure}

\clearpage
\section{Linear Evaluation\label{app:linear}}

Here we describe the setup for linear evaluation. 
We follow ~\citep{kolesnikov2019revisiting} and evaluate the frozen representation by training a linear logistic regression model. 
We use exactly the same hyperparameters as described in lightweight sweep from Section~\ref{app:hyperparameter_sweep}.
The only difference here is that only the linear layer is trained instead of fine tuning the whole network. 

\begin{figure}[H]
\centering
\begin{tabular}{cc}
    \includegraphics[width=0.45\textwidth]{plots/finetune_vs_linear_1k.pdf}&
    \includegraphics[width=0.45\textwidth]{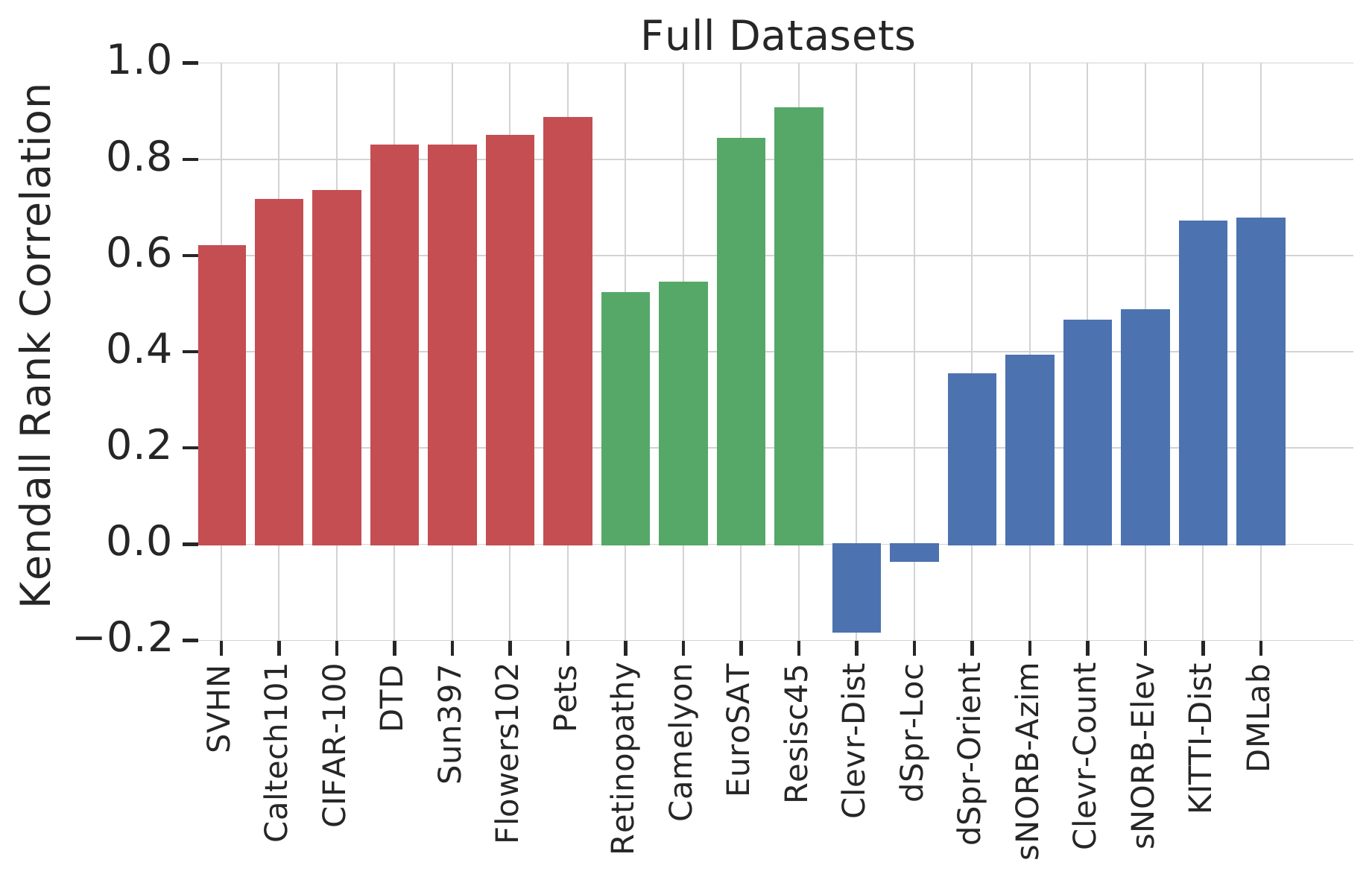}
\end{tabular}
\caption{Kendall rank correlation coefficient between finetuning and linear evaluation on each dataset.
}
\label{app:finetune_vs_linear}
\end{figure}

\begin{table}[H]
\fontsize{7pt}{7pt}\selectfont
\newcolumntype{C}{>{\centering\arraybackslash}X}
\setlength{\tabcolsep}{0pt}
\setlength{\extrarowheight}{5pt}
\renewcommand{\arraystretch}{0.75}
\begin{tabularx}{\linewidth}{p{10pt}p{1.6cm}!{\color{lightgray}\vline} CCCCCCC!{\color{lightgray}\vline}CCCC!{\color{lightgray}\vline}CCCCCCCC!{\color{lightgray}\vline}C}
\toprule
 &
 & \rotatebox{90}{\tikz\fill[natural] (0,0) circle (.5ex); Caltech101}
 & \rotatebox{90}{\tikz\fill[natural] (0,0) circle (.5ex); CIFAR-100}
 & \rotatebox{90}{\tikz\fill[natural] (0,0) circle (.5ex); DTD}
 & \rotatebox{90}{\tikz\fill[natural] (0,0) circle (.5ex); Flowers102}
 & \rotatebox{90}{\tikz\fill[natural] (0,0) circle (.5ex); Pets}
 & \rotatebox{90}{\tikz\fill[natural] (0,0) circle (.5ex); Sun397}
 & \rotatebox{90}{\tikz\fill[natural] (0,0) circle (.5ex); SVHN}
 & \rotatebox{90}{\tikz\fill[specialized] (0,0) circle (.5ex); Camelyon}
 & \rotatebox{90}{\tikz\fill[specialized] (0,0) circle (.5ex); EuroSAT}
 & \rotatebox{90}{\tikz\fill[specialized] (0,0) circle (.5ex); Resisc45}
 & \rotatebox{90}{\tikz\fill[specialized] (0,0) circle (.5ex); Retinopathy}
 & \rotatebox{90}{\tikz\fill[structured] (0,0) circle (.5ex); Clevr-Count}
 & \rotatebox{90}{\tikz\fill[structured] (0,0) circle (.5ex); Clevr-Dist}
 & \rotatebox{90}{\tikz\fill[structured] (0,0) circle (.5ex); DM-Lab}
 & \rotatebox{90}{\tikz\fill[structured] (0,0) circle (.5ex); dSpr-Loc}
 & \rotatebox{90}{\tikz\fill[structured] (0,0) circle (.5ex); dSpr-Ori}
 & \rotatebox{90}{\tikz\fill[structured] (0,0) circle (.5ex); KITTI-Dist}
 & \rotatebox{90}{\tikz\fill[structured] (0,0) circle (.5ex); sNORB-Azim}
 & \rotatebox{90}{\tikz\fill[structured] (0,0) circle (.5ex); sNORB-Elev}
 & \rotatebox{90}{\tikz\fill[all] (0,0) circle (.5ex); Mean} \\
 
\midrule

\multirow{16}{*}{\rotatebox{90}{\hspace*{-2pt}1000}}
& WAE-GAN &        33.9 &        7.1 &  6.8 &        10.6 &   6.7 &     2.8 &  24.1 &      64.4 &     42.3 &      10.7 &         71.3 &         27.6 &        52.1 &   26.1 &      59.7 &          8.4 &        40.7 &        16.0 &        19.9 &  28.0 \\
      & WAE-UKL &        36.7 &        7.8 &  6.8 &        10.4 &   7.3 &     3.2 &  22.7 &      63.0 &     40.8 &       9.8 &         73.3 &         26.8 &        52.3 &   25.1 &      61.5 &         10.1 &        41.0 &        15.3 &        20.7 &  28.1 \\
      & VAE &        36.0 &        7.9 &  7.7 &         9.5 &   7.7 &     3.5 &  21.8 &      65.4 &     38.5 &      13.8 &         73.5 &         27.0 &        51.4 &   25.6 &      61.0 &         11.1 &        39.1 &        17.8 &        21.5 &  28.4 \\
      & WAE-MMD &        36.9 &        9.3 &  7.3 &        13.6 &   8.5 &     3.8 &  22.4 &      65.1 &     43.6 &      14.4 &         71.7 &         27.9 &        52.9 &   26.5 &      66.5 &          9.8 &        42.1 &        19.0 &        22.9 &  29.7 \\
      & Un.C.-BigGAN &        50.0 &       14.0 & 30.2 &        37.5 &  11.4 &     8.2 &  38.4 &      77.5 &     69.5 &      30.1 &         67.2 &         33.9 &        47.8 &   29.3 &      30.6 &         23.5 &        31.6 &        19.9 &        28.1 &  35.7 \\
      & Jigsaw &        57.0 &       14.4 & 39.4 &        40.8 &  15.1 &     9.4 &  41.2 &      73.6 &     84.5 &      50.7 &         69.5 &         40.2 &        46.6 &   34.6 &      31.8 &         20.2 &        48.9 &        18.6 &        23.2 &  40.0 \\
      & Rel.Pat.Loc. &        62.0 &       15.5 & 44.7 &        47.9 &  21.7 &    12.5 &  39.8 &      73.5 &     86.6 &      53.7 &         71.6 &         41.0 &        46.9 &   34.8 &      28.8 &         19.7 &        46.4 &        21.3 &        25.6 &  41.8 \\
      & Cond-BigGAN &        63.4 &       20.7 & 34.5 &        60.6 &  23.2 &    11.9 &  47.3 &      71.4 &     75.3 &      42.8 &         63.6 &         40.6 &        48.0 &   31.0 &      40.4 &         42.9 &        35.6 &        24.1 &        31.3 &  42.5 \\
      & Exemplar &        57.0 &       20.7 & 42.7 &        55.8 &  26.0 &    13.5 &  37.8 &      80.8 &     88.4 &      59.4 &         73.7 &         45.4 &        49.6 &   33.8 &      55.0 &         28.9 &        58.7 &        18.7 &        32.2 &  46.2 \\
      & Rotation &        67.7 &       23.0 & 44.5 &        48.1 &  18.3 &    13.4 &  53.0 &      78.2 &     86.6 &      50.8 &         70.3 &         41.2 &        50.9 &   33.1 &      60.0 &         30.2 &        60.3 &        21.7 &        37.2 &  46.8 \\
      & Sup-10\% &        80.1 &       39.0 & 53.2 &        71.6 &  79.2 &    26.1 &  45.3 &      77.6 &     90.0 &      65.1 &         61.3 &         40.6 &        42.3 &   32.2 &      35.2 &         30.5 &        60.8 &        19.6 &        25.2 &  51.3 \\
      & Semi-Ex-10\% &        82.6 &       39.9 & 56.2 &        74.8 &  82.2 &    28.1 &  49.3 &      79.3 &     88.9 &      65.1 &         61.3 &         35.3 &        39.7 &   33.5 &      29.6 &         35.4 &        64.2 &        18.8 &        24.8 &  52.0 \\
      & Semi-Rot-10\% &        82.0 &       39.0 & 57.6 &        77.0 &  81.6 &    27.5 &  40.7 &      79.1 &     91.2 &      70.9 &         64.3 &         40.4 &        39.1 &   34.5 &      45.1 &         30.7 &        56.8 &        16.9 &        25.9 &  52.6 \\
      & BigBiGAN &        82.2 &       34.8 & 55.5 &        77.6 &  42.1 &    21.0 &  69.8 &      77.4 &     93.1 &      71.4 &         71.9 &         45.2 &        55.2 &   37.6 &      57.4 &         31.7 &        62.8 &        24.3 &        33.0 &  54.9 \\
      & Sup-Rot-100\% &        86.6 &       50.8 & 65.1 &        86.1 &  88.6 &    35.7 &  42.7 &      79.0 &     92.7 &      76.7 &         63.0 &         40.2 &        35.7 &   36.1 &      40.2 &         34.8 &        66.3 &        18.4 &        26.0 &  56.0 \\
      & Sup-100\% &        88.0 &       51.6 & 66.6 &        82.9 &  87.7 &    36.4 &  50.5 &      81.1 &     91.8 &      73.2 &         65.1 &         40.7 &        38.2 &   36.7 &      42.7 &         37.1 &        67.7 &        20.4 &        25.6 &  57.1 \\
      & Sup-Ex-100\% &        88.1 &       51.4 & 63.5 &        83.8 &  88.4 &    34.9 &  54.2 &      80.8 &     92.3 &      72.4 &         63.8 &         39.2 &        39.6 &   37.5 &      41.4 &         38.6 &        70.3 &        19.7 &        25.8 &  57.1 \\

\arrayrulecolor{lightgray}\specialrule{.5pt}{0.6pt}{-0.5pt}\arrayrulecolor{black}

\multirow{16}{*}{\rotatebox{90}{\hspace*{-4pt}Full}}
& VAE &        42.9 &       16.7 &  8.4 &        12.3 &   9.6 &     7.8 &  24.6 &      69.4 &     45.9 &      20.1 &         73.6 &         30.7 &        57.3 &   30.3 &      79.7 &         13.6 &        41.0 &        20.1 &        25.9 &  33.1 \\
      & WAE-UKL &        42.2 &       17.3 &  7.8 &        13.1 &   9.1 &     8.2 &  24.8 &      67.5 &     49.7 &      21.3 &         73.6 &         31.0 &        59.5 &   31.1 &      82.0 &         13.9 &        45.3 &        21.0 &        26.2 &  33.9 \\
      & WAE-GAN &        40.3 &       19.5 &  8.4 &        14.4 &   9.0 &     8.7 &  27.5 &      70.1 &     57.4 &      22.1 &         73.6 &         34.0 &        62.3 &   32.9 &      82.7 &         13.0 &        48.9 &        21.5 &        29.4 &  35.6 \\
      & WAE-MMD &        43.3 &       19.4 &  8.6 &        17.4 &   9.3 &     9.5 &  26.2 &      71.0 &     55.0 &      23.6 &         73.6 &         33.0 &        62.1 &   31.9 &      85.5 &         13.7 &        50.6 &        22.1 &        27.7 &  36.0 \\
      & Un.C.-BigGAN &        60.5 &       35.7 & 38.0 &        49.1 &  17.1 &    26.7 &  55.3 &      79.4 &     77.1 &      50.9 &         74.1 &         40.4 &        55.6 &   37.3 &      57.4 &         40.4 &        30.9 &        32.5 &        39.2 &  47.2 \\
      & Jigsaw &        68.7 &       29.9 & 49.3 &        53.4 &  27.7 &    29.2 &  50.6 &      78.2 &     90.6 &      70.2 &         73.6 &         49.7 &        57.2 &   41.7 &      40.4 &         23.0 &        55.2 &        31.2 &        36.6 &  50.3 \\
      & Rel.Pat.Loc. &        73.2 &       29.0 & 55.5 &        57.5 &  29.7 &    27.8 &  50.2 &      77.4 &     90.7 &      70.2 &         74.5 &         49.7 &        57.7 &   42.8 &      38.5 &         20.7 &        53.0 &        32.7 &        36.3 &  50.9 \\
      & Exemplar &        68.2 &       49.2 & 52.2 &        67.4 &  38.3 &    43.7 &  59.3 &      81.5 &     94.2 &      79.9 &         74.8 &         56.3 &        58.8 &   41.4 &      70.6 &         34.2 &        62.8 &        32.3 &        45.0 &  58.4 \\
      & Cond-BigGAN &        73.6 &       47.9 & 43.5 &        70.1 &  29.4 &    39.4 &  67.7 &      76.3 &     84.6 &      66.9 &         69.7 &         52.1 &        60.4 &   39.7 &      75.5 &         64.9 &        43.7 &        67.3 &        52.7 &  59.2 \\
      & Rotation &        77.5 &       48.4 & 56.2 &        60.2 &  30.6 &    42.1 &  71.6 &      82.0 &     93.4 &      74.7 &         74.7 &         56.6 &        64.8 &   46.0 &      75.1 &         36.1 &        62.0 &        38.3 &        52.8 &  60.2 \\
      & BigBiGAN &        88.9 &       56.7 & 67.1 &        84.2 &  54.7 &    46.2 &  80.0 &      81.3 &     95.8 &      85.1 &         75.4 &         62.7 &        64.2 &   49.2 &      79.6 &         38.6 &        67.7 &        29.9 &        42.6 &  65.8 \\
      & Sup-10\% &        85.5 &       64.4 & 62.3 &        79.2 &  82.8 &    57.2 &  63.9 &      81.9 &     95.6 &      83.9 &         74.4 &         56.9 &        57.3 &   45.2 &      56.3 &         43.5 &        62.8 &        56.5 &        47.0 &  66.1 \\
      & Semi-Rot-10\% &        86.7 &       65.6 & 65.5 &        83.6 &  84.6 &    58.8 &  62.2 &      84.0 &     96.0 &      87.6 &         74.6 &         56.8 &        53.1 &   45.9 &      69.7 &         45.7 &        64.9 &        41.8 &        44.1 &  66.9 \\
      & Semi-Ex-10\% &        86.8 &       64.6 & 63.6 &        80.7 &  85.3 &    57.7 &  68.7 &      83.2 &     94.5 &      82.8 &         74.3 &         52.1 &        54.6 &   46.8 &      67.2 &         65.2 &        74.1 &        68.8 &        41.7 &  69.1 \\
      & Sup-Rot-100\% &        91.8 &       74.3 & 72.2 &        90.8 &  91.1 &    68.2 &  63.4 &      82.5 &     96.8 &      90.6 &         74.8 &         63.7 &        51.9 &   49.4 &      75.3 &         60.2 &        71.4 &        61.5 &        44.7 &  72.3 \\
      & Sup-100\% &        91.6 &       74.9 & 72.6 &        88.7 &  91.0 &    68.0 &  69.7 &      84.4 &     96.5 &      88.6 &         76.0 &         61.7 &        55.1 &   50.6 &      75.4 &         62.2 &        74.4 &        72.0 &        44.1 &  73.6 \\
      & Sup-Ex-100\% &        92.5 &       74.1 & 72.1 &        89.2 &  91.6 &    67.6 &  73.6 &      84.0 &     96.0 &      87.9 &         75.0 &         62.2 &        57.9 &   51.4 &      75.6 &         62.0 &        74.1 &        73.2 &        46.5 &  74.0 \\

\bottomrule
\end{tabularx}

\caption{Top-1 accuracy of all the models with linear evaluation on VTAB.}
\label{tab:lightweight_linear_test}
\end{table}

\clearpage
\section{Larger Architectures\label{app:scale-architecture}}

\begin{minipage}{\textwidth}
\begin{figure}[H]
    \centering
    \begin{tabular}{cc}
    \includegraphics[width=0.45\textwidth]{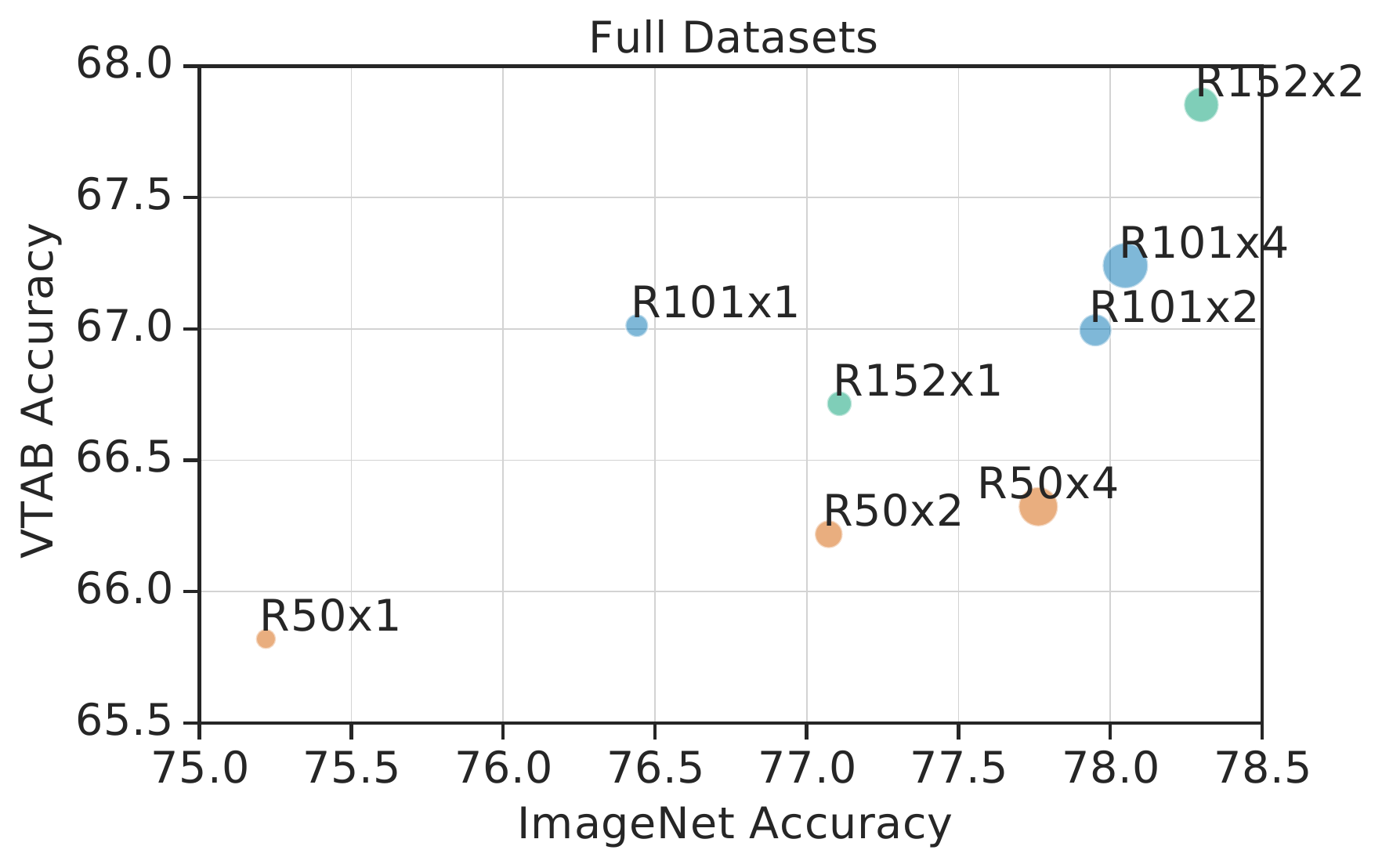}
    \end{tabular}
    \caption{
    VTAB performance on the scaled up architectures.
    Here the architecture is represented by different color and width is represented by different circle size.
    The performance increases when switching from the standard ResNet50 architecture to ResNet152 2x wider architecture.}
    \label{fig:scale-up-downstream}
\end{figure}
\end{minipage}

\begin{figure}[ht]
    \centering
    \includegraphics[width=0.5\textwidth]{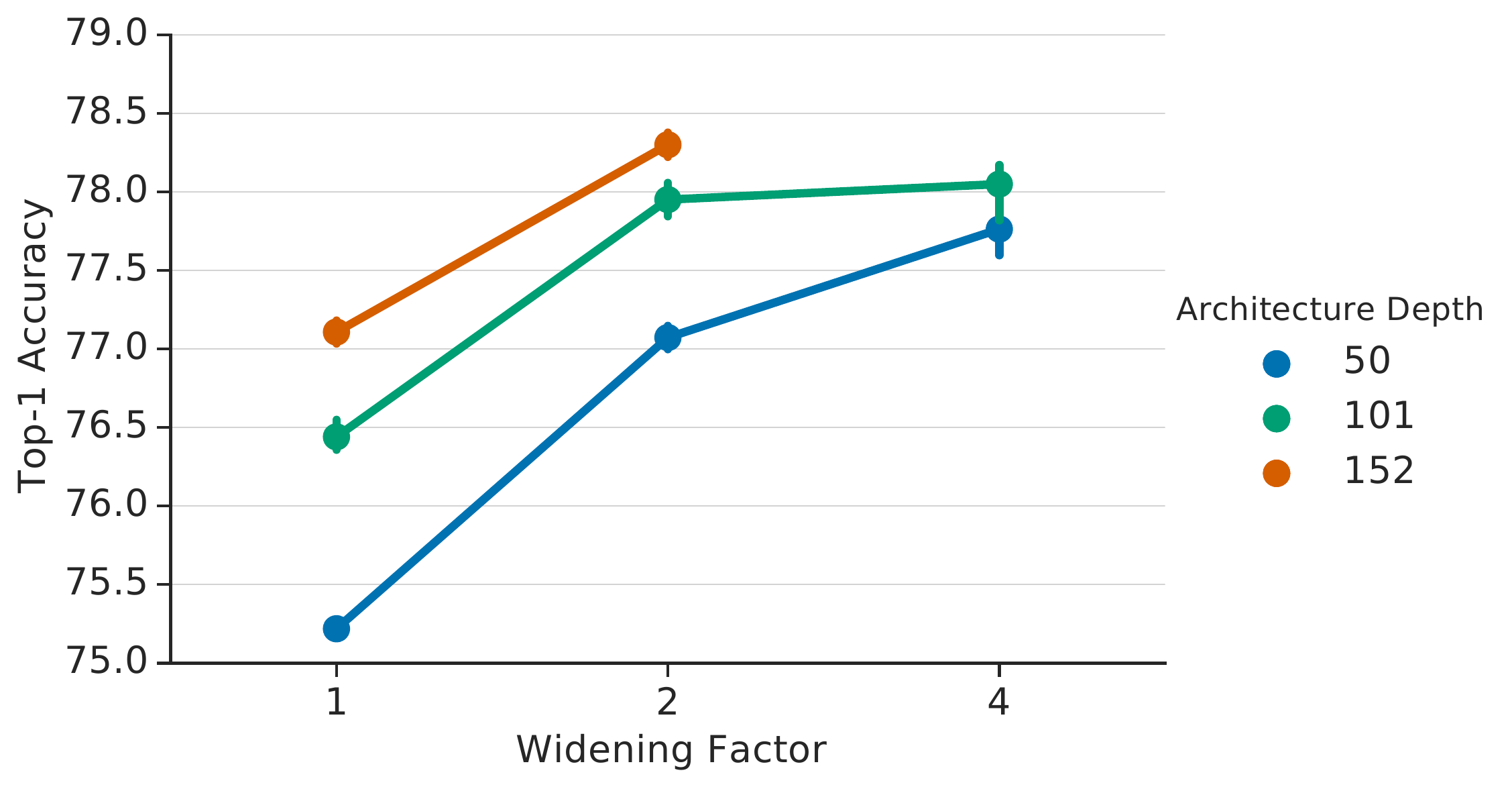}
    \caption{Top-1 accuracy on \imagenet{} public validation set when scaling up the architectures. The accuracy goes up with either wider or deeper architecture. 
    }
    \label{fig:scale-up-upstream}
\end{figure}

In this section, we study the problem of ``is scaling up the architectures helpful?". Figure~\ref{fig:scale-up-upstream} shows top-1 accuracy on \imagenet{} public validation set when scaling up the architectures. 
Widening factor is the multiplier on the network width, where $\times 1$ stands for the standard ResNet architecture.
Depth stands for the number of architecture layers. 
As expected, the model accuracy goes up with either wider or deeper architectures. 
Figure~\ref{fig:scale-up-downstream} shows the results on VTAB benchmark, where the 2x wider ResNet152 architecture performs consistently better than the standard ResNet50 model. 

\clearpage
\section{Budget Analysis\label{app:budget}}

\begin{figure}[H]
    \centering
    \includegraphics[width=0.85\textwidth]{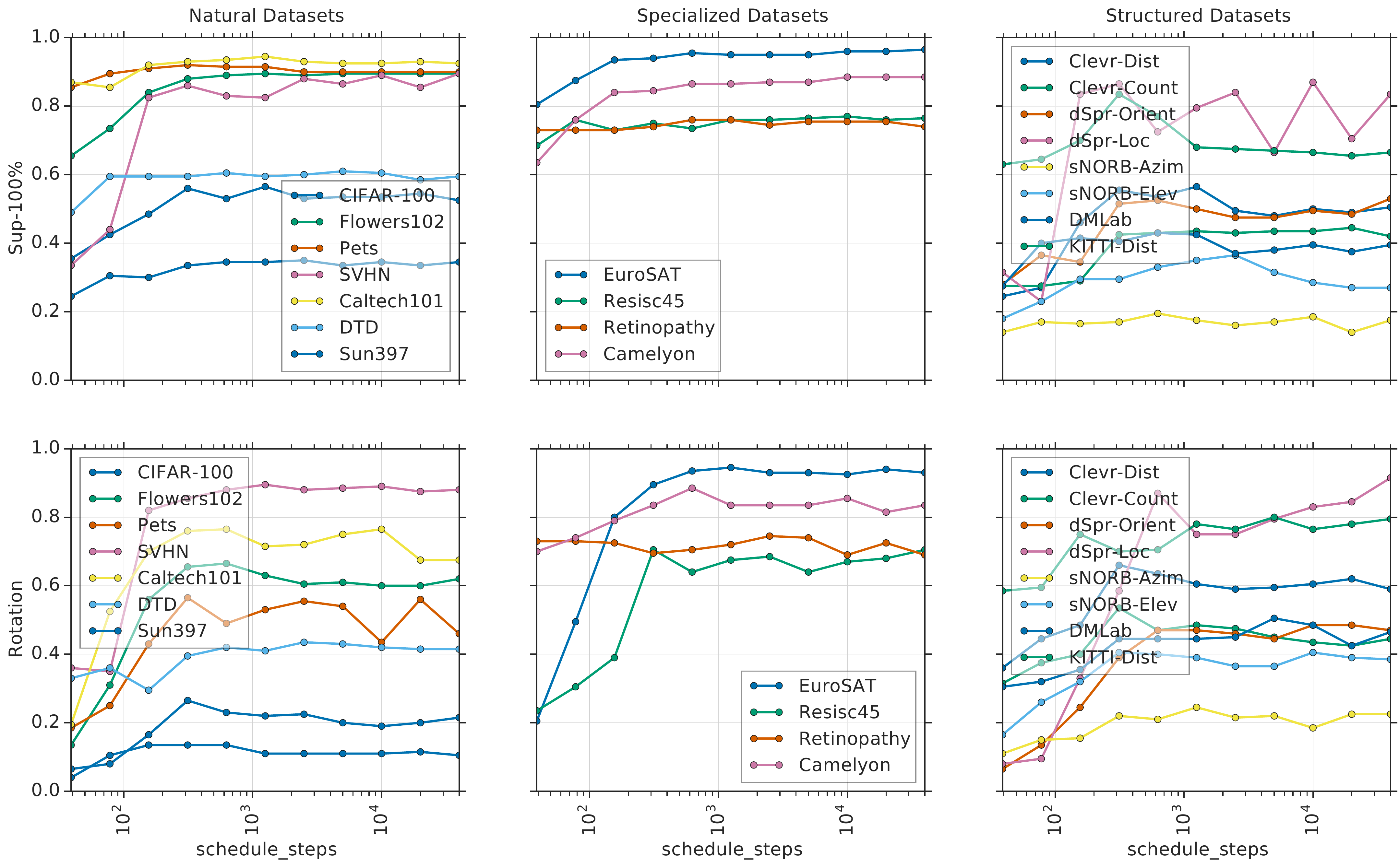}
    \caption{
    Top-1 accuracy for each task attained by \textsc{Sup-100\%} and \textsc{Rotation} with respect to the number of fine-tuning steps on the 1000-example datasets.
    More steps is usually better, but performance is stable after 1000 steps.}
    \label{fig:budget-plot-1k}

    \vspace{5mm}

    \centering
    \includegraphics[width=0.85\textwidth]{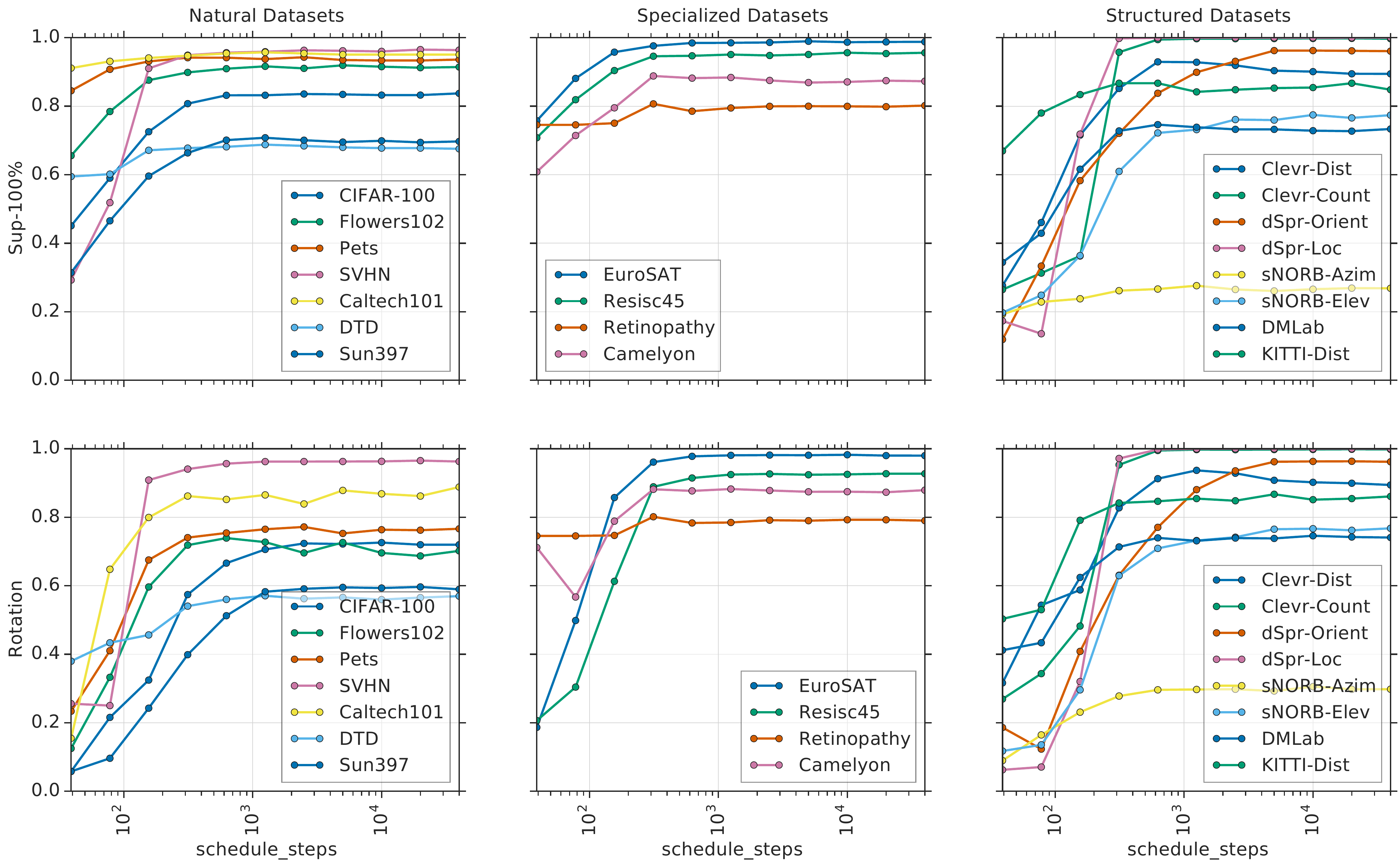}
    \caption{
    Top-1 accuracy for each task attained by \textsc{Sup-100\%} and \textsc{Rotation} with respect to the number of fine-tuning steps on the full datasets.
    More steps is usually better, but performance is stable after 1000 steps.}
    \label{fig:budget-plot-full}
\end{figure}

\clearpage
\section{Comparison to Visual Decathlon}
\label{sec:visual_decathlon}

\begin{figure}[h]
\centering
\includegraphics[width=0.65\linewidth]{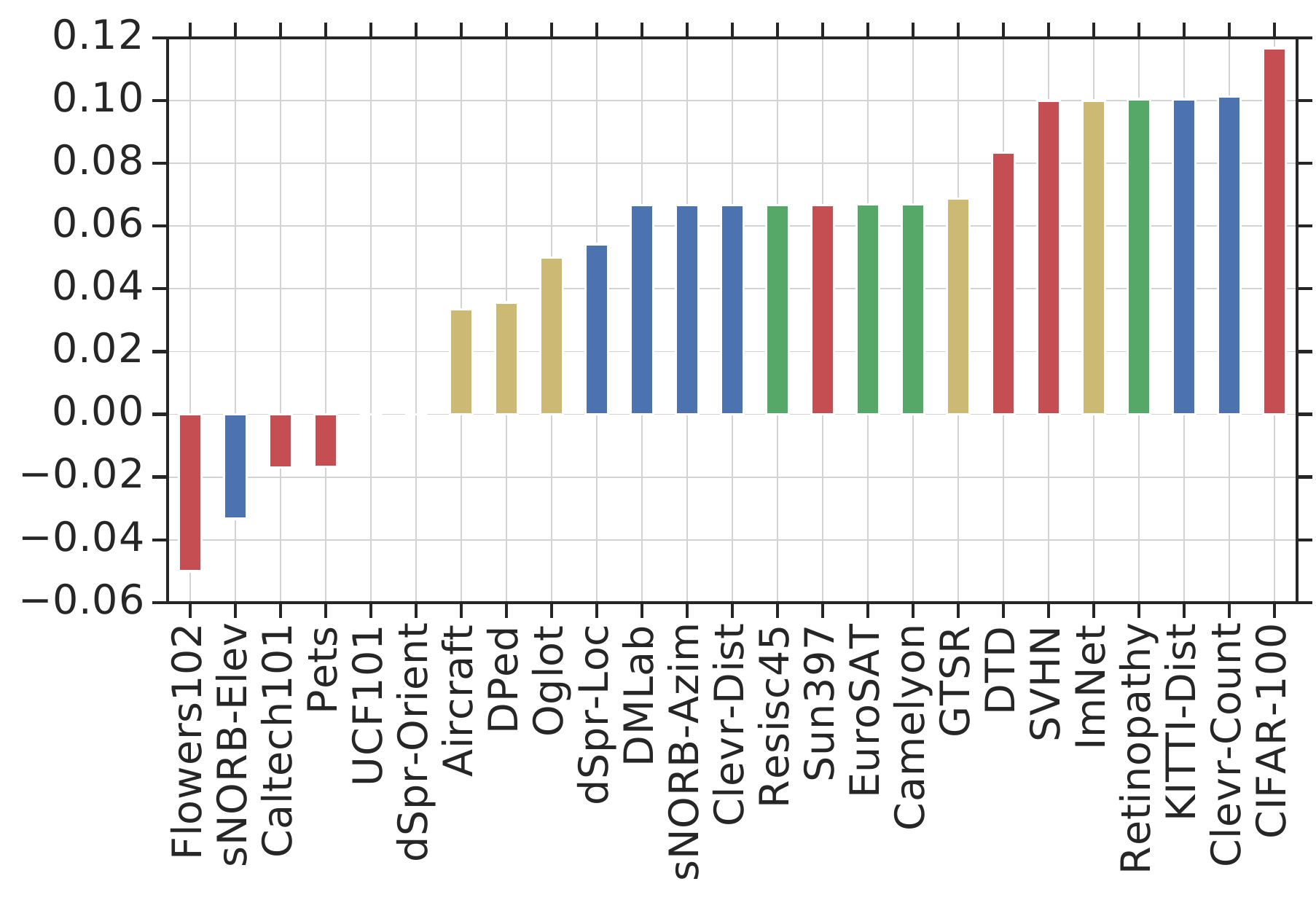}
\caption{%
  Absolute differences in Kendall's rank correlation score between the ``gold'' ranking
  in each dataset, and the ranking obtained with either VTAB or Visual Decathlon.
  Bar colors indicate the category as usual, and yellow indicates datasets only present in Visual Decathlon.
  Positive values indicate that VTAB ranking is closer than Visual Decathlon ranking, when
  compared against the ``gold'' ranking of a particular dataset.
  The average ranking correlation for VTAB is 0.76, and 0.70 for Visual Decathlon.
}
\label{fig:rank_correlation_vtab_vs_visual_decathlon}
\end{figure}

In \cref{sec:related_work} we describe the differences between VTAB
and Visual Decathlon protocols. Here we answer a more practical question: 
which benchmark should one use to compare the adaptation abilities of a set of methods
to unseen tasks? 
We will show that the rank correlation with an unseen task, is expected
to be higher for the ranking obtained using VTAB, than using Visual Decathlon.

First, we fine-tuned each of our 16 baseline models in each of the Visual 
Decathlon datasets (using the lightweight hyperparameter search described in \cref{sec:setup}),
which we downloaded directly from the competition's website. All datasets in the 
Visual Decathlon are provided so that the shorter size of the image is 72 pixels, 
while our baseline models were trained in much larger resolutions.
To overcome this difference in resolution, 
we resize the Visual Decathlon images to the resolution required by each model.
We checked that our baseline models obtained reasonable results on the Visual Decathlon 
benchmark. In fact, our best model,
\textsc{sup-rotation-100\%}, reports a test average accuracy of 78.22\%, and a decathlon score
of 3580, which are both slightly better than the best results reported in
\cite{rebuffi2018} (78.08\% and 3412, respectively), and other works
(e.g. \cite{rebuffi2017,rosenfeld2018incremental}). However, notice that comparing
these numbers is delicate, since we did not use any data augmentation during
training and all our models
are based on the Resnet50 architecture, while these works use heavy data augmentation
(that depends on the dataset), and Resnet26-like architectures.

Then, for each task $T$ in the union of VTAB and Visual Decathlon, we rank the 16 baseline
methods according to their accuracy on task $T$. If we consider $T$ as the unseen dataset,
this is the ``gold'' ranking of the studied methods. Now, we obtain two alternative
rankings: one based on the mean accuracy in VTAB and another on Visual Decathlon,
excluding task $T$ in both cases to avoid any bias.
We can then compute, for each ``unseen'' dataset, the Kendall's ranking correlation
between the ``gold'' ranking and each of the alternative rankings.
\Cref{fig:rank_correlation_vtab_vs_visual_decathlon} shows the absolute differences in the
rank correlation score between the VTAB-based ranking and Visual Decathlon-based ranking.

For most datasets, the difference is positive, which means that the ranking according to
VTAB correlates better with the ``gold'', than the raking obtained using Visual Decathlon.
Notice that even tasks that were not part of VTAB (colored in gold), are better represented
by VTAB's ranking than that of Visual Decathlon. These results are not surprising, since VTAB
contains a more representative set of tasks than Visual Decathlon.
The average ranking correlation for VTAB is 0.76, and 0.70 for Visual Decathlon.

\end{document}